\documentclass[twoside,11pt]{article}

\usepackage{jmlr2e}
\usepackage{cite}
\usepackage{bbold}
\usepackage{amsmath,amssymb,amsfonts}
\usepackage{algorithmic}
\usepackage{graphicx}
\usepackage{rotating}
\usepackage{textcomp}
\usepackage{xcolor}
\usepackage{caption}
\usepackage{framed}
\usepackage{bbold}
\usepackage{mathtools}
\usepackage{mathrsfs}
\makeatletter
\DeclareRobustCommand\bigop[1]{%
	\mathop{\vphantom{\sum}\mathpalette\bigop@{#1}}\slimits@
}
\newcommand{\bigop@}[2]{%
	\vcenter{%
		\sbox\z@{$#1\sum$}%
		\hbox{\resizebox{\ifx#1\displaystyle2.0\fi\dimexpr\ht\z@+\dp\z@}{!}{$\m@th#2$}}%
	}%
}
\makeatother

\makeatletter
\DeclareRobustCommand\normop[1]{%
	\mathop{\vphantom{\sum}\mathpalette\normop@{#1}}\slimits@
}
\newcommand{\normop@}[2]{%
	\vcenter{%
		\sbox\z@{$#1\sum$}%
		\hbox{\resizebox{\ifx#1\displaystyle0.4\fi\dimexpr\ht\z@+\dp\z@}{!}{$\m@th#2$}}%
	}%
}
\makeatother
	
\newcommand{\argmin}{\DOTSB\bigop{\large \text{argmin}}}

\firstpageno{1}

\begin{document}

\title{Stratified Graph Spectra}

\author{\name Fanchao Meng \email mf3jh@virginia.edu \\
       \addr Biocomplexity Institude\\
       University of Virginia\\
       Charlottesville, VA22904-4298, USA
       \AND
       \name Mark Orr \email mo6xj@virginia.edu \\
       \addr Biocomplexity Institude\\
       University of Virginia\\
       Charlottesville, VA22904-4298, USA
   	   \AND
       \name Samarth Swarup \email swarup@virginia.edu \\
       \addr Biocomplexity Institude\\
       University of Virginia\\
       Charlottesville, VA22904-4298, USA
   }

\editor{}
\maketitle

\begin{abstract}
In classic graph signal processing, given a real-valued graph signal, its graph Fourier transform is typically defined as the series of inner products between the signal and each eigenvector of the graph Laplacian. Unfortunately, this definition is not mathematically valid in the cases of \textit{vector-valued} graph signals which however are typical operands in the state-of-the-art graph learning modeling and analyses. Seeking a generalized transformation decoding the magnitudes of eigencomponents from vector-valued signals is thus the main objective of this paper. Several attempts are explored, and also it is found that performing the transformation at hierarchical levels of adjacency help profile the spectral characteristics of signals more insightfully. The proposed methods are introduced as a new tool assisting on diagnosing and profiling behaviors of graph learning models. 
\end{abstract}


\section{Introduction}\label{sec:introduction}
Graph spectral analysis techniques have been widely used in graph learning models. Particularly, graph signal processing (\citet{shuman2013emerging}; \citet{hammond2011wavelets}; \citet{ortega2018graph}) has become the corner stone of many state-of-the-art graph learning models (\citet{defferrard2016convolutional}; \citet{kipf2016semi}; \citet{xu2019graph}; \citet{bruna2013spectral}; \citet{velivckovic2017graph}; \citet{hamilton2017inductive}; \citet{levie2018cayleynets}). Most of the works in this area stipulate designs and diagnoses in various stages of graph learning modeling such as signal construction, transformation, filtering, aggregation, and other processes.  
\par
Designing and diagnosing graph learning models can be easily a painful course in practice, especially when the node embedding is playing a fundamental role, and the characteristics of the embedding vectors (as graph signals) in the ``frequency" domain are of particular interest. One of the major difficulties arises from decoding the magnitudes of eigencomponents from \textit{vector-valued graph signals}, which is the main problem to be addressed in this paper. 
\par
The graph Fourier transform (GFT) has become a standard tool in studying spectral characteristics of real-valued graph signals (where extracting magnitudes of eigencomponents is one of the core topics) (\citet{shuman2013emerging}; \citet{sandryhaila2013discrete}; \citet{sandryhaila2014big}; \citet{ortega2018graph}; \citet{hammond2011wavelets}; \citet{dong2019learning}). Directly extending the GFT to vector-valued signals is a natural thought. And actually, in practice, the GFT has been pervasively used in many state-of-the-art graph learning models such GCN (\citet{defferrard2016convolutional}; \citet{kipf2016semi}) to manipulate spectral features of vector-valued signals. However, such applications did not shed any light on analyzing spectral characteristics of vector-valued signals, if not making it even more confusing. On one hand, the GFT loses its original interpretation and convenience in reflecting magnitudes of eigencomponents when it is mechanically adopted to vector-valued signals even though the calculation looks seemingly valid.
Specifically, in the calculation, each dimension of a vector-valued signal is processed independently as a real-valued signal rather than taking the vector at each node atomically, and the output from the GFT thus consists of a vector, instead of a single value, for each eigencomponent \footnote{Representing nodes using higher (than $1$) dimensional vectors has been a popular practice. The interpretation of dimensions can be individual features (\citet{defferrard2016convolutional}), but it is vague in general. Why higher dimensions are useful and how many dimensions are actually needed are beyond the scope of this paper.}. These significant differences between applying the GFT to real-valued signals and vector-valued signals perplex the interpretation and usage of the GFT. On the other hand, most alternatives to the GFT, profiling vector-valued signals from the ``frequency" perspective, are intrinsically defective. Two typical alternatives, as examples, are briefly described below. 
\par
A straightforward approximation method is to generalize the quadratic form of the graph Laplacian thanks to a notion of signal smoothness. That is, for a real-valued signal $s$, $s^T \cdot \mathcal{L} \cdot s = \sum\limits_{\forall i \sim j} w_{ij} (s_i - s_j)^2$ provides a measure of smoothness of $s$, where $w_{ij}$'s are weights on edges (\citet{shuman2013emerging}). The smoothness implicitly reflects the spectral characteristics of $s$. For a vector-valued signal $\tilde{s}$, this smoothness can be generalized to $\tilde{s}^T \cdot \mathcal{L} \cdot \tilde{s} = \sum\limits_{\forall i \sim j} w_{ij} \big[ d(\tilde{s_i}, \tilde{s_j}) \big]^2$, where $d(\cdot, \cdot)$ is a distance function. The primary defect of this method is also straightforward. Two signals which have distinct spectral characteristics can result in the same value of their quadratic forms. For example, consider respectively, imposed on a 4-node cycle graph, a real-valued pulse signal $s_1=[\sqrt{2}, 0, 0, 0]$ and a real-valued oscillation signal $s_2=[2, 1, 2, 1]$. Although they share the same quadratic form values, it is intuitive to see that they could hardly have similar spectral characteristics. And such examples can also be found in the vector-valued signal cases. 
\par
Another alternative is comparing the input signal with a corresponding filtered signals. If the behaviors of the filter is known, then the difference between the input signal and the filtered signal partially implies the ``frequency" composition of the input signal. A graph filter can be expressed as a matrix function $\hat{h}(\mathcal{L}) := \mathcal{U} \cdot \hat{h}(\Lambda) \cdot \mathcal{U}^*$, where $\mathcal{U}$ is the matrix of considered eigenvectors, $\mathcal{U}^*$ denotes the conjugate transpose of $\mathcal{U}$, and $\hat{h}$ is a multivariate function adjusting the magnitudes of considered eigenvalues (\citet{shuman2013emerging}). Proposed in previous studies on pursuing better running time of the filtering, $\hat{h}(\mathcal{L})$ can be computed using polynomial approximation techniques (e.g. the Chebyshev polynomials of the first kind after normalizing the eigenvalues to $[-1, 1]$ (\citet{hammond2011wavelets})) that bypass the expensive eigendecomposition (\citet{defferrard2016convolutional}; \citet{balcilar2020bridging}). And the filtered signal thereafter is $\hat{\tilde{s}} = \hat{h}(\mathcal{L}) \cdot \tilde{s}$. Although the difference between $\tilde{s}$ and $\hat{\tilde{s}}$ does partially reveal the ``frequency" ingredients of $\tilde{s}$, unfortunately this approach does not treat $\tilde{s}$ atomically, and choosing $\hat{h}$ can be an art in itself. Thus, it is not a satisfactory alternative.
\par
The primary quest of this paper is to seek more reliable and generic methods computing the magnitudes of eigencomponents (i.e. the spectrum) carried by vector-valued graph signals. Several attempts are explored. The methods are motivated from various perspectives such as reducing vector-valued signals to real-valued signals by approximation techniques, utilizing Dirichlet forms between the gradients of signal and the gradients of eigenvectors, and transforming eigenbases between the graph and its line graph. Furthermore, motivated by the fact that the spectra produced by the GFT reflect the relations between adjacent nodes regardless of nodes beyond adjacency (\citet{shuman2013emerging}), a series of auxiliary graphs are proposed, named the \textbf{Stratified Graphs (SGs)}, each induced by a K-hop non-backtracking neighborhood (e.g. the 0-th auxiliary graph is the original). Extracting spectra from the SGs thus helps gain finer resolutions and better insight in profiling the spectral characteristics of signals. The aforementioned methods are all extended to the SGs, and proposed as new tools for diagnosing and profiling graph learning models with both real- and vector-valued signals. They are named the \textbf{stratified graph spectra (SGS)} methods. 
\par
Note that the time complexity usually is not a top consideration in diagnosing and profiling models. In practice, usually thousands or hundreds or even fewer nodes are fair in understanding the behaviors of a model. At this scale, a dominantly expensive step in all SGS methods, the eigendecomposition, can be run in a reasonable time. 
\par
This paper is structured as follows. The motivations, algorithms and limitations of the SGs and the SGS methods are discussed in Section \ref{sec:sgs}. The empirical effectiveness and the utility of the SGS methods are demonstrated with experiments in Section \ref{sec:experiments}. A few related work is reviewed in Section \ref{sec:related_work}. 
\par
\bigbreak
\par
\noindent\textbf{Summary of Contributions}
\par
$\bullet$ The SGs, a series of auxiliary graphs induced from $K$-hop non-backtracking neighborhoods, are proposed as a carrier for the SGS methods helping gain finer resolutions and better insight in understanding the spectral characteristics of both real- and vector-valued graph signals.
\par
$\bullet$ Five SGS methods computing the spectra of signals are proposed.

\section{Stratified Graph Spectra Methods} \label{sec:sgs}

The SGs and SGS methods are primarily motivated by two problems. One is computing the spectra (i.e. magnitudes of eigencomponents) for vector-valued graph signals, and the other is extending the spectra from reflecting relations between $1$-hop neighboring nodes to $K$-hop neighboring nodes. 
\par
Regarding the first problem, five methods (\textbf{Algorithm 2}-\textbf{6}) are proposed. The development of these methods starts with solving a linear least square by approximation transforming the vector-valued signal to a real-valued signal (\textbf{Algorithm 2}). In order to gain more efficiency, a local gradient aggregation based method for the same objective (as in \textbf{Algorithm 2}) is proposed (\textbf{Algorithm 3}), though it is found that this method may only be effective on pulse-like signals. In spite of the limitation, inspired by the utilization of gradients, a simple method based on a Dirichlet form is proposed (\textbf{Algorithm 4}), and this method computes the magnitudes in the edge domain instead of the vertex domain. Following the idea keeping the magnitude computation in the edge domain, another method is proposed computing the GFT on edges and then converting it back to the vertex domain (\textbf{Algorithm 5}). Finally, an empirical ensemble method is discussed (\textbf{Algorithm 6}). 
\par
Regarding the second problem, the SGs are proposed. In brief, a $K$-SG of a given graph preserves all vertices, and links each node to every $K$-hop non-backtracking neighbor if any (e.g. the original graph is the $1$-SG). This is also what ``stratified" means. The maximum $K$ is determined by the diameter of the original graph. The SGs are motivated by the approach computing eigenvalues and eigenvectors of the graph Laplacian induced from the Courant-Fischer Theorem, where eigenvalues reflect the fluctuation of adjacent nodes' signal values which are determined by eigenvectors (Section 1.2 in \citet{chung1997spectral}). And the SGs are designed to capture different levels of adjacency. 
\par
For simplicity, all graphs considered in this paper are undirected, unweighted and self-loop-less.


\subsection{Stratified Graphs}
\begin{figure}
	\centering
	\includegraphics[width=1\textwidth]{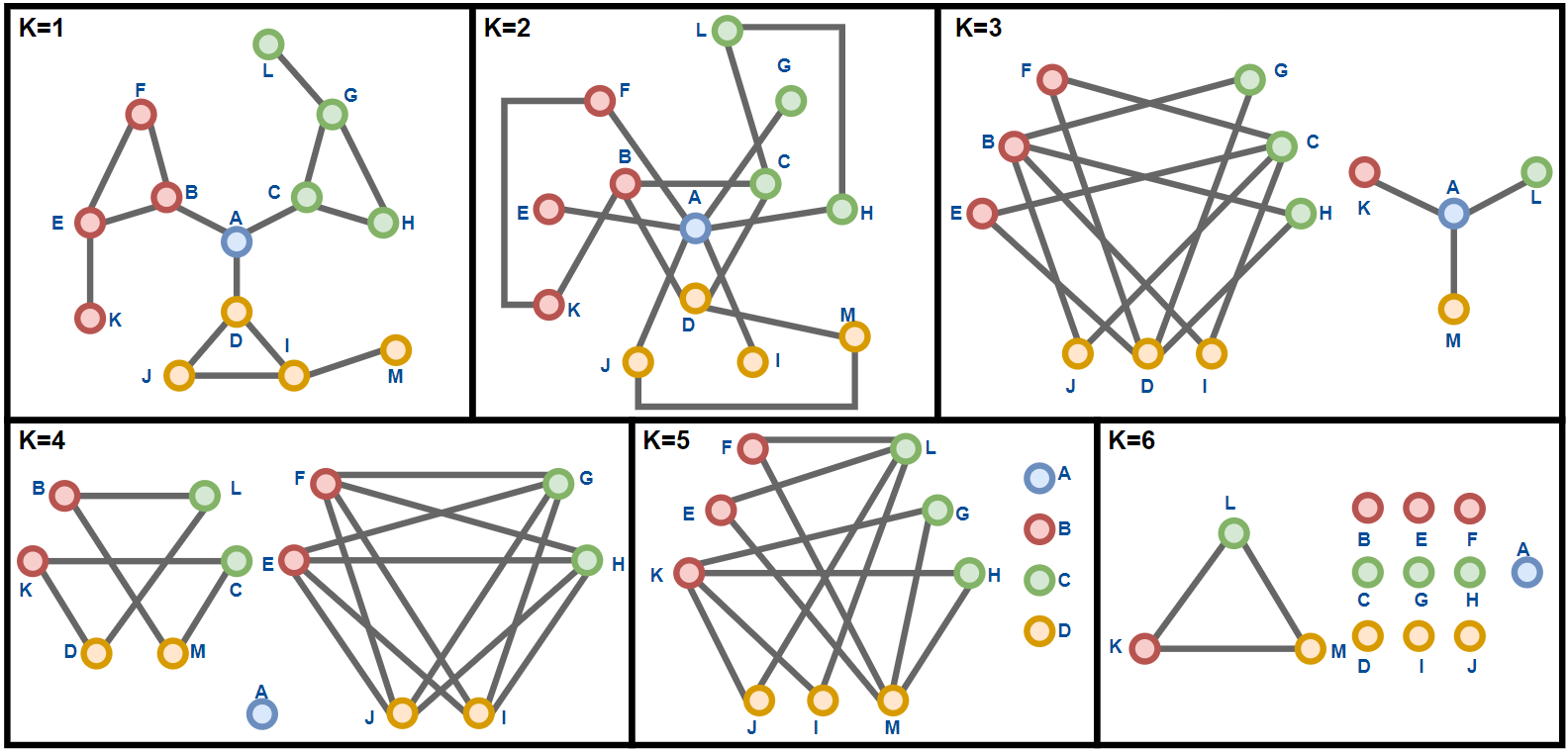}
	\caption{An example of SGs. The graph at $K=1$ is the original.}
	\label{fig:sg_sample}
\end{figure}

The concept \textbf{stratified graph (SG)} is formally defined in this section, and an example of SGs is illustrated in Figure \ref{fig:sg_sample}.
\par
\begin{framed}
	\noindent\textbf{Definition 1: Stratified Graphs (SGs) \& Line Stratified Graphs (LSGs)}
	\par
	Let $\mathcal{G} = (\mathcal{V}, \mathcal{E})$ be a connected graph, and let $\rho_{\mathcal{G}}$ be the graph diameter (i.e. the longest shortest path length). For each integer $1 \leq K \leq \rho_{\mathcal{G}}$, construct a new graph $\mathcal{G}_{K} = (\mathcal{V}_{K}, \mathcal{E}_{K})$ satisfying: (a) $\mathcal{V}_{K} = \mathcal{V}$, (b) $\mathcal{E}_{K} = \{ \forall e_{xy} \in \mathcal{V} \times \mathcal{V} | d_{\mathcal{G}}(x, y) = K \}$, where $d_{\mathcal{G}}$ denotes the shortest path length between $x$ and $y$ in $\mathcal{G}$. $\{\mathcal{G}_{K}\}$ are the \textbf{stratified graphs (SGs)} of $\mathcal{G}$. The SG at $K$ is denoted by $K$-SG.
	\par
	Each $\mathcal{G}_{K}$ can be converted to a line graph (\citet{biggs1993algebraic}; \citet{godsil2001algebraic}), denoted by $L(\mathcal{G}_{K})$ and named as \textbf{line stratified graph (LSG)}.
\end{framed}
\par
\par
The adjacency matrices of $\mathcal{G}_{K}$'s are necessary to the SGS methods, and they are computed by \textbf{Algorithm 1}. 
\par
\begin{framed}
	\noindent\textbf{Algorithm 1: Adjacency Matrices of SGs}\\
	$\triangleright$ \textbf{Given:}
	\par
	$\bullet$ A connected graph $\mathcal{G}$
	\par
	\noindent
	$\triangleright$ \textbf{Seek:}
	\par
	$\bullet$ The adjacency matrices $\{\mathcal{A}_K\}$ of $\{\mathcal{G}_K\}$.
	\par
	\noindent
	$\triangleright$ \textbf{Steps:}
	\par
	(1) Compute $\rho_{\mathcal{G}}$. \par
	(2) For $K = 1$, $\mathcal{A}_1 = \mathcal{A}$, where $\mathcal{A}$ is the adjacency matrix of $\mathcal{G}$. \par
	(3) For $1 < K \leq \rho_{\mathcal{G}}$, $\mathcal{A}_K = \Phi_0\Big( \delta(\mathcal{A}^{K}) \vec{-} \sum\limits_{i=1}^{K-1} \mathcal{A}_i \Big)$, where $\mathcal{A}^{K}$ is the $K$ power of $\mathcal{A}$, $\delta$ is the function setting all non-zero elements to $1$, and $\Phi_0$ sets the diagonal to zeros, and $\vec{-}$ is a non-negative subtraction operator defined as follows: 
	\[x \vec{-} y = 
	\begin{cases}
		x - y	& \quad \text{if } x \geq y \\
		0		& \quad \text{otherwise}
	\end{cases}
	\] 
\end{framed}
\par
Note that the time complexity of the step (3) for each $K$ is dominated by the matrix power, and the loop is determined by $\rho_{\mathcal{G}}$. Computing $\rho_{\mathcal{G}}$ at the step (1) costs $\mathcal{O}(|\mathcal{E}| + |\mathcal{V}|)$ for unweighted graphs by the breadth-first search. The range of $\rho_{\mathcal{G}}$, though beyond the scope of this paper, depends on many intrinsic characteristics of graphs (\citet{chung2001diameter}; \citet{bollobas1981diameter}; \citet{bollobas2004diameter}). Empirically, observed from the experiments in Section \ref{sec:sgs_vs_gft}, a 50-node random graph generated from either the Erdős–Rényi model (ERM) or the stochastic block model (SBM) with various settings can have $\rho_{\mathcal{G}} \leq 7$. 
\par
The SGs are the carriers on which the SGS methods are performed. The five SGS methods are discussed in the following sections. As all of the SGS methods share the same inputs and seek the same targets, for convenience, the inputs and outputs are specified here:
\begin{framed}
	\noindent\textbf{Inputs \& and Outputs of SGS Methods}\\
	$\triangleright$ \textbf{Given:}
	\par
	$\bullet$ A connected graph $\mathcal{G} = (\mathcal{V}, \mathcal{E})$ of $N$ nodes.
	\par
	$\bullet$ A normalized vector-valued signal $s$ on $\mathcal{G}$.
	\par
	\noindent
	$\triangleright$ \textbf{Seek:}
	\par
	$\bullet$ The magnitude of each eigencomponent of $\mathcal{G}_K$ at each $K$ that $s$ carries.
\end{framed}
\par

\subsection{Linear Approximation Based Transform}
In classic GSP, the magnitudes of eigencomponents carried by a real-valued graph signal can be computed straightforwardly by the GFT (\citet{shuman2013emerging}). Naturally, when handling a vector-valued graph signal, an immediate thought is to find an isometric transform converting the vector-valued signal to a real-valued signal. However, as such a transform does not always exist and may not be unique when existing (\textbf{Fact 1}), approximate solutions are needed. 
\par
\begin{framed}
	\noindent \textbf{Fact 1:}
	\par
	For a simple graph $\mathcal{G} = (\mathcal{V}, \mathcal{E})$, a normalized node embedding $s: \mathcal{V} \rightarrow \mathbb{S}^{M}$, where $\mathbb{S}^{M}$ denotes the $M$-sphere, and a desired real-valued function $f: \mathcal{V} \rightarrow \mathbb{R} \text{ s.t. } |f| < \infty$, each edge $(x, y) \in \mathcal{E}$ establishes a non-linear function: $|f(x) - f(y)| = d(s(x), s(y))$, where $d$ is a distance in $\mathbb{S}^{M}$. As directly computing absolute values is not computational friendly, a typical construction is a system of quadratic equations of the form $[f(x) - f(y)]^2 - d^2(s(x), s(y)) = 0, \forall (x, y) \in \mathcal{E}$. This system can be consistent or inconsistent, which can be determined by checking if the corresponding Gröbner basis can be reduced to 1 (i.e. as the Gröbner basis can be thought of as a simplification of the system, the basis being reduced to $1$ implies that the system is equivalent to $\{0 = 1\}$ which obviously is inconsistent.) (\citet{adams1994introduction} Chapter 2; \citet{froberg1997introduction} Chapter 6). Solving non-linear polynomial systems is typically an expensive and complicated task \protect\footnotemark. 
	\par
	Instead, an approximate yet easier linear system can be constructed. First, each edge is oriented by an uniformly random choice. Second, for each oriented edge $x \rightarrow y$, a linear equation is created: $f(x) - f(y) = d(s(x), s(y))$. And thus the linear system is constructed as follows: 
	\begin{equation}\label{eq:vec_to_real_ls}
		\mathcal{B} \boldsymbol{f} = D
	\end{equation}, where $\mathcal{B}$ denotes the incidence matrix upon the oriented edges, $\boldsymbol{f}$ denotes the vector form of $f$ and $D=\{d(s(x), s(y))\}|_{\forall x \rightarrow y}$. It is clear that only when $rank(\mathcal{B}) = rank(\mathcal{B}|D) \leq |\mathcal{V}|$, a solution to $\boldsymbol{f}$ exists. To cover both consistent and inconsistent cases, an approximation method is needed. 
\end{framed}
\footnotetext{Popular methods solving polynomial systems include Gröbner-basis-based, homotopy-continuation-based, resultant-based and other methods. This topic is beyond this paper, and readers who are particularly interested are referred to \citet{sturmfels2002solving}, \citet{adams1994introduction}, \citet{verschelde1999algorithm}, \citet{manocha1994solving} and \citet{bates2013numerically}.}
\par

For underdetermined cases of the system specified by \textbf{Equation} \ref{eq:vec_to_real_ls}, the output solution is arbitrarily selected. On the other hand, for the overdetermined cases, empirically a least square approximation can be found. Then solving the linear system is reduced to solving a linear least square (LLS) problem: 
\begin{equation}
	\min\limits_{\boldsymbol{\hat{f}}} || \mathcal{B} \boldsymbol{\hat{f}} - D ||_2
\end{equation}
Many techniques can be utilized to address this problem (\citet{friedman2001elements}; \citet{james2013introduction};  \citet{strang2019linear}; \citet{trefethen1997numerical}). Particularly, in the Appendix A of \citet{markovsky2012low}, a brief summary of classic approaches solving overdetermined systems is provided. Among these techniques, the singular value decomposition (SVD) based approaches are one of the most popular choices to compute least squares approximation (see Section 7.4 in \citet{poole2014linear}). In the proposed \textbf{Algorithm 2}, an SVD-based approach\footnote{LAPACK (\citet{anderson1999lapack}) provides the implementation of the SVD-based methods.} is utilized to seek an approximate real-valued graph signal for a given vector-valued signal preserving the distances between nodes to a maximal extent, and then the GFT of the approximated signal is finally computed. Before the algorithm is detailed, a couple of necessary concepts are provided in \textbf{Definition 2}.
\par
\begin{framed}
	\noindent\textbf{Definition 2: Gradient and Divergence on Graphs}
	\par
	Generalizing the gradient on graphs proposed in \citet{lim2020hodge}, the gradient $\nabla s$ of a normalized vector-valued graph signal $s$ with respect to two given nodes $u, v \in \mathcal{V}$ is defined as 
	\begin{equation}\label{eq:graph_gradient}
		(\nabla s)(u, v) := \sqrt{\frac{1 - \cos\big(\theta(u, v)\big)}{2}}
	\end{equation}
	where $\theta(u, v)$ denotes the angle between $s(u)$ and $s(v)$. Hence, $(\nabla s)(\cdot, \cdot)$ is actually the Euclidean distance normalized to $[0, 1]$.
	\par
	The divergence $\Delta s$ with respect to $v \in \mathcal{V}$ is defined as 
	\begin{equation}\label{eq:graph_divergence}
		(\Delta s)(v) := \sum\limits_{\forall u \sim v} (\nabla s)(u, v)
	\end{equation}
	where $u \sim v$ denotes the adjacency between $u$ and $v$. 
\end{framed}
\par
\begin{framed}
	\noindent\textbf{Algorithm 2: Linear Approximation Based Transform on SGs (APPRX-LS)}
	\\
	(1) Compute incidence matrices $\{\mathcal{B}_K\}$ of  $\{\mathcal{G}_K\}$. \\
	(2) For each $K$ and for each $(x, y) \in \mathcal{E}$, compute $(\nabla s)(x, y)$, and $\nabla s_K$ denotes the vector of all $(\nabla s)(x, y)$ at $K$ in a predetermined edge order. \\
	(3) For each $K$, solve by approximation
	\begin{equation}
		\mathcal{B}_K \cdot \hat{f}_K = \nabla s_K
	\end{equation}
	where $\hat{f}_K$ is the desired real-valued signal.
	\\
	(4) Compute $\{\mathcal{A}_K\}$ of $\{\mathcal{G}_K\}$ by \textbf{Algorithm 1}.
	\\
	(5) Compute graph Laplacians $\{\mathcal{L}_K\}$ from  $\{\mathcal{A}_K\}$.
	\\
	(6) For each $\mathcal{L}_K$, compute the eigendecomposition 
	\begin{equation}
		\mathcal{L}_K = \mathcal{U}_K \cdot \Lambda_K \cdot \mathcal{U}_K^*
	\end{equation}
	where $u^K_i \in \mathcal{U}_K$ is the $i^{th}$ eigenvector associated to the $i^{th}$ eigenvalue $\lambda^K_i \in \Lambda_K$ when $\Lambda_K$ is sorted as $0 = \lambda^K_0 \leq \lambda^K_1 \leq \dots \leq \lambda^K_{N-1}$.
	\\
	(7) For each $\hat{f}$ and $\mathcal{U}_K$, compute the GFT by 
	\begin{equation}
		\hat\eta^K_{s} = \mathcal{U}^*_K \cdot\hat{f}_K
	\end{equation}
	and compute the magnitudes $\mathcal{M}^K_{APPRX-LS}$ of eigencomponents carried by $s$ by 
	\begin{equation}
		\mathcal{M}^K_{APPRX-LS}(i) = \big| \hat\eta^K_{s}(i) \big|
	\end{equation}
\end{framed}
\par
Note that the time complexity of \textbf{Algorithm 2} is lower than that of eigendecompostion which costs $\mathcal{O}(n^3)$ or slightly lower. The dominant step is computing the SVD which typically costs $\mathcal{O}(mn^2)$ (depending on what is requested for the outputs, e.g. both orthonormal bases and the singulars) where $(m, n)$ ($m \geq n$) is the size of $\mathcal{B}_K$ (\citet{golub2013matrix} Section 8.6). Noticing that computing the SVD is somewhat expensive in practice, \textbf{Algorithm 3} explores a simplified method without this step, though its utility is limited. 
\par

\subsection{Incidence Aggregation Based Transform}
The idea of \textbf{Algorithm 3} is also approximating the input vector-valued signal by a real-valued signal. The real-valued signal at each node is computed by a local aggregation. Recall that the GFT is a linear operator (\citet{shuman2013emerging}), and thus it holds that 
\begin{equation}
	\sum\limits_i \mathcal{U}^* \cdot s_i = \mathcal{U}^* \cdot \sum\limits_i s_i
\end{equation}
where $\mathcal{U}$ is the eigenvector matrix and $s_i$'s are real-valued signals. The idea is constructing a pulse signal at each node, and then combining them together. The amplitude of each pulse is determined by an aggregation of the distances between the center node and its 1-hop neighbors. This construction is inspired by the linear approximation of function (e.g. by the Taylor' theorem for the case $k=1$, for a function $f \in C^1$, $f$ can be approximated at a given point $a \in \mathbb{R}$ by $f(x) = f(a) + f'(a)(x-a) + h_1(x)(x-a)$). Specifically, the value at a given node can be approximated by the weighted divergence of gradients (\citet{lim2020hodge}) at this node. The GFT is then applied to the approximated real-valued signal. This method is detailed in \textbf{Algorithm 3}.
\par
\begin{framed}
	\noindent\textbf{Algorithm 3: Incidence Aggregation Based Transform on SGs (IN-AGG)}
	\\
	(1) Compute incidence matrices $\{\mathcal{B}_K\}$ of  $\{\mathcal{G}_K\}$.
	\\
	(2) Compute $\nabla s_K$.
	\\
	(3) For each $\nabla s_K$, compute $\Delta s_K$ for all nodes by 
	\begin{equation}\label{eq:alg2_step3}
		\Delta s_K = \mathcal{B}_K \cdot \nabla s_K
	\end{equation}
	where $\Delta s_K$ denotes the vector of all $\Delta s_K(v), \forall v \in \mathcal{V}$. 
	\\
	(4) Compute the desired approximated real-valued signal by 
	\begin{equation}\label{eq:in_agg_f_hat}
		\hat{f}_K(v) = \mathbb{E}\big[ (\nabla s) (v, u) \big] \Big|_{\forall u \sim v} = \frac{\Delta s_K (v)}{|\mathcal{N}(v, 1)|} 
	\end{equation}
	where $\mathbb{E}[\cdot]$ denotes expectation, and $\mathcal{N}(v, 1)$ denotes the set of 1-hop neighbors of $v$.
	\\
	(5) Compute the magnitudes of eigencomponents by 
	\begin{equation}
	\mathcal{M}^K_{IN-AGG}(i) = \big| (\mathcal{U}^* \cdot \hat{f}_K)(i) \big|
	\end{equation}
	where $\hat{f}_K$ denotes the vector of $\hat{f}_K(v)$ for all $v$.
\end{framed}
\par
Undoubtedly, the effectiveness of IN-AGG is conditioned, and it is discussed in \textbf{Proposition 1}. 
\par
\begin{framed}
	\noindent\textbf{Proposition 1:}
	\par
	Given a graph, the degrees of nodes are large enough. Then $\forall x \sim y, |\hat{f}(x) - \hat{f}(y)|$ trends to be monotonic to $(\nabla s)(x, y)$, only if $s$ is a constant signal or a linear combination of pulse-like signals (i.e. each of which is centered at a node $v$ such that $\forall a \neq v, b \neq v \in \mathcal{V}, (\nabla s)(v, a) \gg (\nabla s)(a, b)$) which are at least 2-hop distant from each other.
\end{framed}
\par
\noindent\textit{Proof:}
\par
If $s$ is a constant signal, then the statement is clearly true. The second case is as follows.
\par
W.l.o.g., two nodes $x, y$ are selected s.t. $x \sim y$, and it is assumed that $\hat{f}(x) \geq \hat{f}(y)$. Then it suffices to show that $\hat{f}(x) - \hat{f}(y)$ trends to be monotonic to $(\nabla s)(x, y)$ when at most two pulses exists in $\mathcal{N}(x, 1) \cup \mathcal{N}(y, 1) \cup \{x, y\}$ at: (i) $x$ and/or $\exists_1 b \neq x \in \mathcal{N}(y, 1)$, or (ii) $\exists_1 a \in \mathcal{N}(x, 1) \backslash \{x, y\}$ and/or $\exists_1 b \in \mathcal{N}(y, 1) \backslash \{x, y\}$, where $\exists_1$ means existing only one.
\par 
By \textbf{Equation} \ref{eq:in_agg_f_hat},
\begin{equation}\label{eq:in_agg_f_hat_low_bd}
	\begin{split}
		\hat{f}(x) - \hat{f}(y) 
		&= \mathbb{E}\big[ (\nabla s) (x, a)  \big]\Big|_{\forall a \sim x} - \mathbb{E}\big[ (\nabla s) (y, b) \big]\Big|_{\forall b \sim y} \\
		&= \mathbb{E}\big[ (\nabla s) (x, a) - (\nabla s) (y, b) \big]\Big|_{\forall a, b \in \mathcal{N}(x, 1) \times \mathcal{N}(y, 1)} \\
		&= \mathbb{E}\big[ [(\nabla s) (x, a) + (\nabla s) (y, a)] - [(\nabla s) (y, b) + (\nabla s) (y, a)] \big] \\
		&\geq \mathbb{E}\big[ (\nabla s) (x, y) - (\nabla s) (y, b) - (\nabla s) (y, a) \big] \\
		&= (\nabla s) (x, y) - \mathbb{E}\big[(\nabla s) (y, b)\big] - \mathbb{E}\big(\nabla s) (y, a) \big]
	\end{split}
\end{equation}
Similarly, 
\begin{equation}\label{eq:in_agg_f_hat_up_bd}
	\begin{split}
		\hat{f}(x) - \hat{f}(y) 
		&= \mathbb{E}\big[ [(\nabla s) (x, a) - (\nabla s) (y, a)] - [(\nabla s) (y, b) - (\nabla s) (y, a)] \big] \\
		&\leq (\nabla s) (x, y) - \mathbb{E}\big[ (\nabla s) (y, b)\big] + \mathbb{E}\big[(\nabla s) (y, a) \big] 
	\end{split}
\end{equation}
It is straightforward that, 
\begin{equation}
	|\hat{f}(x) - \hat{f}(y)| \rightarrow (\nabla s) (x, y) - \mathbb{E}\big[ (\nabla s) (y, b)\big], \text{ as } \mathbb{E}\big[(\nabla s) (y, a)\big] \rightarrow 0
\end{equation}
Then $|\hat{f}(x) - \hat{f}(y)|$ trends to be monotonic to $(\nabla s) (x, y)$ only when $\mathbb{E}\big[ (\nabla s) (y, b)\big] \rightarrow \mathcal{C}_{b}$ where $\mathcal{C}_{b} \geq 0$ is a constant. It needs to show that the two conditions $\mathbb{E}\big[(\nabla s) (y, a)\big] \rightarrow 0$ and $\mathbb{E}\big[ (\nabla s) (y, b)\big] \rightarrow \mathcal{C}_{b}$ are satisfied in the aforementioned two cases. 
\par
For the case (i), $\mathbb{E}\big[(\nabla s) (y, a)\big] = 0$. And, if $\exists_1 b_\delta \in \mathcal{N}(y, 1)$ carries a pulse, $\mathbb{E}\big[ (\nabla s) (y, b)\big] = \frac{(\nabla s)(y, b_\delta)}{|\mathcal{N}(y, 1)|} =  \frac{1}{|\mathcal{N}(y, 1)|}$, and when $\deg(y)$ is large $\mathbb{E}\big[ (\nabla s) (y, b)\big]$ is close to $0$ (i.e. $\mathcal{C}_{b} > 0$). Otherwise, $\mathbb{E}\big[ (\nabla s) (y, b)\big] = \mathcal{C}_{b} = 0$.
\par
For the case (ii), if $\exists_1 a_\delta \in \mathcal{N}(x, 1)$ carries a pulse, $\mathbb{E}\big[(\nabla s) (y, a)\big] = \frac{1}{|\mathcal{N}(x, 1)|}$, and when $\deg(x)$ is large $\mathbb{E}\big[(\nabla s) (y, a)\big]]$ is close to $0$. Otherwise, $\mathbb{E}\big[(\nabla s) (y, a)\big] = 0$. And $\mathbb{E}\big[ (\nabla s) (y, b)\big]$ has the same situation as case (i). $\Box$
\par
In addition to $\mathbb{E}\big[ (\nabla s) (y, a)\big]$ and $\mathbb{E}\big[ (\nabla s) (y, b)\big]$, it is also necessary to understand the asymptotics of $\mathbb{E}\big[ (\nabla s) (x, a)\big]$ appearing in both \textbf{Equations} \ref{eq:in_agg_f_hat_low_bd} and \ref{eq:in_agg_f_hat_up_bd}. $\mathbb{E}\big[(\nabla s) (y, a)\big] \rightarrow 0$ implies that $\forall a_i, a_j \in \mathcal{N}(x, 1), \mathbb{E}\big[(\nabla s) (a_i, a_j)\big] \rightarrow 0$ by the triangle inequality. Also, $0 \leq \mathbb{E}\big[(\nabla s) (x, a_i) - (\nabla s) (x, a_j)\big] \leq \mathbb{E}\big[(\nabla s) (a_i, a_j)\big]$, w.l.o.g. assuming that $(\nabla s) (x, a_i) \geq (\nabla s) (x, a_j)$, which further implies that $(\nabla s) (x, a_i)$ and $(\nabla s) (x, a_j)$ trends to be close as $\mathbb{E}\big[(\nabla s) (a_i, a_j)\big] \rightarrow 0$. Hence, $(\nabla s) (x, a) \rightarrow \mathcal{C}_a \geq 0$ as $\mathbb{E}\big[(\nabla s) (a_i, a_j)\big] \rightarrow 0$. And, when $x$ carries a pulse, $\mathcal{C}_a > 0$ is equal to the amplitude of the pulse, otherwise, $\mathcal{C}_a = 0$. This result necessarily confirms \textbf{Proposition 1}.
\par
\textbf{Proposition 1} and the analysis unveil the primary limitation of \textbf{IN-AGG}. When the two conditions, $\mathbb{E}\big[(\nabla s) (y, a)\big] \rightarrow 0$ and $\mathbb{E}\big[ (\nabla s) (y, b)\big] \rightarrow \mathcal{C}_{b}$, are relaxed, both the upper and lower bounds of $|\hat{f}(x) - \hat{f}(y)|$ can be arbitrary. Hence, \textbf{IN-AGG} may not be effective in the cases beyond \textbf{Proposition 1}, and in the experiments (Section \ref{sec:sgs_vs_gft}) this limitation is empirically discussed. 
\par
\begin{figure}
	\centering
	\includegraphics[width=0.5\textwidth]{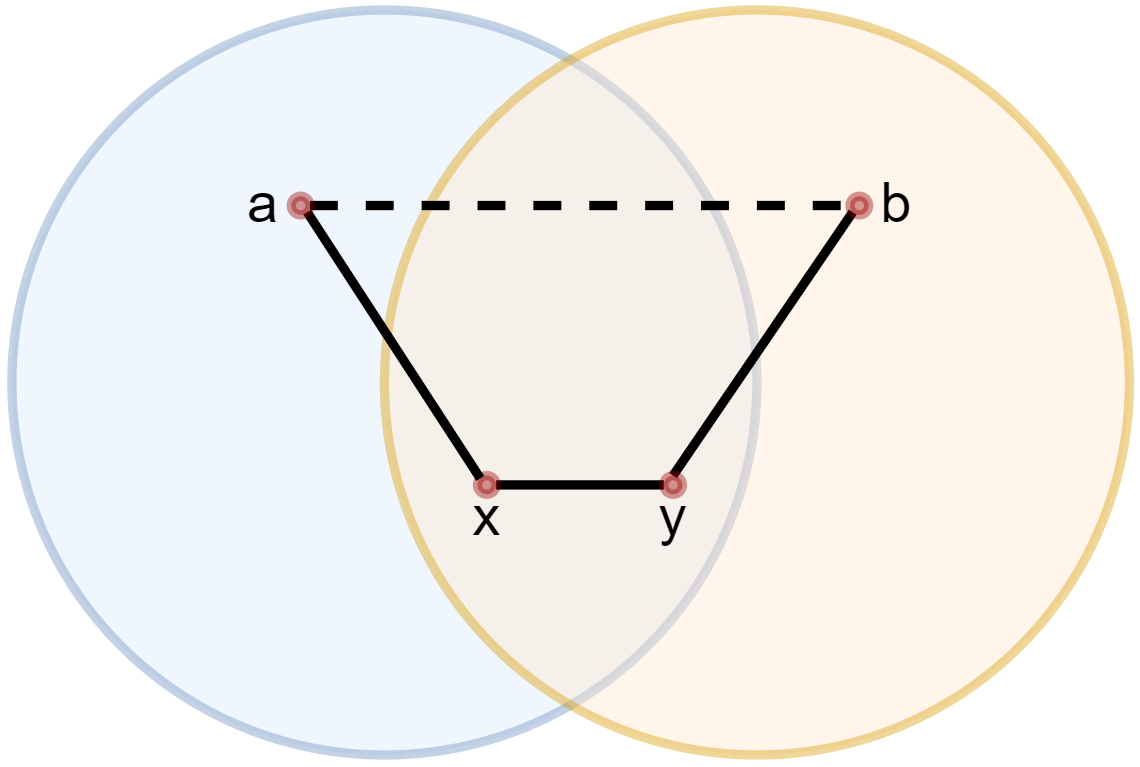}
	\caption{\textbf{Proposition 1.} The illustration of the relations between $x, y, a, b$, $\mathcal{N}(x, 1)$, and $\mathcal{N}(y, 1)$. The blue disc shows $\mathcal{N}(x, 1)$, and the yellow disc shows $\mathcal{N}(y, 1)$. $a$ (resp. $b$) exclusively belongs to $\mathcal{N}(x, 1)$ (resp. $\mathcal{N}(y, 1)$). There may or may not be a path (in an arbitrary length) between $a$ and $b$.}
	\label{fig:in_agg_prop1}
\end{figure}
\par

\subsection{Adjacent Difference Based Transform}
Despite the limitation of \textbf{IN-AGG}, leveraging gradients to compute magnitudes of eigencomponents keeps inspiring. Particularly, Dirichlet forms are a famous family of studying ``how one function changes relative to the changes of another" (e.g. defining energy measures (\citet{fabes1993}; \citet{fukushima2010dirichlet}; \citet{taylor2011partial})), and the forms have become one of the primary tools in analyzing Markov processes (\citet{fukushima2010dirichlet}; \citet{oshima2013semi}). In addition, many results in continuous cases have been migrated to discrete graphs (\citet{haeseler2017analysis}; \citet{chung1997spectral}; \citet{diaconis1996logarithmic}; \citet{bobkov2006modified}). The discrete versions of Dirichlet form defined in these works essentially agree with each other except minor difference in technical details. Typically, an early version proposed in \citet{diaconis1996logarithmic} is specific to Markov processes
\begin{equation*}
	\mathscr{E}(f, g) = \langle \mathcal{L}_r f, g \rangle = \frac{1}{2} \sum\limits_{x, y}  (f(x) - f(y))(g(x) - g(y)) P(x, y) \pi(x)
\end{equation*}
where $f, g$ are functions defined on vertices, $\mathcal{L}_r$ is the random walk Laplacian, $P$ is the transition probability and $\pi$ is the stationary distribution; a generalized version is proposed in \citet{bobkov2006modified} (Example 3.3 therein)
\begin{equation*}
	\mathscr{E}(f, g) = \int \langle \nabla f(x), \nabla g(x) \rangle d\mu(x) = \sum\limits_x \sum\limits_{y \sim x} (f(x) - f(y))(g(x) - g(y))\mu(x)
\end{equation*}
where $f, g$ are functions defined on vertices, and $\mu$ is a probability measure defined on vertices; and in \citet{haeseler2017analysis} (Section 4 therein) the definition is further generalized by adding in self-loops
\begin{equation*}
	\mathscr{E}(f, g) = \frac{1}{2} \sum\limits_{x, y} j(x, y)(f(x) - f(y))(g(x) - g(y)) + \sum\limits_{x} k(x, x)f(x)g(x)
\end{equation*}
where $j$ is the jump weight satisfying $\sum j < \infty$ and $k$ is the killing weight valuating non-negatives on the diagonal and zeros elsewhere (i.e. weighting self-loops). 
\par
It is intuitive that the discrete Dirichlet forms can be possibly utilized to compare a vector-valued signal $s$ to an eigenvector $u_i$ avoiding computing $|s(x) - s(y)|$ for $x \sim y$. To concretely leverage the forms in computing magnitudes of eigencomponents, a couple of simplifications and modifications need to be made. First, since the graphs considered in this paper are unweighted and self-loop-less, no self-loop term is included in the form, and the measure on vertices is simply induced by degrees. Second, regarding the gradient of a pair of adjacent nodes $x \sim y$, since only $\nabla s$ is available for computation and it is always non-negative, the corresponding gradient for the eigenvector $u_i$ is defined as
\begin{equation}
	(\nabla u_i)(x, y) = \big|u_i(x) - u_i(y)\big|
\end{equation}
Taking the absolute value is reasonable because typically the relative quantitative difference between elements (\citet{hata2017localization}; \citet{shuman2013emerging}) and the signs alone (primarily in analyzing nodal domains) (\citet{band2007nodal}; \citet{berkolaiko2008lower}; \citet{davies2000discrete}; \citet{helffer2009nodal}) are more meaningful than the signed difference. These simplifications and modifications finally give a very simple form to be utilized
\begin{equation}\label{eq:dirichlet_form}
	\mathscr{E}(s, u_i) := \langle \nabla s, \nabla u_i \rangle = \sum\limits_{\forall x \sim y} (\nabla s)(x, y) (\nabla u_i)(x, y)
\end{equation}
On the other hand, recall that the Dirichlet form of the graph Laplacian computes the eigenvalues (i.e. $\lambda_i = u_i^* \cdot \mathcal{L} \cdot u_i = \sum\limits_{\forall x \sim y}(u_i(x) - u_i(y))^2 = \sum\limits_{\forall x \sim y} [(\nabla u_i)(x, y)]^2$ for an eigenvector $u_i$). Then, asymptotically, when $\nabla s$ gets close to $\nabla u_i$ up to a scale $\mathcal{C}_i > 0$, $\mathscr{E}(s, u_i)$ gets close to $\mathcal{C}^2_i \lambda_i$. This fact significantly supports the rationale of this method. The full algorithm based on the form in \textbf{Equation} \ref{eq:dirichlet_form} to compute the magnitudes of eigencomponents is described in \textbf{Algorithm 4}.
\par
\begin{framed}
	\noindent\textbf{Algorithm 4: Adjacent Difference Based Transform on SGs (ADJ-DIFF)}\\
	(1) For each $K$, compute the eigenvector matrix $\{\mathcal{U}_K\}$.
	\\
	(2) For each $\mathcal{U}_K$ and for each $u^K_i \in \mathcal{U}_K$, compute the vector of $(\nabla u^K_i)(x, y), \forall x \sim y$, denoted by $\nabla u^K_i$. The elements of $\nabla u^K_i$ are indexed by a predetermined order of edges. Additionally, for $u^K_i$ corresponding to $\lambda_i=0$, manually set $\nabla u^K_i$ to an all-one vector $[1, \dots, 1]$ because otherwise $\nabla u^K_i$ will be a zero vector and thus be unable to capture any non-trivial magnitude. 
	\\
	(3) For each $K$, compute the magnitudes of eigencomponents by 
	\begin{equation}\label{eq:adj_diff}
		\mathcal{M}^K_{ADJ-DIFF}(i) = \frac{\mathscr{E}(s, u^K_i)}{\lambda_i}
	\end{equation}
\end{framed}
\par
The computation of \textbf{ADJ-DIFF} is very straightforward and fast. Nevertheless, the interpretation of \textbf{ADJ-DIFF} is not yet completely clear. To fully understand what \textbf{ADJ-DIFF} actually expresses, the upper and lower bounds of $\mathcal{M}^K_{ADJ-DIFF}(i)$ are discussed in \textbf{Proposition 2}. It is explored in depth what factors are tightly related to $\mathcal{M}^K_{ADJ-DIFF}$, as well as the limitations of \textbf{ADJ-DIFF}. 
\par
\begin{framed}
	\noindent\textbf{Proposition 2:}
	\par
	Given a graph $\mathcal{G} = (\mathcal{V}, \mathcal{E})$ and a normalized vector-valued graph signal $s$, 
	\begin{equation}\label{eq:prop2}
		\frac{\Bigg[\alpha \sum\limits_{\forall x \in \mathcal{V}} u_i^4(x) d_x \log \frac{u_i^4(x)}{\sum\limits_{\forall y \in \mathcal{V}} u_i^4(y) \pi(y)}\Bigg]^{\frac{1}{2}}}{\mathcal{O}(\mathcal{V}) ||u_i||_{\infty} \lambda_i} \leq \mathcal{M}_{ADJ-DIFF}(i) \leq \frac{\big[\mathcal{C}_{\lambda_i}\lambda_i + \mathcal{C}_{\kappa}\kappa \big] ||\nabla s||^2_2 + 2 \sum\limits_j \lambda_j}{2\lambda_i}
	\end{equation}
	where $u_i$ is the $i^{th}$ eigenvector and the corresponding eigenvalue is $\lambda_i$, $\alpha$ is a constant, $d_x = \sum\limits_{\mathclap{\forall y \sim x}} \sin^2\big(\theta(x, y)\big)$, $\pi(y) = \frac{d_y}{Vol(\mathcal{G})}$, $Vol(\mathcal{G}) = \sum\limits_{\forall x \in \mathcal{V}} d_x$, $\mathcal{C}_{\lambda_i}$ and $\mathcal{C}_{\kappa}$ are constants, $\kappa$ is a graph curvature. 
\end{framed}
\par
Note that $\alpha$ is the log-Sobolev constant (\citet{gross1975logarithmic}; \citet{chung1997spectral} Chapter 12; \citet{bobkov2006modified}; \citet{diaconis1996logarithmic}). By Lemma 12.1 in \citet{chung1997spectral}, under specific boundary conditions, it may hold that $\alpha \leq \frac{\beta_1}{2}$, where $\beta_1$ is the $1^{th}$ eigenvalue of the Laplacian of $\mathcal{G}$ with each edge $(x, y) \in \mathcal{E}$ weighted by $\sin^2(x, y)$. $\kappa$ is defined in \citet{chung2017strong} (Section 3.2). 
\par
\noindent\textit{Proof:}
\par
This proof starts with the lower bound. Let 
\begin{equation}
	\varphi_+(\mathcal{B^T} \cdot u_i) := \big[|u_i(x) + u_i(y)|\big]|_{\forall x \sim y}
\end{equation}
where $[\cdot]|_\mathcal{H}$ denotes the vector induced by the index set $\mathcal{H}$. Let
\begin{equation}
	\Phi(A, B) := I \cdot A \cdot B = \big[ A(\eta) B(\eta) \big]|_{\forall \eta \in \mathcal{H}}
\end{equation}
where $A$ and $B$ are vectors indexed by $\mathcal{H}$, and $I$ is the identity matrix. Let
\begin{equation}
	\varphi_+(A) = \big[ |A(\eta)| \big]_{\forall \eta \in \mathcal{H}}
\end{equation}
Let
\begin{equation}
	(\Gamma s)(x, y) := \sqrt{\frac{1 + \cos \big(\theta(x, y)\big)}{2}}
\end{equation}
and let
\begin{equation}
	\Gamma s := \big[ (\Gamma s)(x, y) \big]|_{\forall x \sim y}
\end{equation}
Then, it is obtained that
\begin{equation}
	\begin{split}
		\big\langle \Phi(\nabla s, \varphi_+(\mathcal{B^T} \cdot u_i)), \Phi(\Gamma s, \nabla u_i) \big\rangle 
		&= \sum\limits_{\forall x \sim y} |u_i(x) + u_i(y)| (\nabla s)(x, t) |u_i(x) - u_i(y)| (\Gamma s)(x, y) \\
		&= \big\langle \Phi(\nabla s, \nabla u_i), \Phi(\Gamma s, \varphi_+(\mathcal{B} \cdot u_i)) \big\rangle
	\end{split}
\end{equation}
where $\mathcal{B}$ denotes the incidence matrix.
By the Cauchy–Schwarz inequality, 
\begin{equation}
	\Big| \big\langle \Phi(\nabla s, \varphi_+(\mathcal{B^T} \cdot u_i)), \Phi(\Gamma s, \nabla u_i) \big\rangle \Big| \leq ||\Phi(\nabla s, \nabla u_i)||_2 ||\Phi(\Gamma s, \varphi_+(\mathcal{B} \cdot u_i))||_2
\end{equation}
and straightforwardly,
\begin{equation}
	\Big| \big\langle \Phi(\nabla s, \varphi_+(\mathcal{B^T} \cdot u_i)), \Phi(\Gamma s, \nabla u_i) \big\rangle \Big|^2 \leq ||\Phi(\nabla s, \nabla u_i)||_2^2 ||\Phi(\Gamma s, \varphi_+(\mathcal{B} \cdot u_i))||_2^2
\end{equation}
For $(\nabla s)(x, y) \geq 0$ and $(\nabla u_i)(x, y) \geq 0$, then 
\begin{equation}
	||\Phi(\nabla s, \nabla u_i)||_2^2 = \sum\limits_{\forall x \sim y} (\nabla s)^2(x, u) (\nabla u_i)^2(x, y) \leq \mathscr{E}^2(s, u_i) = \mathcal{M}^2_{ADJ-DIFF}(i) \lambda_i^2
\end{equation}
and for $(\Gamma s)(x, y) \geq 0$ and $\varphi_+(\mathcal{B} \cdot u_i)(x, y) \geq 0$, then
\begin{equation}
	||\Phi(\Gamma s, \varphi_+(\mathcal{B} \cdot u_i))||_2^2 = \sum\limits_{\forall x \sim y} (\Gamma s)^2(x, y) |u_i(x) + u_i(y)|^2 \leq \Big[\sum\limits_{\forall x \sim y} (\Gamma s)(x, y) |u_i(x) + u_i(y)|\Big]^2
\end{equation}
Hence,
\begin{equation}
	\Big| \big\langle \Phi(\nabla s, \varphi_+(\mathcal{B^T} \cdot u_i)), \Phi(\Gamma s, \nabla u_i) \big\rangle \Big|^2 \leq \Big[ \mathcal{M}_{ADJ-DIFF}(i) \lambda_i \Big]^2 \Big[ \sum\limits_{\forall x \sim y} (\Gamma s)(x, y) |u_i(x) + u_i(y)| \Big]^2
\end{equation}
By the definitions of $(\nabla s)(\cdot, \cdot)$ and $(\Gamma s)(\cdot, \cdot)$,
\begin{equation}
	(\nabla s)(x, y) (\Gamma s)(x, y) = |\sin\big( \theta(x, y) \big)|
\end{equation}
It can be rewritten that
\begin{equation}
	\Big| \big\langle \Phi(\nabla s, \varphi_+(\mathcal{B^T} \cdot u_i)), \Phi(\Gamma s, \nabla u_i) \big\rangle \Big|^2 
	= \Big[ \sum\limits_{\forall x \sim y} |u_i^2(x) - u_i^2(y)| |\sin\big( \theta(x, y) \big)| \Big]^2
\end{equation}
By $|u_i^2(x) - u_i^2(y)| \geq 0$ and $|\sin\big( \theta(x, y) \big)| \geq 0$, 
\begin{equation}
	\Big| \big\langle \Phi(\nabla s, \varphi_+(\mathcal{B^T} \cdot u_i)), \Phi(\Gamma s, \nabla u_i) \big\rangle \Big|^2 \geq \sum\limits_{\forall x \sim y} \big[u_i^2(x) - u_i^2(y)\big]^2 \sin^2\big( \theta(x, y) \big)
\end{equation}
Hence, if it is assumed that not all adjacent nodes have signal vectors in the opposite directions, then
\begin{equation}
	\mathcal{M}^2_{ADJ-DIFF}(i) \geq \frac{\sum\limits_{\forall x \sim y} \big[u_i^2(x) - u_i^2(y)\big]^2 \sin^2\big( \theta(x, y) \big)}{\lambda_i^2 \Big[ \sum\limits_{\forall x \sim y} (\Gamma s)(x, y) |u_i(x) + u_i(y)| \Big]^2}
\end{equation}
Consider the graph $\mathcal{G}$ with each edge $(x, y) \in \mathcal{E}$ weighted by $\sin^2(x, y)$, and apply the log-Sobolev inequality for discrete graphs (\citet{diaconis1996logarithmic}; \citet{chung1997spectral} Chapter 12) on $u_i$. Then it is obtained that
\begin{equation}
	\sum\limits_{\forall x \sim y} \big[u_i^2(x) - u_i^2(y)\big]^2 \sin^2\big( \theta(x, y) \big) \geq \alpha \sum\limits_{\forall x \in \mathcal{V}} u_i^4(x) d_x \log \frac{u_i^4(x)}{\sum\limits_{\forall y \in \mathcal{V}} u_i^4(y) \pi(y)}
\end{equation}
where $\alpha$, $d_x$, $\pi(y)$ and $Vol(\mathcal{G})$ are the same concepts mentioned in the statement of \textbf{Proposition 2}. In additions, by $(\Gamma s)(x, y) \leq 1$ and $|u_i(x) + u_i(y)| \leq 2||u_i||_{\infty}$, 
\begin{equation}
	\Big[ \sum\limits_{\forall x \sim y} (\Gamma s)(x, y) |u_i(x) + u_i(y)| \Big]^2 \leq 4 |\mathcal{E}|^2 ||u_i||^2_{\infty} \leq 4 \mathcal{O}^2(\mathcal{V}) ||u_i||^2_{\infty}
\end{equation}
Hence, 
\begin{equation}
	\mathcal{M}_{ADJ-DIFF}(i) \geq \frac{\Bigg[ \alpha \sum\limits_{\forall x \in \mathcal{V}} u_i^4(x) d_x \log \frac{u_i^4(x)}{\sum\limits_{\forall y \in \mathcal{V}} u_i^4(y) \pi(y)} \Bigg]^{\frac{1}{2}}}{2  \mathcal{O}(\mathcal{V}) ||u_i||_{\infty} \lambda_i}
\end{equation}
\par
Next, the upper bound is discussed. By the definition of $\mathcal{M}_{ADJ-DIFF}$, 
\begin{equation}
	\begin{split}
		\mathcal{M}_{ADJ-DIFF}(i) \lambda_i &= \mathscr{E}(s, u_i) = \sum\limits_{\forall x \sim y} (\nabla s)(x, y) (\nabla u_i)(x, y) \\
		& \leq \sum\limits_{\forall x \sim y} \frac{1}{2} \Big[ (\nabla s)^2(x, y)(x, y) \big[ u_i(x) - u_i(y) \big]^2 + 1 \Big]
	\end{split}
\end{equation}
And, by the Harnack inequality for general graphs proposed in \citet{chung2017strong}, 
\begin{equation}
	\big[ u_i(x) - u_i(y) \big]^2 \leq \mathcal{C}_{\lambda_i} \lambda_i + \mathcal{C}_{\kappa} \kappa
\end{equation}
where $\mathcal{C}_{\lambda_i}$ and $\mathcal{C}_{\kappa}$ are determined by $d_{\max} = \max\limits_{\forall x \in \mathcal{V}}\{d_x\}$ up to constant scales. Hence,
\begin{equation}
	\begin{split}
		\mathcal{M}_{ADJ-DIFF}(i) 
		&\leq \frac{\big[\mathcal{C}_{\lambda_i} \lambda_i + \mathcal{C}_{\kappa} \kappa\big] \sum\limits_{\forall x \sim y} (\nabla s)^2(x, y) + |\mathcal{E}|}{2 \lambda_i} \\
		&= \frac{\big[\mathcal{C}_{\lambda_i} \lambda_i + \mathcal{C}_{\kappa} \kappa\big] ||\nabla s||^2_2 + 2 \sum\limits_j \lambda_j}{2 \lambda_i}
	\end{split}
\end{equation}
by the fact that $\sum\limits_j \lambda_j = \frac{|\mathcal{E}|}{2}$. $\Box$
\par
In \textbf{Proposition 2}, the lower bound is primarily determined by the numerator which is an entropy-like quantity multiplied by the log-Sobolev constant. This entropy-like quantity is determined by $|u_i(x)|$ and $d_x$. On one hand, when $x$ is dissimilar to its neighbors, $d_x$ trends to be great. On the other hand, the term $u_i^4(x) d_x \log \frac{u_i^4(x)}{\sum\limits_{\forall y \in \mathcal{V}} u_i^4(y) \pi(y)}$ being great requires the log term being positive and great. The log term can be rewritten as
\begin{equation}
	\log \frac{u_i^4(x)}{\sum\limits_{\forall y \in \mathcal{V}} u_i^4(y) \pi(y)} = \log \frac{u_i^4(x) d_x + \sum\limits_{\forall y \neq x} u_i^4(x) d_y}{u_i^4(x) d_x + \sum\limits_{\forall y \neq x} u_i^4(y) d_y}
\end{equation} 
The denominator is actually an inner product of $u_i^4$ and $\boldsymbol{d}$ (i.e. the vector of $d_y$), which essentially has a similar meaning as \textbf{IN-AGG}. It is clear that only when $|u_i(x)| > |u_i(y)|, \forall y \neq x$ this log term is positive, and, the larger the difference, the greater the log term. Thus, understanding which $|u_i(x)|$'s are great at a given $i$ is the key to answer when the log term is great. Plenty of studies have argued that in various cases the localization of eigenvectors exists (Laplacian based: \citet{grebenkov2013geometrical}; \citet{hata2017localization}; adjacency based: \citet{pastor2016distinct}; \citet{pastor2018eigenvector}), though it is still unknown if there exists a generic pattern of the localization. For example, according to the main results proposed in \citet{hata2017localization}, in large random networks, $|u_i(x)|$ is linearly correlated to $i$, and only the nodes sharing similar degrees take large absolute values in $u_i$ while others are very small. This may not hold in other types of graphs (e.g. the Fiedler vectors of graphs with strong partition structures) \footnote{The localization of eigenvectors is a topic beyond this paper.}. Based on the analysis above, it can be concluded that, for each $u_i$, there are a particular subset of nodes $\mathcal{V}_i$ at which $|u_i(x)|, \forall x \in \mathcal{V}_i$ are greater than others, and, when these nodes are dissimilar to their neighbors (i.e. $d_x$'s are great), the lower bound in \textbf{Proposition 2} is high. This conclusion unveils a crucial difference between \textbf{ADJ-DIFF} and the GFT. The latter one produces a high (absolute) value only when $|s(x)|, \forall x \in \mathcal{V}_i$ are great (when $s$ is a real-valued signal) but $x$'s may not be dissimilar to their neighbors.
\par
Another key factor in the lower bound is the log-Sobolev constant defined in \citet{chung1997spectral} (Equation 12.4) as follows. 
\begin{equation}
	\alpha := \inf\limits_{f \neq 0} \frac{\sum\limits_{\forall x \sim y} \big[f(x) - f(y)\big]^2 w_{x, y}}{\sum\limits_{\forall x \in \mathcal{V}} f^2(x) d_x \log \frac{f^2(x)}{\sum\limits_{\forall y \in \mathcal{V}} f^2(y) \pi(y)}}
\end{equation}
A potential upper bound for $\alpha$ is given in \citet{chung1997spectral} (i.e. $\alpha \leq \frac{\beta_1}{2}$), though therein it is also stated that this bound may not always hold (e.g. \citet{diaconis1991geometric}). Suppose it holds or $\alpha$ is within a constant factor of $\beta_1$ \footnote{The topic of the log-Sobolev constant in depth is beyond this paper. Up to this point, the bound for this constant is typically discussed case by case. Readers who are interested are referred to Chapter 9 and Chapter 12 in \citet{chung1997spectral}, \citet{diaconis1996logarithmic}, \citet{wang1999harnack} and \citet{jerrum2004elementary}.}. Then $\alpha$ is actually determined by $\nabla s$ and the partition structures of the graph. Specifically, when $\nabla s$ trends to agree the partition structures, $\beta_1$ is close to $\lambda_1$ (i.e. the Fiedler value of $\mathcal{G}$), otherwise, $\beta_1$ can be greater than $\lambda_1$ and keeps increasing. 
\par
\begin{figure}
	\centering
	\includegraphics[width=0.2\textwidth]{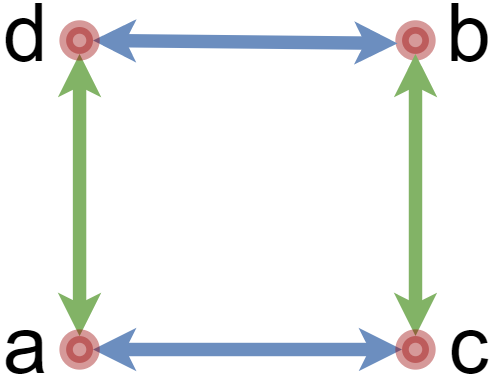}
	\captionsetup{singlelinecheck=off}
	\caption{\textbf{Proposition 2}. An example of homogeneous graph (defined in \citet{chung2017strong}). The adjacency in the graph follows a group structure. Specifically, the generating set of the group consists of four colored arrows: \{blue-right, blue-left, green-up, green-down\}. And, given a node (e.g. $a$), every other node can be expressed as this node being applied with a word of the group (e.g. $b = \text{green-up}(\text{blue-right}(a))$). For such graphs, it always holds that, for instance, 
	\begin{displaymath}
		\begin{split}
			&[f(a) - f(\text{blue-right}(a))] - [f(\text{green-up}(a)) - f(\text{blue-right}(\text{green-up}(a)))] \\ 
			&= [f(a) - f(\text{green-up}(a))] - [f(\text{blue-right}(a)) - f(\text{green-up}(\text{blue-right}(a)))]
		\end{split}	
	\end{displaymath}
	For homogeneous graphs, the curvature is always zero, and those graphs are called ``Ricci flat" (\citet{chung2017strong}).}
	\label{fig:curvature}
\end{figure}

On the other hand, the upper bound of $\mathcal{M}_{ADJ-DIFF}(i)$ is primarily determined by $\kappa$ and $||\nabla s||_2$. The curvature $\kappa$, roughly speaking, measures the extent to which, from a node $x$ to another node $y$, the cumulative distance between adjacent nodes varies along different paths linking $x$ and $y$. $\kappa$ is a characteristics of the graph, and is exclusively determined by all eigenvectors (\citet{chung2017strong}). In \citet{chung2017strong}, it is also justified that $\kappa$ is consistent with Ollivier's Ricci curvature proposed in \citet{ollivier2009ricci}. For homogeneous graphs (\citet{chung1994harnack}) (an example is demonstrated in Figure \ref{fig:curvature}), $\kappa = 0$. Otherwise, $\kappa$ can be positive or negative. Thus, $\kappa$ can be considered as a fixed constant (can be positive, negative or zero) reflecting an intrinsic geometric characteristics of the graph. Therefore, $||\nabla s||_2$ is the most variational factor in the upper bound, and it is scaled by $\kappa$ of the graph. 
\par
The analysis above has shed some light on what $\mathcal{M}_{ADJ-DIFF}$ is related to (e.g. the localization of eigenvectors and the curvature of graph) as well as how \textbf{ADJ-DIFF} is different from the GFT. Particularly, the difference between \textbf{ADJ-DIFF} and the GFT also implies a potential limitation of \textbf{ADJ-DIFF}. For example, for a graph with strong partition structures, the Fiedler vectors indicate the clusters, and the vector values at the nodes in the same cluster are likely to be very similar (or even identical). In this case, the magnitudes corresponding to the Fiedler vectors may mostly depend on the divergence of the nodes at the cluster boundary rather than the resemblance of the members. Another potential limitation is that the intrinsic characteristics of the input graph can significantly affect the scale of $\mathcal{M}_{ADJ-DIFF}$, which may result in numerical issues. Finally, $\nabla u_i$'s may not be orthogonal to each other, which may further lead to confusion between magnitudes of eigencomponents. 
\par
Regarding the implementation, the step (2) of \textbf{Algorithm 4} can be benefited by utilizing matrix operations to compute $\big|u^K_i(x) - u^K_i(y)\big|$ in relatively large cases in terms of the empirical running time. Specifically,
\begin{equation*}
	\big|u^K_i(x) - u^K_i(y)\big| = \sqrt{u^K_i(x)^2 + u^K_i(y)^2 - 2 u^K_i(x) u^K_i(y)}
\end{equation*}
Thus, firstly compute 
\begin{equation}
	\Xi^K_i = \sqrt{\Big[\big(u^K_i\big)^2\Big]^T_N + \Big[\big(u^K_i\big)^2\Big]_N - 2 u^K_i \cdot \big(u^K_i\big)^T} 
\end{equation}
where $[\cdot]_N$ denotes stacking $N$ copies of a column vector, then compute 
\begin{equation*}
	\Xi^K_i = \Xi^K_i \odot \delta(\mathcal{A}_K)
\end{equation*}
where $\odot$ denotes the element-wise matrix multiplication, and finally express $\Xi^K_i$ as a vector (i.e. $\nabla u_i^K$) following the order of edges ruled by $\mathcal{E}_K$ after removing duplicates. 
\par

\subsection{Line-to-Vertex Conversion Based Transform}
\textbf{Algorithm 2} and \textbf{3} compute the magnitudes of eigencomponents in the vertex domain while \textbf{Algorithm 4} actually has migrated this computation to the edge domain. It is well known that the line graph is a dual of the original graph (\citet{hemminger1983line}; \citet{godsil2001algebraic}), and many characteristics of the line graph are tightly related to those of the original graph (e.g. partitions of nodes and partitions of edges can be obtained by same methods, and can further be converted to each other (\citet{evans2009line})). Thus, another idea of computing magnitudes of eigencomponents is that: if there exists a transform between the eigenbasis of a graph and that of its line graph, then, by a design, the Fourier-transformed signals can be possibly transformed between the two domains directly or indirectly. Before discussing the implementation of this idea, it is necessary to understand how such a transform would look like if exists, and \textbf{Fact 2} starts this topic. 
\par
\begin{framed}
	\noindent\textbf{Fact 2: }
	\par
	The transform between the two Laplacian eigenbases, denoted by $\mathcal{U}$ and $L(\mathcal{U})$, of a graph $\mathcal{G} = (\mathcal{V}, \mathcal{E})$ and its line graph $L(\mathcal{G}) = (L(\mathcal{V}), L(\mathcal{E}))$, where $L(\mathcal{V}) = \mathcal{E}$, may not be linear (and thus may not exist a linear inverse). 
\end{framed}
\par
First, in general, $\mathcal{U}$ is a $|\mathcal{V}| \times |\Lambda|$ matrix, where $|\Lambda|$ denotes the number of eigencomponents in consideration, and similarly $L(\mathcal{U})$ is $|\mathcal{E}| \times |L(\Lambda)|$. Their ranks are $|\Lambda|$ and $|L(\Lambda)|$ respectively. Thus, unless $|\Lambda| = |L(\Lambda)|$ (or $|\mathcal{V}| =|\mathcal{E}|$, which in general does not hold), there is no linear transform between $\mathcal{U}$ and $L(\mathcal{U})$. Second, transforming between $\mathcal{U}$ and $L(\mathcal{U})$ requires a transpose of $\mathcal{U}$ or $L(\mathcal{U})$, if the two eigenbases are not truncated, and the transform exists. Typically, w.l.o.g., the transform can be written as 
\begin{equation}\label{eq:eigenbases_transform}
	\mathcal{U}^T = \mathcal{H}_{ln} \cdot L(\mathcal{U}) \cdot \mathcal{H}_{vx}^T
\end{equation}
where $\mathcal{H}_{ln}$ is $|\Lambda| \times |\mathcal{E}|$ and $\mathcal{H}_{vx}$ is $|\mathcal{V}| \times |L(\Lambda)|$. The transpose is not a linear transformation, and $\mathcal{H}_{ln}$ and $\mathcal{H}_{vx}$ are not necessarily invertible. Hence, this transform from $L(\mathcal{U})$ to $\mathcal{U}$ may not be linear and may not exist a linear inverse (and the same for the other direction).
\par
If the transform in \textbf{Equation} \ref{eq:eigenbases_transform} exists, \textbf{Fact 3} also holds. 
\begin{framed}
	\noindent\textbf{Fact 3: }
	\par
	For any $u_i \in \mathcal{U}$, $u_i$ in general cannot be localized to any particular subset of $\{L(u_j)\} \subseteq L(\mathcal{U})$ by the transform in \textbf{Equation} \ref{eq:eigenbases_transform} (i.e. $u_i$ can be dependent of every eigencomponents in $L(\mathcal{U})$). $\mathcal{H}_{ln}$ and $\mathcal{H}_{vx}$ do not transform Fourier-transformed signals or even Laplacians. 
\end{framed}
\noindent
Directly induced from \textbf{Equation} \ref{eq:eigenbases_transform}, it is obtained that
\begin{equation*}
	 L(\mathcal{U})^T = \big[\mathcal{H}_{vx}^T \cdot \mathcal{H}_{vx}\big]^{-1} \cdot \mathcal{H}_{vx}^T \cdot \mathcal{U} \cdot \mathcal{H}_{ln}^T \cdot \big[\mathcal{H}_{ln} \cdot \mathcal{H}_{ln}^T\big]^{-1}
\end{equation*}
assuming that $\big[\mathcal{H}_{vx}^T \cdot \mathcal{H}_{vx}\big]^{-1}$ and $\big[\mathcal{H}_{ln} \cdot \mathcal{H}_{ln}^T\big]^{-1}$ exist. Hence, the Fourier transform of $\nabla s$ in the line graph can be written as 
\begin{equation*}
	L(\mathcal{U})^T \cdot \nabla s = \big[\mathcal{H}_{vx}^T \cdot \mathcal{H}_{vx}\big]^{-1} \cdot \mathcal{H}_{vx}^T \cdot \mathcal{U} \cdot \mathcal{H}_{ln}^T \cdot \big[\mathcal{H}_{ln} \cdot \mathcal{H}_{ln}^T\big]^{-1} \cdot \nabla s
\end{equation*}
However, the RHS could hardly be rewritten in the form of the Fourier transform in the original graph (i.e. $\mathcal{U}^T \cdot \xi$, where $\xi$ is a transformed signal from the line graph). On the other hand, if $\mathcal{U}$ is substituted by \textbf{Equation} \ref{eq:eigenbases_transform}, then the eigendecomposition of $\mathcal{L}$ (i.e. the Laplacian of the original graph) is rewritten as
\begin{equation*}
	\mathcal{L} = \big[\mathcal{H}_{vx} \cdot L(\mathcal{U})^T\big] \cdot \big[\mathcal{H}_{ln}^T \cdot \Lambda \cdot \mathcal{H}_{ln}\big] \cdot \big[L(\mathcal{U}) \cdot \mathcal{H}_{vx}^T\big]
\end{equation*}
where $\Lambda$ denotes the diagonal matrix of the eigenvalues of $\mathcal{L}$. Clearly, this is a reduced-SVD-like decomposition but not a transform from the Laplacian of the line graph unless $L(\mathcal{U})$ is symmetric. 
\par
Despite these pessimistic facts, by \textbf{Equation} \ref{eq:eigenbases_transform}, it is clear that if the norms of $L(u_j) \in L(\mathcal{U})$ are kept unchanged (i.e. being $1$), then $\mathcal{U}$ can be recovered from $L(\mathcal{U})$. Then, weighting the $L(u_j)$'s by the Fourier-transformed $\nabla s$ in the line graph domain and transforming the weighted $L(u_j)$'s back to the original graph is the idea of \textbf{Algorithm 5} to approximate the eigencomponent magnitudes of the original graph carried by $s$.
\par
\begin{framed}
	\noindent\textbf{Algorithm 5: Line-to-Vertex Conversion Based Transform on SGs (LN-VX)}\\
	(1) Compute $\{\mathcal{A}_K\}$ of $\{\mathcal{G}_K\}$.
	\\
	(2) Compute incidence matrices $\{\mathcal{B}_K\}$ of  $\{\mathcal{G}_K\}$. 
	\\
	(3) For each line graph $L(\mathcal{G}_K)$ of $\mathcal{G}_K$, compute its adjacency matrices $L(\mathcal{A}_K)$ by
	\begin{equation}
		L(\mathcal{A}_K) = \mathcal{B}_K^T \cdot \mathcal{B}_K - 2\mathcal{I}
	\end{equation}
	where $\mathcal{I}$ is the identity matrix. 
	\\
	(4) Compute graph Laplacians $\{L(\mathcal{L}_K)\}$ for the line graphs from $\{L(\mathcal{A}_K)\}$.
	\\
	(5) For each $L(\mathcal{G}_K)$, compute the eigenbasis $L(\mathcal{U}_K)$.
	\\
	(6) For each $\mathcal{G}_K$, compute the eigenbasis $\mathcal{U}_K$.
	\\
	(7) For each pair of $\big( \mathcal{U}_K, L(\mathcal{U}_K) \big)$, following \textbf{Equation} \ref{eq:eigenbases_transform} construct a transform from $L(\mathcal{U}_K)$ to $\mathcal{U}_K$ by learning two matrices $\mathcal{H}_{ln}$ and $\mathcal{H}_{vx}$ on the objective 
	\begin{equation}
		\argmin\limits_{\mathcal{H}_{ln}, \mathcal{H}_{vx}} MSE \Big(\mathcal{U}^T_K, \mathcal{H}_{ln} \cdot L(\mathcal{U}_K) \cdot \mathcal{H}^T_{vx} \Big)
	\end{equation}
	where $MSE$ denotes the mean squared error. 
	\\
	(8) For each $K$, compute $\nabla s_K$.
	\\
	(9) For each $\nabla s_K$, compute its GFT in the line graph domain by 
	\begin{equation}
		\hat\eta^K_s = L(\mathcal{U}_K)^*_K \cdot \nabla s_K
	\end{equation}
	\\
	(10) For each $L(\mathcal{U}_K)$, compute the weighted eigenbasis $L^w(\mathcal{U}^K)$ by 
	\begin{equation}
		L^w(u_i^K) = |\hat\eta^K_s(i)| L(u_i^K)
	\end{equation}
	where $L(u_i^K) \in L(\mathcal{U}_K)$ is the $i^{th}$ eigenvector of $L(\mathcal{L}_K)$.
	\\
	(11) For each $L^w(\mathcal{U}^K)$, transform it to the vertex domain by 
	\begin{equation}\label{eq:weighted_eigenbasis_transform}
		\big( \mathcal{U}^w_K \big)^T = \mathcal{H}_{ln} \cdot L^w(\mathcal{U}_K) \cdot \mathcal{H}^T_{vx}
	\end{equation}
	\\
	(12) For each $\mathcal{U}^w_K$, and for each pair of $u^{wK}_i \in \mathcal{U}^w_K$ and $u^{K}_i \in \mathcal{U}_K$, compute the magnitudes $\mathcal{M}^K_{LN-VX}$ of eigencomponents carried by $s$ by 
	\begin{equation}
		\mathcal{M}^K_{LN-VX}(i) = \big| \langle u^{wK}_i, u^{K}_i \rangle \big|
	\end{equation}
\end{framed}
\par
The learning problem at the step (7) of \textbf{Algorithm 5} can be solved in various ways, and also this problem can further be generalized as $\mathcal{U}^T_K \approx \mathcal{F}\big(L(\mathcal{U}_K)\big)$, where $\mathcal{F}$ is learnable, and it may not be linear transform as argued in \textbf{Fact 2}. Additionally, when implementing this method, it is found that attaching a non-linear activation function (e.g. SeLU (\citet{klambauer2017self})) to $\mathcal{H}_{ln}$ and $\mathcal{H}_{vx}$ respectively can slightly benefit the performance. Nonetheless, the choice of activation function can significantly impact the performance as well (e.g. ReLU (\citet{nair2010rectified}), in the experiments, performed much worse than that even without activation function) \footnote{Exploring empirical constructions of the learning architecture solving the step (7) of \textbf{Algorithm 5} is not a primary concentration of this paper, and it is left to the future work.}. In Section \ref{sec:experiments} (\textbf{Task 1}), the quality of this learning is further discussed empirically. 
\par
\textbf{LN-VX} also has its own limitations. First, the transform defined in \textbf{Equation} \ref{eq:eigenbases_transform} actually needs to be conditioned on connected graphs over all $K$'s. Theoretically, the eigenbases of disconnected components are independent of each other, and they should be handled independently. However, if the connectivity condition is violated, the learning step has to take all connected components as a whole into consideration, which can easily worsen the hardness of the learning, and may introduce more confusion into the transformed eigenbasis. Second, by \textbf{Equation} \ref{eq:weighted_eigenbasis_transform}, 
\begin{equation}\label{eq:weighted_eigenbasis_transform_element}
	\mathcal{U}^w(j, i) = \sum\limits_{k=0}^{|L(\Lambda)| - 1} \Big|\big\langle \nabla s, L(u_k)^* \big\rangle\Big| \big\langle h_{{ln}_i}, L(u_k) \big\rangle h^T_{{vx}_j}(k)
\end{equation}
where $h_{{ln}_i} \in \mathcal{H}_{ln}$ and $h_{{vx}_j} \in \mathcal{H}_{vx}$ are row vectors. It is clear that this product is affected by the norms of $h_{{ln}_i}$'s and $h_{{vx}_j}$'s which may not be normalized. And, as $h_{{ln}_i}$'s and $h_{{vx}_j}$'s are resulted from the learning, their norms highly depend on the spectral characteristics of the input graph and how the learning is proceeded, which can be various case by case. Hence, $h_{{ln}_i}$'s and $h_{{vx}_j}$'s are likely to introduce numerical biases into the step (11), and further weaken the quality of resulting magnitudes of eigencomponents. Finally, the learned transform (i.e. $\mathcal{H}_{ln}$ and $\mathcal{H}_{vx}$) may not be unique. This can be observed by
\begin{equation*}
	\mathcal{H}_{ln} \cdot L(\mathcal{U})  = \mathcal{U}^T \cdot \mathcal{H}_{vx} \cdot \big( \mathcal{H}^T_{vx} \cdot \mathcal{H}_{vx} \big)^{-1}
\end{equation*}
where $\big( \mathcal{H}^T_{vx} \cdot \mathcal{H}_{vx} \big)^{-1}$ is assumed existing. As $L(\mathcal{U})$ is a basis, then $\mathcal{H}_{ln}$ is unique up to $\mathcal{H}_{vx}$ given $\mathcal{U}^T$ (by Theorem 28.4 and its corollary in \citet{warner1965modern}). However, it may not hold that $\mathcal{H}_{vx}$ is uniquely determined by $\mathcal{U}$ or $L(\mathcal{U})$, and thus the transform may not be unique. Note that empirically this limitation can be offset to a great extent by running multiple independent trials for the learning, though doing so will increase the running time\footnote{Seeking the best trade-off point between the number of learning trials and the concentration of resulting transforms is an interesting problem. It is not solved in this paper, and left to the future work.}. The discussion on this topic is continued in Section \ref{sec:sgs_vs_gft}.
\par

\subsection{Ensemble Based Transform}
Practically, it is more convenient to have only one method rather than multiple to compute the magnitudes of eigencomponents. Also, empirically, the SGS methods can have different performance in various scenarios (detailed in Section \ref{sec:sgs_vs_gft}). Thus, the ensemble can offer a more convenient and robust method. 
\begin{framed}
	\noindent\textbf{Algorithm 6: Ensemble Based Transform on SGs (ENS)}\\
	(1) For each $K$, compute $\mathcal{M}^K_{APPRX-LS}$, $\mathcal{M}^K_{ADJ-DIFF}$, $\mathcal{M}^K_{IN-AGG}$ and $\mathcal{M}^K_{LN-VX}$.
	\\
	(2) For each $K$, compute the magnitudes of eigencomponents carried by $s$ by 
	\begin{equation}
		\begin{split}
			\mathcal{M}^K_{ENS} &= w_{APPRX-LS} \mathcal{M}^K_{APPRX-LS} + w_{ADJ-DIFF} \mathcal{M}^K_{ADJ-DIFF} \\
			&+ w_{IN-AGG} \mathcal{M}^K_{IN-AGG} + w_{LN-VX} \mathcal{M}^K_{LN-VX}
		\end{split}
	\end{equation}
	where $w_{APPRX-LS}, w_{ADJ-DIFF}, w_{IN-AGG}, w_{LN-VX} \geq 0$.
\end{framed}
\par
Note that the weights for \textbf{APPRX-LS}, \textbf{IN-AGG}, \textbf{ADJ-DIFF} and \textbf{LN-VX} can be customized depending on the application scenarios. Particularly, as \textbf{IN-AGG} has been proved being effective exclusively on pulse-like signals, it can be assigned a low weight elsewhere. A limitation of \textbf{ENS} is that, when ${M}^K$'s \footnote{For convenience, the notation ${M}^K$ is used to denote in general $\mathcal{M}^K_{APPRX-LS}$, $\mathcal{M}^K_{ADJ-DIFF}$, $\mathcal{M}^K_{IN-AGG}$ and $\mathcal{M}^K_{LN-VX}$.} are not normalized, the weighted sum is likely to encounter the numerical bias issue (i.e. the norms of ${M}^K$'s can be significantly different from each other) if the weights are not carefully assigned. Constructing $\mathcal{M}^K_{ENS}$ on normalized ${M}_{\mathbb{1}}^K$'s is a solution; however, preserving the norms of ${M}^K$'s is useful in some scenarios (empirical examples are discussed in Sections \ref{sec:low_pass_filter} and \ref{sec:amplitude}). On the other hand, the norms of ${M}^K$'s, though primarily depend on $s$, are also affected by the internal mechanisms of the SGS methods (e.g. the learned transform in \textbf{LN-VX}) as well as some intrinsic characteristics of the considered graph (e.g. the localization of eigenvectors and the graph curvature in \textbf{ADJ-DIFF}), which makes it difficult to provide a generic weighting strategy \footnote{This issue is not well solved in this paper, and left to the future work.}. Empirically, the weighting on normalized ${M}_{\mathbb{1}}^K$'s is further discussed in Section \ref{sec:sgs_vs_gft}. 
\par
As discussed, each SGS method has its own limitations. In addition to these defects, there is another limitation shared by all methods. That is, the SGS methods are weak at decoding the magnitudes at zero eigencomponents (i.e. $\lambda_i = 0$). In classic GSP, given a constant real-valued signal, the magnitudes at zero eigencomponents should be the greatest and all others are valuated by $0$ (because eigenvectors are orthonormal). However, it can be shown that this property does not always hold in the SGS methods on constant vector-valued signals (i.e. all nodes share the same signal vector). First, \textbf{APPRX-LS} requires to solve $\mathcal{B}_K \cdot \hat{f}_K = \nabla s_K$ (\textbf{Algorithm 2} step (3)). For a constant $s$, $\nabla s_K = \boldsymbol{0}$, where $\boldsymbol{0}$ denotes the zero vector, and thus $\hat{f}_K = \boldsymbol{0}$ is a solution. Consequently, the GFT over $\hat{f}_K$ (\textbf{Algorithm 2} step (7)) can also result in $\boldsymbol{0}$, which is the worst case against the expectation. Second, \textbf{IN-AGG} aggregates $\nabla s_K$ at each node from its neighborhood (\textbf{Algorithm 3} step (3)), and thus the approximated signal $\hat{f}_K$ based on the aggregation (\textbf{Algorithm 3} step (4)) is $\boldsymbol{0}$ if $s$ is constant. And, thus, the resulting GFT over $\hat{f}_K$ (\textbf{Algorithm 3} step (5)) is $\boldsymbol{0}$. Third, \textbf{ADJ-DIFF} requires to compute $\langle \nabla s, \nabla u_i \rangle$ (\textbf{Algorithm 4} step (3)), which results in $0$ for any constant $s$. Finally, \textbf{LN-VX} requires to compute the GFT of $\nabla s_K$ in the line graph domain (\textbf{Algorithm 5} step 9), which results in $\boldsymbol{0}$ for any constant $s$, and further leads to the weighted eigenbasis becoming a zero matrix (\textbf{Algorithm 5} step 10), then eventually lands in $\boldsymbol{0}$ as the output magnitudes. Therefore, when utilizing the SGS methods in practice, the magnitudes at zero eigencomponents need to be handled carefully.
\par
Both the rationale and limitations of the five SGS methods have been discussed in theory. Next, in Section \ref{sec:experiments}, the empirical effectiveness and utility of these methods are elaborated. 
\par

\section{Experiments} \label{sec:experiments}
The primary concentration of this section is experimentally justifying the effectiveness of the SGS methods and demonstrating their utility. Firstly, in Section \ref{sec:sgs_vs_gft}, the SGS methods are compared to the GFT using real-valued graph signals to examine if they agree with each other \footnote{Recall that the SGS methods actually take $\nabla s$ as input, and thus they are applicable to both real-valued and vector-valued signal.}. Secondly, in Section \ref{sec:low_pass_filter}, the SGS methods are applied in a low-pass filtering case study to examine if they are able to capture the effect of filtering in the spectral domain. Finally, in Section \ref{sec:amplitude}, another case study analyzing an over-smoothed node embedding learning model is demonstrated showing the utility of the SGS methods in diagnosing and understanding behaviors of node embedding learning models.

\subsection{Compare SGS Methods to GFT} \label{sec:sgs_vs_gft}
Two primary tasks are performed in this section. One is to examine if the magnitudes of eigencomponents produced by the SGS methods agree with those produced by the GFT on real-valued signals, and the other one is to examine if the SGS methods agree with each other. Positive results from both of the tasks will substantiate the effectiveness of the SGS methods. In addition, an empirical study of the learning step of \textbf{LN-VX} is discussed thereafter. 
\par
\begin{framed}
	\noindent\textbf{Task 1: SGS vs GFT}\\
	$\triangleright$ \textbf{Objective:}
	\par
	Justify the effectiveness of the SGS methods by comparing them to the GFT on real-valued signals. Specifically, the resulting normalized magnitudes of eigencomponents produced by the SGS methods and those produced by the GFT are compared by the cosine similarity. The higher the similarities, the better effectiveness.
	\par
	\noindent
	$\triangleright$ \textbf{Settings:}
	\par
	$\bullet$ A random graph $\mathcal{G} = (\mathcal{V}, \mathcal{E})$ of 50 nodes generated by the ERM or the SBM. The ERM is configured by $p = 0.1$. The SBM is configured by a random choice of the number of blocks in the range of $[2, 10]$, evenly distributed block sizes, and uniformly assigned edge probabilities. Only connected graphs are considered. 
	\par
	$\bullet$ A normalized real-valued graph signal $s$ which is either a random signal generated uniformly or a pulse signal with a randomly chosen pulse position. 
	\par
	$\bullet$ For \textbf{LN-VX}, the terminal condition for its learning step is set to $MSE \leq 0.000001$ or $1000$ epochs whichever is met first. And 50 learning trials (for the step (7) in \textbf{Algorithm 5}) are configure. 
	\par
	$\bullet$ For \textbf{ENS}, all element SGS methods are evenly weighted by $1$.
	\par
	\noindent
	$\triangleright$ \textbf{Trials:}
	\par
	$\bullet$ \textbf{ERM-Rand}: 100 ERM graphs, a random signal for each.
	\par
	$\bullet$ \textbf{ERM-Pulse}: 100 ERM graphs, a pulse signal for each.
	\par
	$\bullet$ \textbf{SBM-Rand}: 100 SBM graphs, a random signal for each.
	\par
	$\bullet$ \textbf{SBM-Pulse}: 100 SBM graphs, a pulse signal for each.
	\par
	\noindent
	$\triangleright$ \textbf{Steps:}
	\par
	(1) Construct SGs: $\{ \mathcal{G}_K \}$.
	\par
	(2) For each $\mathcal{G}_K$, compute the magnitudes of eigencomponents $\mathcal{M}^K_{GFT}$ using the GFT by $\mathcal{M}^K_{GFT}(i) = |\langle \big (u^K_i \big)^* \cdot s \rangle|$.
	\par
	(3) For each $\mathcal{G}_K$, compute $\mathcal{M}^K$'s by using the SGS methods.
	\par
	(4) Compute the normalized inner product between $\mathcal{M}^K_{GFT}$ and each of $\mathcal{M}^K$'s, and denote $\mathcal{C}_{ADJ-DIFF} = \langle \mathcal{M}^K_{GFT}, \mathcal{M}^K_{ADJ-DIFF} \rangle_{\mathbb{1}}$, where $\langle \cdot, \cdot \rangle_{\mathbb{1}}$ denotes the dot product on $l2$-normalized operands, and $\mathcal{C} \in [0, 1]$. Note that the norms of $\mathcal{M}^K_{GFT}$ and $\mathcal{M}^K$'s are not considered in this task in order to eliminating the numerical impact from the amplitude of $s$ and the internal mechanisms of the methods.
\end{framed}
\par
\begin{figure}
	\centering
	\includegraphics[width=0.8\textwidth]{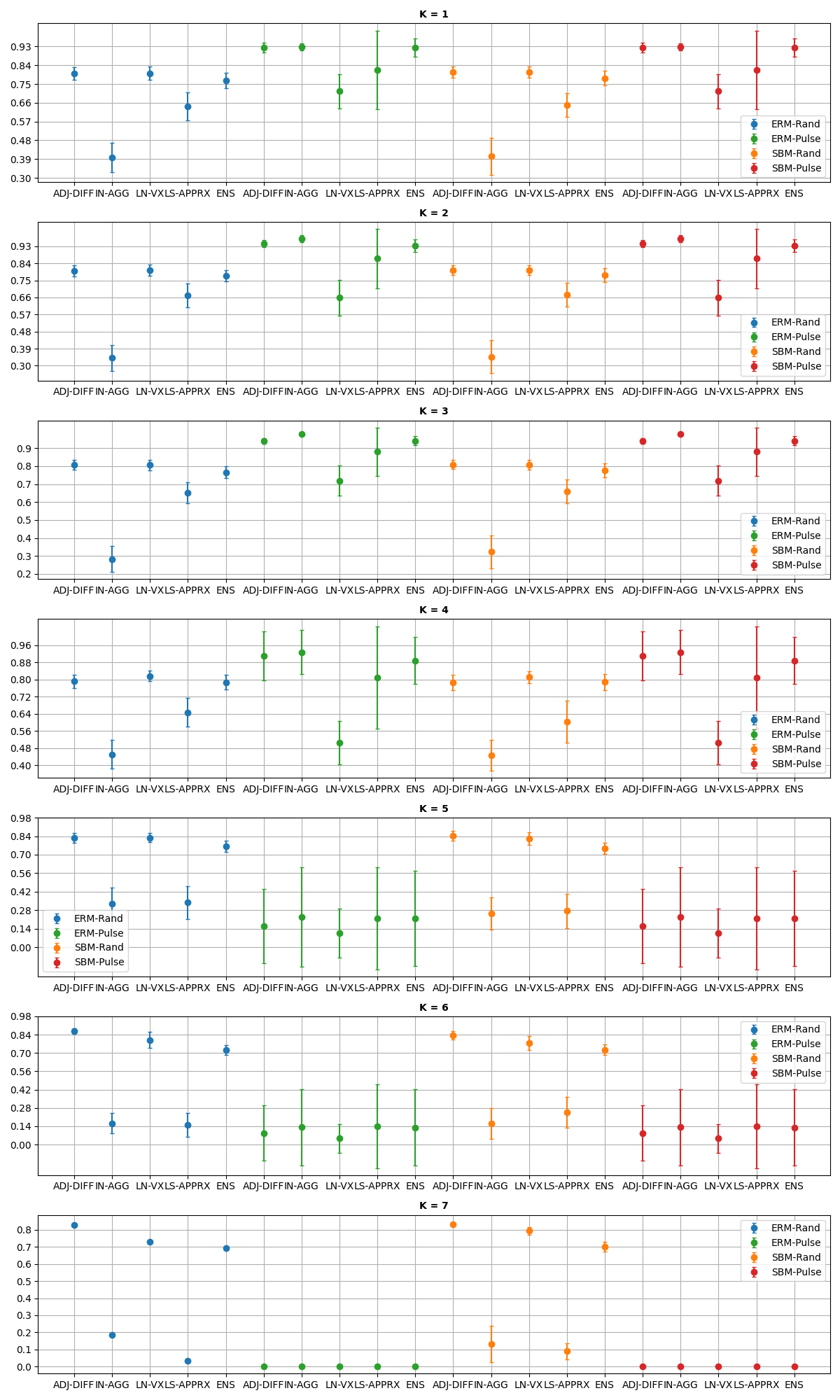}
	\caption{\textbf{Task 1}. $\mathcal{C}_{ADJ-DIFF}$, $\mathcal{C}_{APPRX-LS}$, $\mathcal{C}_{IN-AGG}$, $\mathcal{C}_{LN-VX}$ and $\mathcal{C}_{ENS}$ over all $K$'s and all trials. Trials are colored, x-axis shows the SGS methods, y-axis shows the means of metrics, and the standard deviation of each metric is displayed as the length of vertical bar \protect\footnotemark.}
	\label{fig:all_classic_vs_sas_and_trans}
\end{figure}
\begin{figure}
	\centering
	\includegraphics[width=1\textwidth]{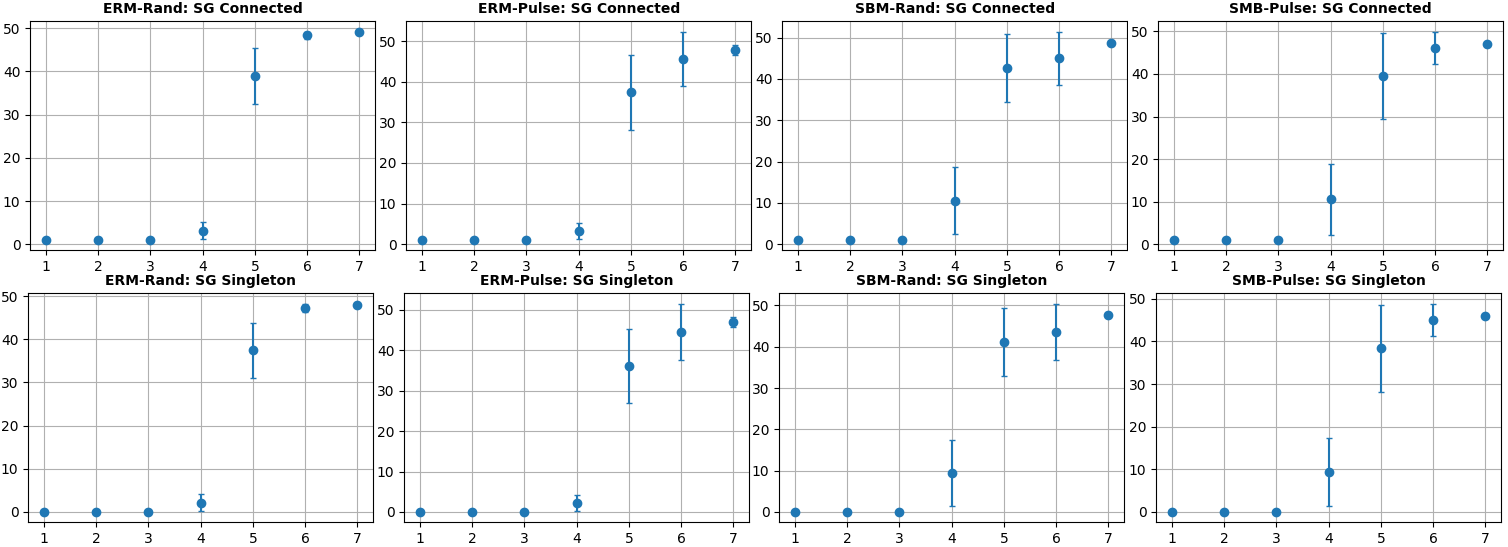}
	\caption{\textbf{Task 1}. The statistics of numbers of connected components (the upper row) and numbers of singleton components (the lower row) of all trials. X-axis is $K$, and y-axis is the number of components.}
	\label{fig:sg_stats}
\end{figure}

The results of \textbf{Task 1} are shown in Figure \ref{fig:all_classic_vs_sas_and_trans}. Several observations need to be highlighted. First, $K \leq 4$, at least one SGS method performs well (typically $\mathcal{C} \geq 0.8$). This observation strongly supports the effectiveness of SGS methods. Second, for $K \geq 5$, \textbf{ADJ-DIFF} and \textbf{LN-VX} keep performing fairly well (typically $\mathcal{C} \geq 0.7$) on random signals. Third, no method is effective on pulse signals at $K \geq 5$. This is primarily caused by the increasing number of singleton components as $K$ increases, and the randomly assigned pulses are more likely to fall on the singleton nodes. When this happens, unfortunately, $\nabla s$, as the key ingredient to all SGS methods, is actually undefined. To justify this explanation, the numbers of connected components and singleton components in all trials are counted and summarized in Figure \ref{fig:sg_stats}. The high correlation between the two counts (can be easily observed by eye) indicates that the connected components at $K \geq 5$ are primarily singleton. The correlation between the performance of SGS methods on pulse signals and the numbers of singleton components then can be easily observed by comparing Figure \ref{fig:all_classic_vs_sas_and_trans} and \ref{fig:sg_stats}. Fourth, \textbf{IN-AGG} is exclusively effective on the pulse cases. This empirically justifies \textbf{Proposition 1}, and thus evidences the limitation of \textbf{IN-AGG}. Fifth, \textbf{LN-VX} is typically weaker than others at pulses. This weakness can be interpreted based on \textbf{Equation} \ref{eq:weighted_eigenbasis_transform_element}. For instance, for any $K$ and a signal $s$ singly pulsing at the $j^{th}$ node, if the magnitude of the $i^{th}$ eigencomponent carried by $s$ is great (i.e. $\langle s, u_i^* \rangle$ is great), $u_i(j)$ must be great. By the transform (i.e. \textbf{Equation} \ref{eq:eigenbases_transform}), $\mathcal{U}(j, i) = \sum\limits_{k=0}^{|L(\Lambda)| - 1} \big\langle h_{{ln}_i}, L(u_k) \big\rangle h^T_{{vx}_j}(k)$ is thus great. However, by \textbf{Equation} \ref{eq:weighted_eigenbasis_transform_element}, the weight (i.e. $\Big|\big\langle \nabla s, L(u_k)^* \big\rangle\Big|$) imposed on $\big\langle h_{{ln}_i}, L(u_k) \big\rangle$ is not necessarily great as most values of $\nabla s$ are likely to be $0$, which is the major cause of this issue. Finally, \textbf{APPRX-LS} typically has larger variance of performance than others, which is as expected due to the approximation in solving the LLS.
\par
\footnotetext{Errorbars, in this paper, always indicate means and standard deviations if no particular specification is noted.}
\par 
\par
In \textbf{Task 2}, the SGS methods are examined if they agree with each other. Strong agreement enhances the effectiveness of the methods.
\begin{framed}
	\noindent\textbf{Task 2: Agreement Between SGS Methods}\\
	$\triangleright$ \textbf{Objective:}
	\par
	Examine if the SGS methods produce accordant results. Specifically, pairwise similarities of their results are computed. The higher the similarities, the stronger the agreement. 
	\par
	\noindent
	$\triangleright$ \textbf{Settings and Trials:}
	\par
	Same as \textbf{Task 1}
	\par
	\noindent
	$\triangleright$ \textbf{Steps:}
	\par
	(1) Obtain $\mathcal{M}^K_{APPRX-LS}$, $\mathcal{M}^K_{ADJ-DIFF}$, $\mathcal{M}^K_{LN-VX}$ and $\mathcal{M}^K_{IN-AGG}$ computed in the step (3) of \textbf{Task 1}.
	\par
	(2) Compute pairwise cosine similarities of $\mathcal{M}_{\mathbb{1}}^K$'s, where $\mathbb{1}$ denotes the $l2$-normalization.
\end{framed}
\par
\begin{figure}
	\centering
	\includegraphics[width=1\textwidth]{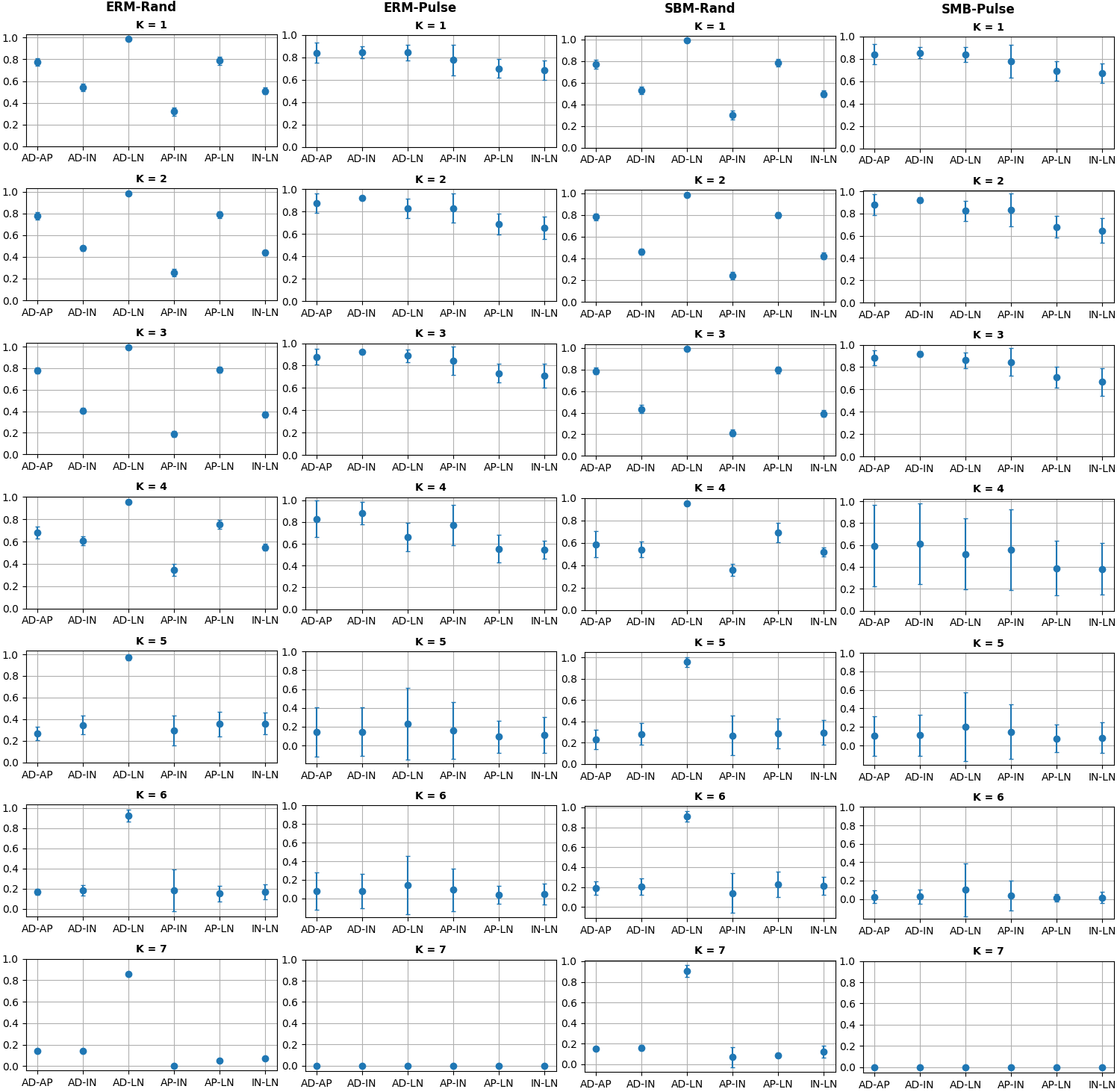}
	\caption{\textbf{Task 2}. Pairwise cosine similarities of $\mathcal{M}_{\mathbb{1}}^K$'s produced by the SGS methods. \textbf{AD}, \textbf{AP}, \textbf{IN} and \textbf{LN} denote \textbf{ADJ-DIFF}, \textbf{APPRX-LS}, \textbf{IN-AGG} and \textbf{LN-VX} respectively.}
	\label{fig:sgs_pw_comp}
\end{figure}

The results of \textbf{Task 2} are shown in Figure \ref{fig:sgs_pw_comp}. In the random signal cases, \textbf{ADJ-DIFF} and \textbf{LN-VX} highly agree with each other over all $K$'s. Another two pairs, \{\textbf{APPRX-LS}, \textbf{ADJ-DIFF}\} and \{\textbf{APPRX-LS}, \textbf{LN-VX}\} have relatively high accordance at $K \leq 4$, but become increasingly discrepant at $K \geq 5$. \textbf{IN-AGG} is not similar to anyone. On the other hand, in the pulse single cases, all methods highly agree with each other at $K \leq 3$, but this agreement sharply descends at $K \geq 4$. These results are consistent with the results of \textbf{Task 1} as shown in Figure \ref{fig:all_classic_vs_sas_and_trans}.
\par
\par
Based on the results from \textbf{Task 1} and \textbf{2}, several conclusions can be made.
\begin{framed}
	\noindent\textbf{Conclusions of Tasks 1 and 2}\\
	(1) When the input signals are not pulse-like, \textbf{ADJ-DIFF} and \textbf{LN-VX} are more effective than others, and they typically produce agreeing results. \\
	(2) \textbf{APPRX-LS} is effective at lower $K$'s, it agrees with \textbf{ADJ-DIFF} and \textbf{LN-VX} to a great extent.\\
	(3) \textbf{ADJ-DIFF}, \textbf{LN-VX}, \textbf{APPRX-LS} and \textbf{IN-AGG} all perform acceptably on pulse-like signals at lower $K$'s, and produce similar results. \\
	(4) No SGS method is effective on pulse-like signals at higher $K$'s.
\end{framed}
\par
\begin{figure}
	\centering
	\includegraphics[width=1\textwidth]{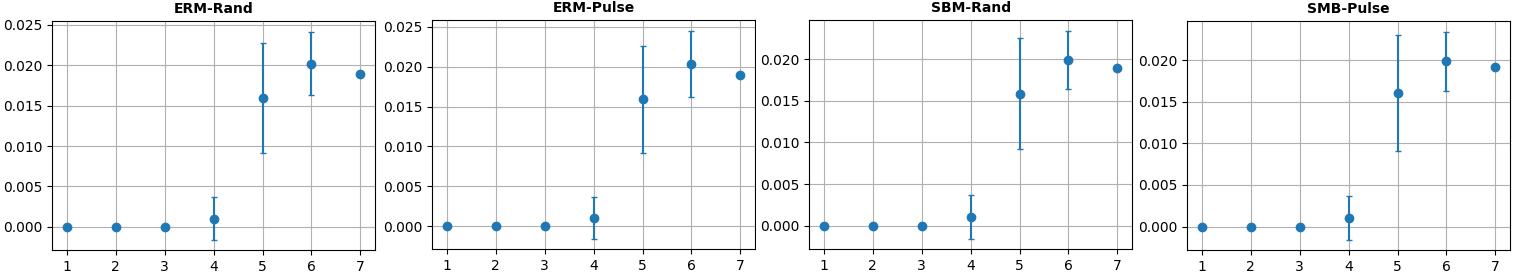}
	\caption{\textbf{Task 1}. The final MSEs of the learning step of \textbf{LN-VX} for the four trials in \textbf{Task 1}. X-axis is $K$, and y-axis is MSE.}
	\label{fig:ln_to_vs_eigs_stats}
\end{figure}
\begin{figure}
	\centering
	\includegraphics[width=1\textwidth]{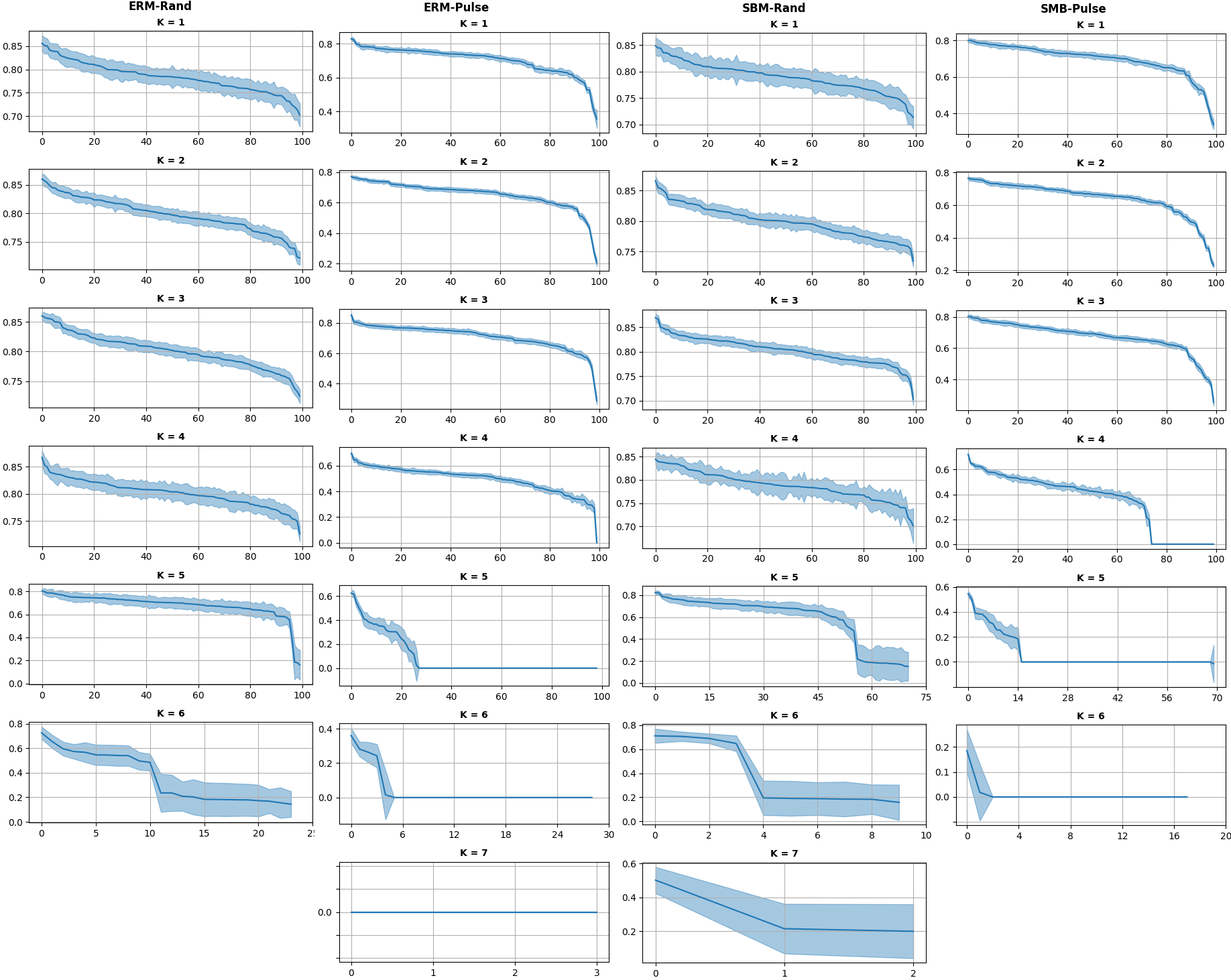}
	\caption{\textbf{Task 1}. The distributions of cosine similarities between \textbf{LN-VX} and the GFT for all testing graphs in all trials. The results are sorted by the means. X-axis is indexed by the testing graphs (after the sorting), y-axis is cosine similarity, and the bands show the standard deviations.}
	\label{fig:gsp_vs_ln_vx_50_learn_trials}
\end{figure}

$\mathcal{M}^K_{LN-VX}$, in \textbf{Task 1} and \textbf{2}, is computed by the expectation over $50$ learning trials. However, the learning performance and the empirical impact from the nonuniqueness limitation are not yet clear. These two topics are discussed here, and the discussion is based on the leaning trials performed in \textbf{Task 1}. The final MSEs of the learning trials over all testing graphs in \textbf{ERM-Rand}, \textbf{ERM-Pulse}, \textbf{SBM-Rand} and \textbf{SBM-Pulse} (and each testing graph corresponds to $50$ learning trials) are shown in Figure \ref{fig:ln_to_vs_eigs_stats}. The learning performance is fairly acceptable for all $K$'s in all trials. Specifically, the MSEs are close to zero for $K \leq 4$, and lower than $0.025$ for the rest. The rise of MSEs at $K \geq 5$ attributes to the increasing numbers of singleton components. The correlation between the MSEs and the numbers of singleton components is self-explanatory by comparing Figure \ref{fig:ln_to_vs_eigs_stats} to Figure \ref{fig:sg_stats}. This correlation also evidences the first limitation of \textbf{LN-VX}. On the other hand, to understand how the nonuniqueness of learning impacts the results of \textbf{LN-VX}, the distribution of cosine similarities between $\mathcal{M}^K_{\mathbb{1}LN-VX}$ and $\mathcal{M}^K_{\mathbb{1}GFT}$ for each testing graph is computed, and the results are shown in Figure \ref{fig:gsp_vs_ln_vx_50_learn_trials}. For random signal cases, at $K \leq 4$, the performance of \textbf{IN-VX} is fairly stable. On most graphs, no sharp drop is observed, and the variances are also acceptable. However, at $K \geq 5$, explicit plunges can be observed, and, when $K \geq 6$, on a half of the graphs, $\mathcal{M}^K_{\mathbb{1}LN-VX}$ is nearly independent of $\mathcal{M}^K_{\mathbb{1}GFT}$. For pulse signal cases, the drops occur at all $K$'s, though, at $K \leq 4$, \textbf{LN-VX} performs poorly on a small portion (about $20\%$) of the graphs. Moreover, at $K \geq 5$, most $\mathcal{M}^K_{\mathbb{1}LN-VX}$ can be irrelevant. These results affirm the stability of \textbf{IN-VX} on non-pulse signals, and suggest that \textbf{IN-VX} should be carefully used on pulse signals, especially at higher $K$'s. The results also further confirm the aforementioned conclusions of \textbf{Task 1} and \textbf{2} on \textbf{IN-VX}. 
\par
According to the above conclusions, a practical suggestion on weighting the element SGS methods for \textbf{ENS} is provided:
\begin{framed}
	\noindent\textbf{Practical Suggestion 1: Weight SGS Methods for \textbf{ENS}} \\
	(1) When weighting the element methods for \textbf{ENS}, \textbf{ADJ-DIFF}, \textbf{LN-VX} and \textbf{APPRX-LS}, can be particularly emphasized for lower $K$'s (typically $K \leq 4$ \protect\footnotemark), and \textbf{ADJ-DIFF}, \textbf{LN-VX} can be assigned higher weights than \textbf{APPRX-LS}. For higher $K$'s, \textbf{APPRX-LS} needs to be suppressed. \\
	(2) If the signals are given being pulse-like, then \textbf{IN-AGG} can be joined to \textbf{ADJ-DIFF}, \textbf{APPRX-LS} and \textbf{LN-VX} while \textbf{LN-VX} needs be moderately suppressed. \\
	(3) When the time complexity is stressed, \textbf{LN-VX} and \textbf{APPRX-LS} can be abandoned, though the robustness of \textbf{ENS} may be traded off to some extent.
\end{framed}
\footnotetext{$K \leq 4$ is an empirical criterion to define ``lower $K$'s" without rigorous theoretical justification, and this criterion can vary in other cases. Future work is needed to make the weighting strategies of \textbf{ENS} more rigorous and effective.}
\par
In the next section, the discussion on the effectiveness of SGS methods is extended. One of the most important applications in GSP, filtering, is concentrated. It is examined if the SGS methods are able to capture the effects of filtering. A low-pass filtering case study is elaborated.

\subsection{A Low-Pass Filtering Case Study}\label{sec:low_pass_filter}
Filtering has been widely applied to graph learning models (\citet{kipf2016semi}; \citet{defferrard2016convolutional}). Traditionally, the GFT is the standard approach to capture the effects of filtering by decoding the magnitudes of eigencomponents for real-valued signals. Regarding vector-valued signals, the SGS methods are expected to possess the same functionality. To justify the effectiveness of the SGS methods in filtering, a low-pass filtering use case is studied.
\par
The low-pass filtering has been utilized and even attested to be an essential functionality of many graph learning models (\citet{nt2019revisiting}; \citet{yu2020graph}; \citet{wu2019simplifying}; \citet{li2020dirichlet}). The effect of low-pass filtering, without additional adjustments (e.g. learning on specific tasks), is embodied by a fact. That is, relatively, adjacent nodes are more likely to become similar (i.e. the signal is smoothed) \footnote{On weighted graphs, the fact is that adjacent nodes linked by a heavy-weight edge are more likely to become similar.} (\citet{shuman2013emerging}). Leveraging this effect, for graphs endowed with partition structures, low-pass filters can help learn node embeddings highly agreeing the partitions. A typical graph of this kind is concentrated in this case study. The construction of low-pass filters is a variety (\citet{nt2019revisiting}; \citet{shuman2013emerging}; \citet{hammond2011wavelets}). Amid various candidates, the total variation is a popular choice (\citet{buades2005review}; \citet{shuman2013emerging}), and it is well known with its smoothing effect (\citet{berger2018graph}; \citet{berger2020efficient}; \citet{chen2015signal}). Thus, it is chosen as the primary objective of the node embedding learning model in this case study. However, a notorious issue of low-pass filters is that an ill-controlled filtering process can result in over-smoothing (i.e. roughly, all nodes become hardly distinguishable in the output) which is likely to further obfuscate the spectral patterns of partitions carried by the resulting embeddings. To avoid the over-smoothing, a simple regularizer is introduced keeping non-adjacent nodes as distant as possible. Combining the total variation and the regularizer, this regularized objective, intuitively, trends to raise both the density of each partition and the discrepancies between partitions. 
\par
The experiments of this case study are conducted over a shallow node embedding learning model equipped with the regularized low-pass filtering objective. A manual signal which guarantees significant magnitudes at high-frequency components and being irrelevant to the partition structures is assigned to the initial condition of the learning. A resulting embedding which perfectly solves the node clustering task (defined by the natural partition structures of the input graph), and thus guarantees being well low-pass filtered yet not over-smoothed, is selected for the spectral analysis. The magnitudes of eigencomponents of the initial condition and the selected resulting embedding are then respectively computed and compared aiming at examining if the effect of filtering is captured. Details are described in \textbf{Task 3.}
\par
%
\par
\begin{framed}
	\noindent\textbf{Task 3: A Regularized Low-Pass Filtering Case Study}\\
	$\triangleright$ \textbf{Objective:}
	\par
	Examine if the SGS methods are able to detect the frequency changes in signals before and after a regularized low-pass filtering. Specifically, as the effect of the filtering is reflected by the changing in relative distances between nodes, the partition indicators (e.g. the Fiedler eigencomponents) are expected to be significantly accentuated, and other key changes in eigencomponent magnitudes are expected to capture and explain the smoothing effect. 
	\par
	\noindent
	$\triangleright$ \textbf{Settings and Trials:}
	\par
	$\bullet$ A variant of the Caveman graph $\mathcal{G}=(\mathcal{V}, \mathcal{E})$ (visualized in Figure \ref{fig:sg_sample} at $K=1$) which contains four communities (colored in Figure \ref{fig:sg_sample}): \{A\}, \{B, E, F, K\}, \{D, I, J, M\} and \{C, G, H, L\}. $|\mathcal{V}|=13$ and $|\mathcal{E}|=15$. 
	\par
	$\bullet$ A shallow learning model is implemented to learn node embeddings. The objective is a total variation \protect\footnotemark defined as 
	\begin{equation}\label{eq:tv_objective}
		\tau = \mathbb{E} \big[\nabla s^2 (x, y)\big] \Big|_{\forall x \sim y \in \mathcal{E}}
	\end{equation}
	which is regularized by 
	\begin{equation}
		\varepsilon = \mathbb{E} \big[\nabla \Gamma^2 (x, y)\big] \Big|_{\forall x \not\sim y, x \neq y \in \mathcal{V}}
	\end{equation}
	The learning problem is then formulated as
	\begin{equation}
		\argmin\limits_{s} \big(w_{\tau} \tau +  w_{\varepsilon} \varepsilon \big)
	\end{equation}
	where $s$ denotes the node embedding in desire, $w_{\tau}$ is set to $1$ and $w_{\varepsilon}$ varies in the range $[0, 1]$ on the stride of $0.1$.
	\par
	$\bullet$ The initial condition of the learning is a manually assigned normalized 3-dimensional vector-valued signal:
	\begin{equation*}
		s_0 (x) = 
		\begin{cases}
			[0.58, 0.58, 0.58]	& \quad \text{if } x = \text{A, K, L, M} \\
			[1, 0, 0] & \quad \text{if } x = \text{B} \\
			[0, 1, 0] & \quad \text{if } x = \text{C} \\
			[0, 0, 1] & \quad \text{if } x = \text{D} \\
			[-1, 0, 0] & \quad \text{if } x = \text{F, G, J} \\
			[0, -1, 0] & \quad \text{if } x = \text{E, H, I}
		\end{cases}
	\end{equation*} 
	\par
	$\bullet$ The terminal condition of the learning is determined by a fixed number of epochs, $3500$, to empirically guarantee the convergence. 
	\par
	\noindent
	$\triangleright$ \textbf{Steps:}
	\par
	(1) Learn a 3-dimensional node embedding on each $w_{\varepsilon}$, and the resulting vectors are normalized. \par
	(2) Cluster nodes with each learned embedding by the spectral clustering, and compute Adjusted Rand Index (ARI) (\citet{hubert1985comparing}) and Adjusted Mutual Information (AMI) (\citet{vinh2010information}) to evaluate the performance of clustering. Both ARI and AMI are close to $1$ when reaching a perfect clustering and close to $0$ for uniformly random cluster label assignments. 
	\par
	(3) An embedding $s$ with the minimal $w_{\varepsilon}$ resulting in a perfect clustering (i.e. $\text{ARI} = 1$ or $\text{AMI} = 1$ whichever is met) is selected for the spectral analysis.
	\par
	(4) $\mathcal{M}_{\mathbb{1}}^K$'s and $\mathcal{M}^K$'s are computed on this $s$. $20$ learning trials are configured for \textbf{LN-VX}. And $\mathcal{M}^K_{ENS}$ is computed weighting \textbf{ADJ-DIFF}, \textbf{LN-VX} and \textbf{APPRX-LS} respectively by $0.4$, $0.4$, $0.2$ at $K \leq 4$ and $0.5$, $0.5$, $0$ at $K \geq 5$. The spectral analysis on $s_0$ and $s$ are all based on  $\mathcal{M}^K_{ENS}$.
\end{framed}
\footnotetext{Note that the original total variation is defined as the 1-Dirichlet form of signal gradients: $\tau = \sum\limits_{x \sim y \in \mathcal{E}} w_{xy} \big(s(x) - s(y)\big)^2$, where $w_{xy}$ is the edge weight (\citet{shuman2013emerging}).}
\par
$\mathcal{G}$ is constructed so because the Cavemen graphs are good test cases for the node clustering task (\citet{kloster2014heat}; \citet{lim2014slashburn}; \citet{kang2011beyond}; \citet{neubauer2009towards}), and simple enough yet sufficiently non-trivial to demonstrate the partition structures in detail. To accommodate the Cavemen graph more relevantly, a number of modifications are imposed. First, a center node $A$ is introduced, and it is alone forming a singleton cluster (due to the symmetry of $\mathcal{G}$). Second, the ring structure of the partitions is changed to the star structure, in which very non-singleton partition is linked to $A$. Finally, to weaken the perfection of partitions, another three nodes $K$, $L$ and $M$ are linked to each of the non-singleton partitions by a single edge.
\par
\begin{figure}
	\centering
	\includegraphics[width=1\textwidth]{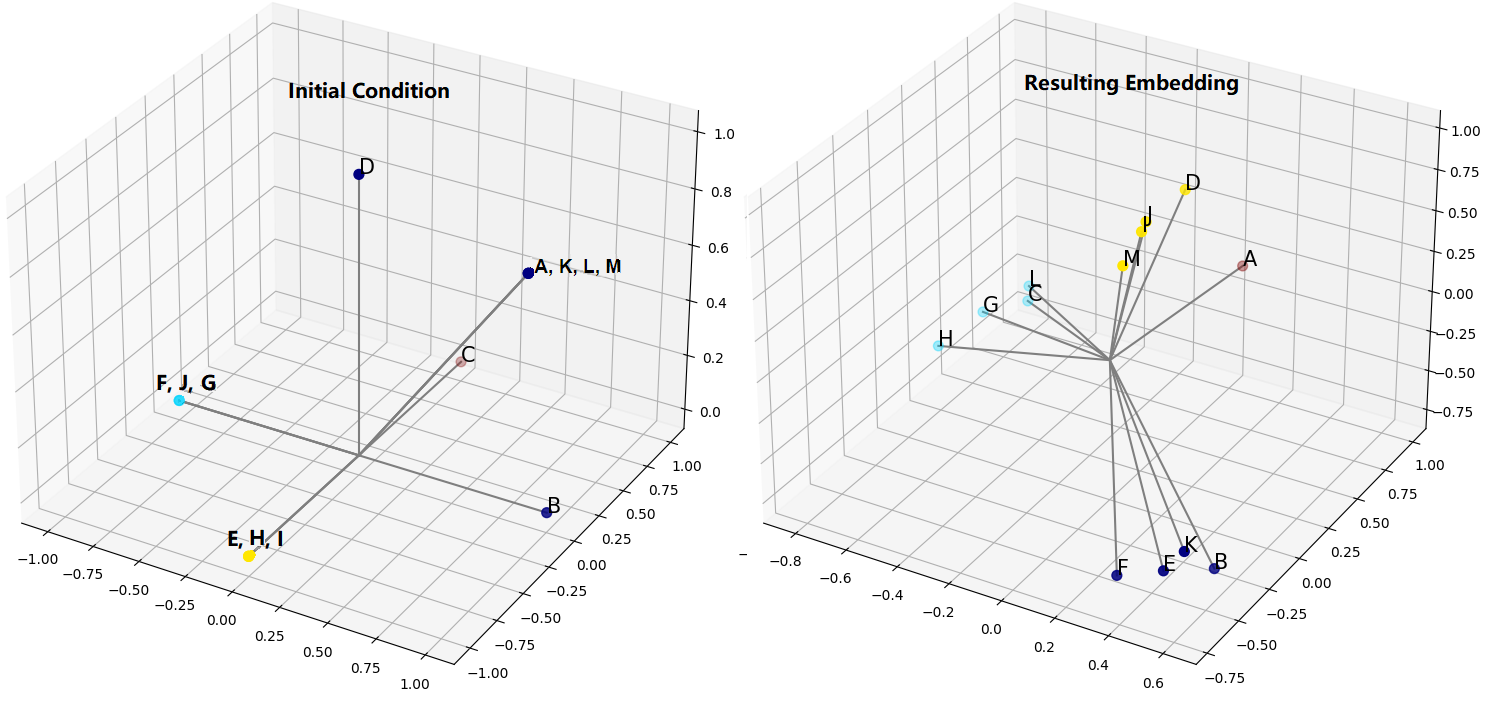}
	\caption{\textbf{Task 3}. The initial condition (left) and the selected resulting embedding (right). The clusterings are colored for the respective embeddings.}
	\label{fig:low_pass_signals}
\end{figure}

The number of dimensions of node embedding vectors is chosen to be $3$ for 
it is friendly for observations. The initial condition and the selected resulting embedding are visualized in Figure \ref{fig:low_pass_signals}.
\par
The manually assigned initial condition is discordant to the natural partition structures, and its ARI and AMI are $-0.19$ and $-0.29$ respectively. This creates great magnitudes at high-frequency eigencomponents. Specifically, these magnitudes are primarily originated from the members of each non-singleton partition (i.e. \{$B$, $E$, $F$, $K$\}, \{$C$, $H$, $G$, $L$\} and \{$D$, $I$, $J$, $M$\}) being scattered.
\par
The shallow learning model is chosen for its simplicity. The more popular graph neural network (GNN) models (e.g. ChebNet (\citet{defferrard2016convolutional}), GCN (\citet{kipf2016semi}), GAT (\citet{velivckovic2017graph}), SAGE (\citet{hamilton2017inductive}) and GGS-NN (\citet{li2015gated})), though playing dominant roles in this area, typically do not directly learn the vectors on nodes but a hidden transform defined in various ways (e.g. an analog to the filtering (\citet{defferrard2016convolutional}) and the attention mechanism (\citet{velivckovic2017graph})) \footnote{Actually the hidden transform is the most important benefit gained from the popular GNN models because it effectively refrains the linear (or even higher order) growth of complexity as the size of input graph increases.}, which is not as straightforward as the shallow learning model in demonstrating behaviors of the objective.
\par
\begin{figure}
	\centering
	\includegraphics[width=1\textwidth]{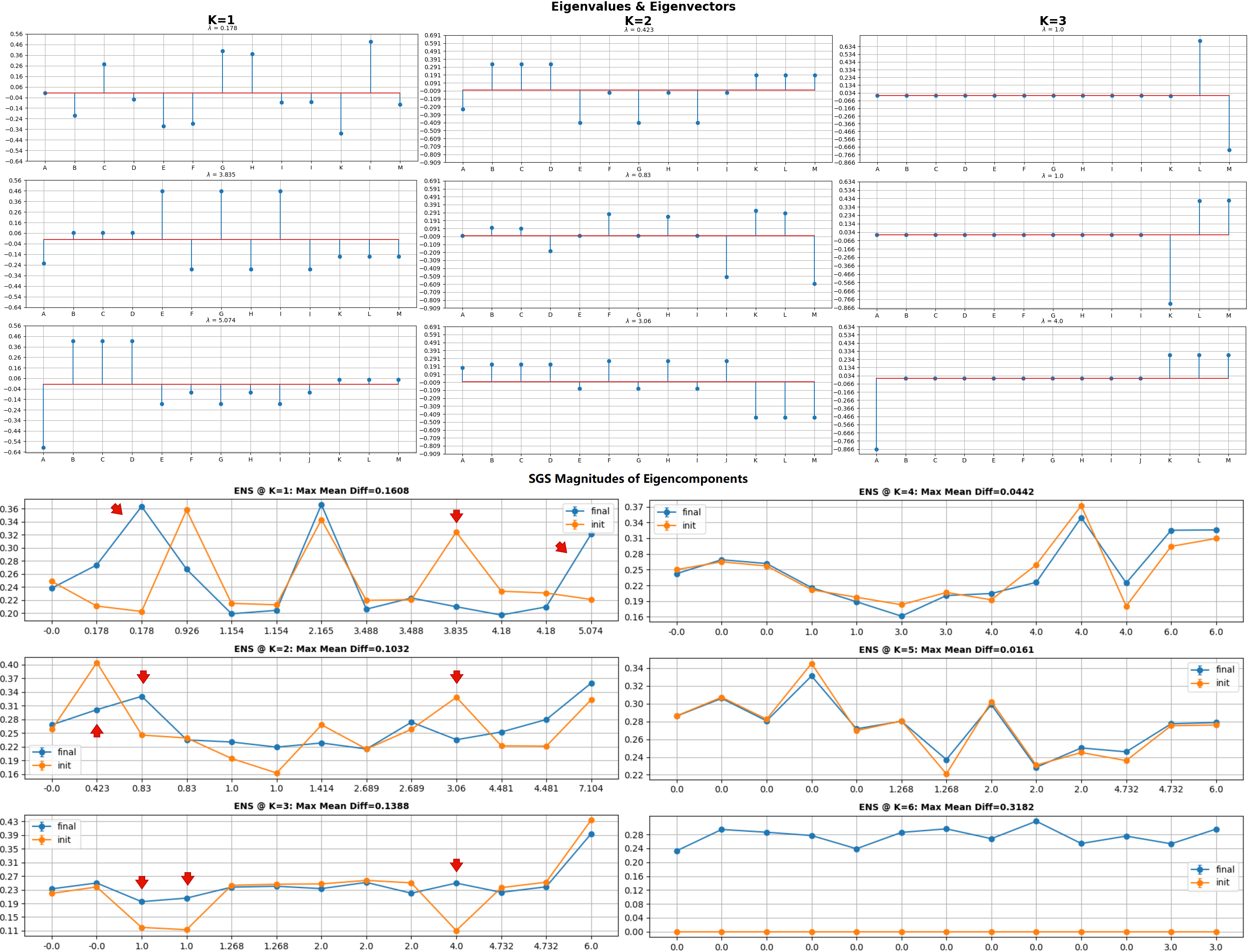}
	\caption{\textbf{Task 3}. Featured difference of the spectral characteristics between the initial condition and those of the low-pass filtered signal (i.e. the selected embedding). Specifically, red arrows in the lower half highlight the featured changes.  $\mathcal{M}_{\mathbb{1}ENS}^K$'s of the initial condition are drawn in yellow and labeled as ``init". And those of the low-pass filtered signal are drawn in blue and labels as ``final". The x-axis shows the eigenvalues, and y-axis shows the magnitudes. In addition, $\text{Max Diff} = ||\mathcal{M}_{\mathbb{1}init} - \mathcal{M}_{\mathbb{1}final}||_{\infty}$. The upper half shows several eigenvalues and eigenvectors that are necessary in understanding why the changes of magnitudes reflect the effect of low-pass filtering. At $K=1$, $\lambda_2=0.178$ (one of the Fiedler values, and $\lambda_1=0.178$), $\lambda_9=3.835$ and $\lambda_{12}=5.074$ are selected; at $K=2$, $\lambda_1=0.423$, $\lambda_2=0.83$ and $\lambda_9=3.06$ are selected; and at $K=3$, $\lambda_2=1.0$, $\lambda_3=1.0$ and $\lambda_9=4.0$ are selected. The x-axis shows the node labels, and y-axis shows the values of eigenvectors. \protect\footnotemark}
	\label{fig:low_pass_spectral_analysis_ens}
\end{figure}

At the step (3), a node embedding at $w_{\varepsilon} = 0.1$ attaining $\text{ARI} = 1$ and $\text{AMI} = 1$ is selected (Figure \ref{fig:low_pass_signals} right) \footnote{A video visualizing the entire learning process of the selected embedding is attached in Appendix \ref{sec:sup_file_list} \textbf{SUP-T3-1}.}. The key difference of the eigencomponent magnitudes between the initial condition and the selected embedding is highlighted (by red arrows) in Figure \ref{fig:low_pass_spectral_analysis_ens}. To justify these difference does capture the effect of low-pass filtering, first, significant accents of magnitudes at the Fiedler eigencomponents (at $K=1$) and other partition indicators are expected, and second, the causes of other magnitude changes are expected to be the smoothing effect. 
\par
\footnotetext{Full sets of eigenvalues and eigenvectors of all SGs can be found in Append \ref{sec:sup_file_list} \textbf{SUP-T3-2}. And full sets of $\mathcal{M}_{\mathbb{1}}^K$'s of the initial condition and the selected resulting embedding can be found in Append \ref{sec:sup_file_list} \textbf{SUP-T3-3}.}
\par
The Fiedler eigencomponents (at $K=1$) do gain great accents on the magnitudes during the learning. This is self-explanatory as shown in Figure \ref{fig:low_pass_spectral_analysis_ens}. And the accent at the second Fiedler eigencomponent reaches the highest difference between the two magnitude curves. The resulting embedding illustrated in Figure \ref{fig:low_pass_signals} (right) explicates that the relative distances between member nodes in each partition become much lower than those in the initial condition, which matches the pattern indicated by the Fiedler eigenvector shown in Figure \ref{fig:low_pass_spectral_analysis_ens}. In addition to the Fiedler eigencomponents, another indicator is at $\lambda_{12}$ (also at $K=1$). The corresponding eigenvector chiefly concentrates on the discrepancies between $A$ and $B, C, D$. As $A$ forms a singleton partition, these discrepancies reflect $A$ repelling other partitions. The noticeable accent at $\lambda_{12}$ thus indicates $A$ being distinguishable from other partitions.
\par
Other highlighted changes (marked in Figure \ref{fig:low_pass_spectral_analysis_ens}) in magnitudes capture the smoothing effect. First, at $K=1$, a sharp descent occurs at $\lambda_9=3.84$. The corresponding eigenvector requires adjacent members in each non-singleton partition being dissimilar to gain a high magnitude. Thus, the member nodes being pulled close to each other, as a consequence of smoothing, is the essential cause of this descent. Second, at $K=2$, two descents occurs at $\lambda_1=0.42$ and $\lambda_9=3.06$ respectively, where $\lambda_1$ is the Fiedler value of $2$-SG. The associated Fielder eigenvector implies different partition structures from those in $1$-SG. For instance, the adjacent pairs $\{B, K\}$ and $\{F, K\}$ (resp. $\{C, L\}$ and $\{H, L\}$ as well as $\{D, M\}$ and $\{J, M\}$), which are in the same partition of $1$-SG, are dissimilar. Thus, the similarities of these pairs, again resulted from the smoothing, significantly weaken the matching between the filtered signal and the Fiedler eigenvector. Similarly, the mismatching on the eigenvector of $\lambda_9=3.06$ attributes to the same cause. Third, the similarities of $\{F, K\}$, $\{H, L\}$ and $\{J, M\}$ further lead to the accent at $\lambda_2=0.83$ ($K=2$), which accords with what the eigenvector of $\lambda_2=0.83$ primarily suggests. Finally, at $K=3$, accents occur at the Fiedler eigencomponents (i.e. $\lambda_1=1.0$ and $\lambda_2=1.0$) and $\lambda_9=4.0$. These eigencomponents indicate the partition $\{A, K, L, M\}$ (of $3$-SG) in a star shape, and profile the oscillation of $K, L, M$ around $A$. Thus, the departure of $K, L, M$ from $A$, which is an indirect consequence of the smoothing, matches the patterns of oscillation better than the initial condition, and thereby contributes to the accents. 
\par
\begin{figure}
	\centering
	\includegraphics[width=0.6\textwidth]{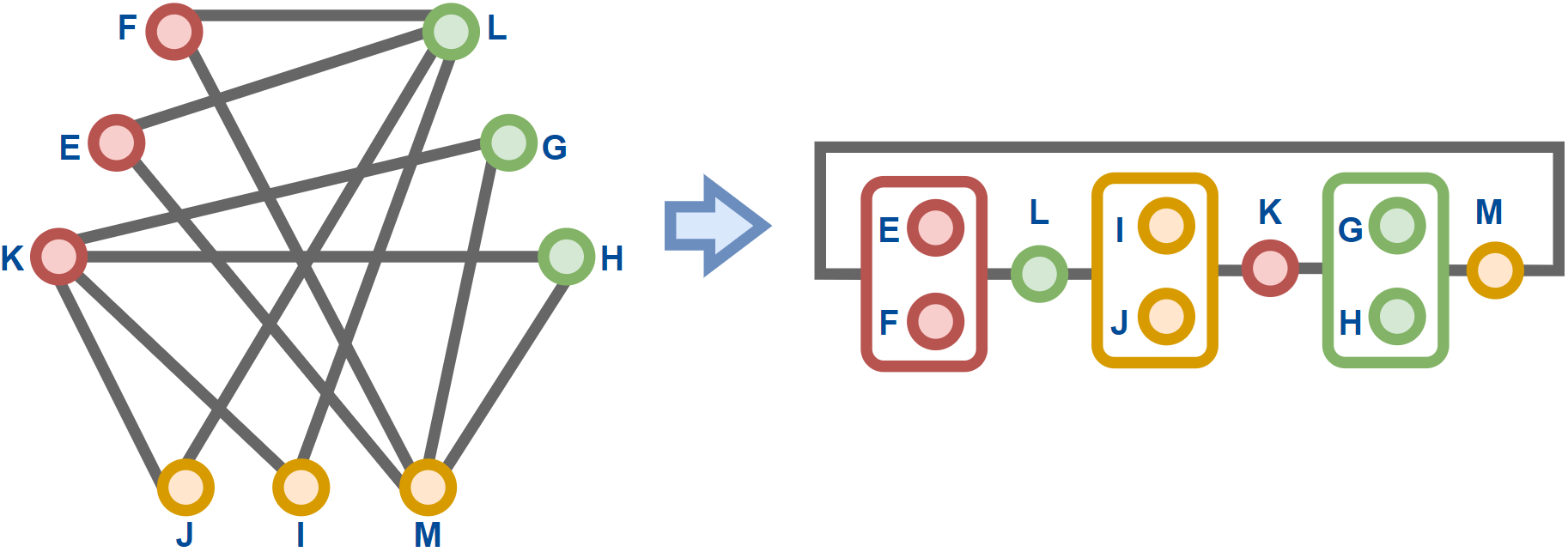}
	\caption{\textbf{Task 3}. The reduced circle graph (right) from the connected component (left) $\{E, F, G, H, I, J, K, L, M\}$ at $K=5$. The nodes in each box are considered to be equivalent.}
	\label{fig:reduced_circle}
\end{figure}

It has been justified that the effect of low-pass filtering can be captured by the difference of the eigencomponent magnitudes between the initial condition and the filtered signal. However, in addition to the difference, another remarkable phenomenon is also related to the behaviors of the learning model. That is, at $K \geq 3$, the $\mathcal{M}_{\mathbb{1}}^K$ curves of the initial condition and the filtered signal trend to become identical \footnote{The curves at $K=6$ are distinct as the magnitudes of the initial condition are all zeros. This is because $K$, $L$ and $M$ exclusively form a clique partition, and they are assigned the same signal vectors in the initial condition, which leads to $\nabla s$ being a zero vector. This issue is essentially the limitation of the SGS methods discussed at the end of Section \ref{sec:sgs}.}. For example, at $K=5$, the connected component $\{E, F, G, H, I, J, K, L, M\}$ can be reduced to a circle graph of $6$ edges as illustrated in Figure \ref{fig:reduced_circle}. It is known that the eigenvectors of a circle graph (except the trivial one) are sinuous (in different frequencies), and the matching between the signal and each eigenvector determines $\mathcal{M}_{\mathbb{1}}^K$'s. From Figure \ref{fig:low_pass_signals}, it is self-evident that, up to scaling, the initial condition and the selected embedding possess similar relative distances between adjacent nodes in the reduced circle graph. Hence, they have similar $\mathcal{M}_{\mathbb{1}}^K$'s corresponding to this connected component (i.e. from $\lambda_5$ to $\lambda_{12}$). As $\mathcal{M}_{\mathbb{1}}^K$'s only reflect relative distances between adjacent nodes in $K$-SGs, this phenomenon can be a consequence of expansion, contraction or inertia on the embeddings. To comprehensively justify the effectiveness of the SGS methods in capturing the filtering effect, this uncertainty needs to be eliminated. 
\par
\begin{figure}
	\centering
	\includegraphics[width=1\textwidth]{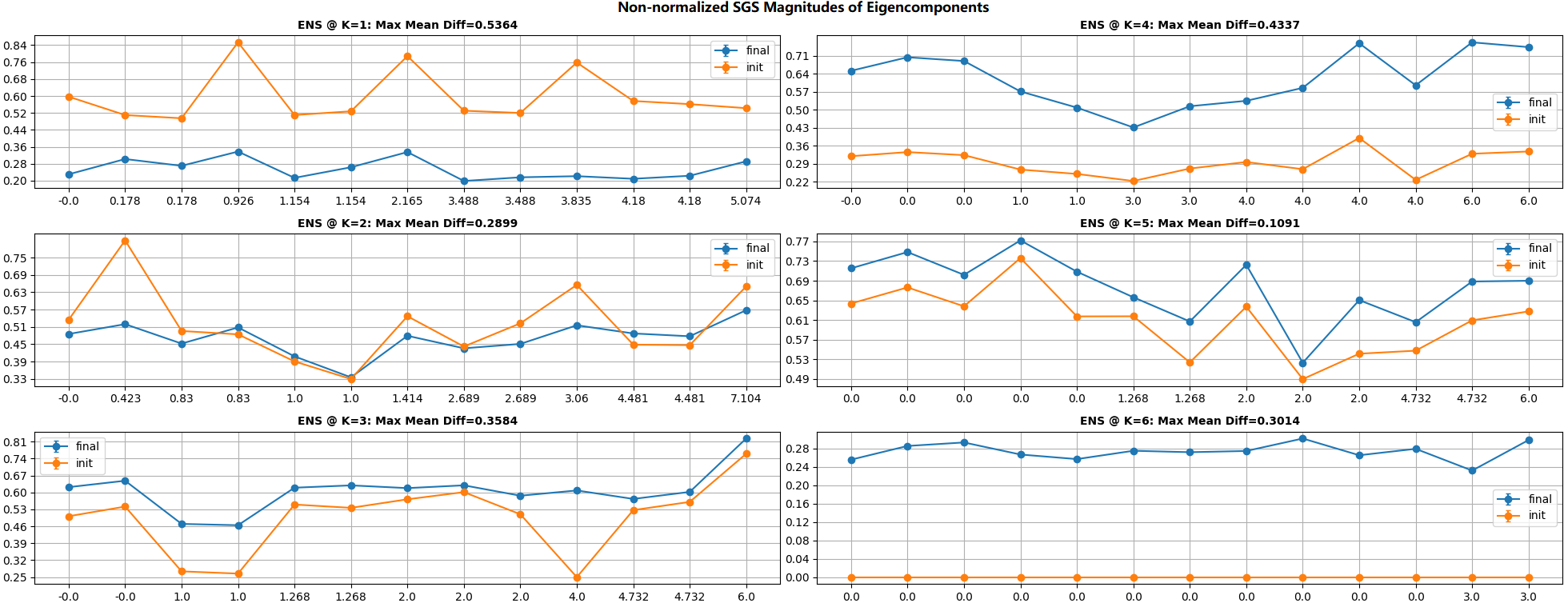}
	\caption{\textbf{Task 3}. $\mathcal{M}_{ENS}^K$'s of the initial condition (in yellow and labeled as ``init") and the low-pass filtered signal (in blue and labels as ``final") \protect\footnotemark.}
	\label{fig:low_pass_spectral_analysis_ens_non_norm}
\end{figure}

The norms of $\mathcal{M}_{ENS}^K$ help address this problem, and $\mathcal{M}_{ENS}^K$'s of the initial condition and the filtered signal are compared and illustrated in Figure \ref{fig:low_pass_spectral_analysis_ens_non_norm}. Two featured observations affirm the smoothing effect of the low-pass filter and the ``anit-smoothing" effect of the regularizer. First, at $K=1$, the amplitude of $\mathcal{M}_{ENS}^1$ of the initial condition is significantly higher than that of the filtered signal. This strongly evidences the smoothing effect. Second, at $K \geq 3$, the relations of amplitudes reverse, which clearly attributes to the regularizer impeding the smoothing. In addition, as the total variation objective and the regularizer directly compete at $K=2$, the amplitudes therein are similar. 
\footnotetext{Note that the shapes of $\mathcal{M}_{ENS}^K$ curves may not be exactly the same as $\mathcal{M}_{\mathbb{1}ENS}^K$ curves due to the limitation of \textbf{ENS} discussed in \textbf{Algorithm 6}.}
\par
The following conclusions summarize the discussions on \textbf{Task 3}:
\begin{framed}
	\noindent\textbf{Conclusions of Task 3}\\
	(1) The SGS methods, especially \textbf{ENS} for convenience, are effective in capturing the effects of filtering by comparing the spectral characteristics between the input signal and the filtered signal. \\
	(2) In case of the spectral characteristics being indistinguishable, the norms of $\mathcal{M}^K$'s help reveal the true behaviors of the learning model. 
\end{framed}
In the next section, a more complex case study is demonstrated showing how the SGS methods help understand the behaviors of a smoothing based node embedding learning model suffering from the over-smoothing issue.
\par

\subsection{A Smoothing Based Node Embedding Learning Case Study}\label{sec:amplitude}
The objective of this case study is to demonstrate the utility of SGS methods in the model diagnostics. Specifically, in \textbf{Task 4}, an ill node embedding learning, as the considered scenario, is performed and empirically evaluated. The learning model is constructed from a modification of the one used in \textbf{Task 3} by dropping the regularizer $\varepsilon$. Doing this leads to the model degenerating to a plain smoothing. Also, as a part of the scenario, the learning is controlled to run into over-smoothing (\citet{chen2020measuring}; \citet{zhao2019pairnorm}; \citet{li2018deeper}). It is shown that the resulting embeddings are not stable in node clustering (i.e. ARIs and AMIs can vary in a wide range). The over-smoothing is suspected to be the essential cause of this phenomenon. The diagnostics aims to justify this claim. Beforehand, it is necessary to show that the unstableness is a pathological issue rooting in the learning model rather than being caused by the settings. For this purpose, in \textbf{Task 5} and \textbf{6} respectively, biased initial conditions and missing local optima are precluded from causing the unstableness. Then, in \textbf{Task 7}, the resulting embeddings are profiled in the spectral domain, and the over-smoothing is identified. Finally, in \textbf{Task 8}, it hosts a discussion on interpreting the performance of node clustering in the over-smoothed learning. 
\par

\begin{framed}
	\noindent\textbf{Task 4: Unstableness of Smoothing Based Node Embedding Learning Model}\\
	$\triangleright$ \textbf{Objective:}
	\par
	Show the unstableness of the considered node embedding learning model in the node clustering task. The $0.1$ and $0.9$ percentiles of ARIs and AMIs of all trials are examined. The larger the spans of ARIs and AMIs between the percentiles, the stronger unstableness. 
	\par
	\noindent
	$\triangleright$ \textbf{Settings:}
	\par
	$\bullet$ The same graph in \textbf{Task 3} is used (visualized in Figure \ref{fig:sg_sample}).
	\par
	$\bullet$ The shallow learning model in \textbf{Task 3} is modified exclusively preserving $\tau$ (\textbf{Equation} \ref{eq:tv_objective}) in the objective.
	\par
	$\bullet$ The initial condition is a normalized $3$-dimensional random real-valued vector for each node. 
	\par
	$\bullet$ The terminal condition is exclusively ruled by a fixed number of epochs, $1000$, guaranteeing the over-smoothing. 
	\par
	\noindent
	$\triangleright$ \textbf{Trials:}
	\par
	$\bullet$ 500 independent node embedding learning trials.
	\par
	\noindent
	$\triangleright$ \textbf{Steps:}
	\par
	(1) Learn a 3-dimensional node embedding vector for each node, and the vector is $l_2$-normalized.
	\par
	(2) Cluster nodes by the spectral clustering, and compute ARI and AMI for each trial.
	\par
	(3) Examine the flatness of ARI and AMI distributions over all trials. 
\end{framed}
\par
\begin{figure}
	\centering
	\includegraphics[width=1\textwidth]{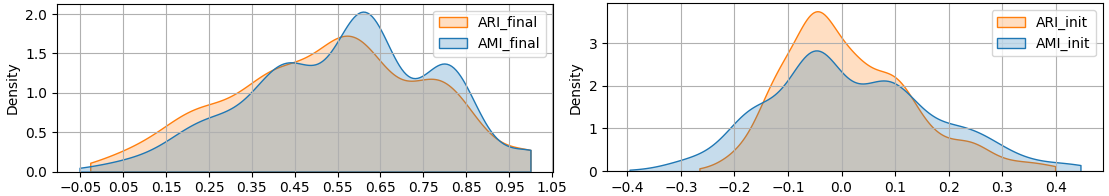}
	\caption{\textbf{Task 4}. The distributions of ARIs and AMIs on the initial conditions (right) and the learned node embeddings (left) respectively over all trials.}
	\label{fig:nc_perf_ep1000}
\end{figure}

The results of \textbf{Task 4} are shown in Figure \ref{fig:nc_perf_ep1000}. For the learned embeddings, the spans of ARIs and AMIs between the $0.1$-percentile and the $0.9$-percentile are about $0.60$; and for the initial conditions, the spans are about $0.35$. Thue, the distributions of ARIs and AMIs on the learned embeddings are flat over a wide range, and much flatter than those on the initial conditions. These observations strongly evidence the unstableness of the model. Nevertheless, the unstableness does not necessarily imply the over-smoothing. To eliminate the possibilities of other causes, in \textbf{Task 5} and \textbf{6}, the initial and terminal conditions respectively are justified not causing the unstableness.
\par

\begin{framed}
	\noindent\textbf{Task 5: Initial Conditions and Unstableness}\\
	$\triangleright$ \textbf{Objective:}
	\par
	Examine if the initial conditions (in \textbf{Task 4}) are biased and further causing the unstableness. Specifically, validated good and bad learned embeddings (in terms of the performance in node clustering) are collected, and it is examined if they can be distinguished by the spectral characteristics of their initial conditions. Strong evidence indicating the indistinguishability implies the unbiasedness of initial conditions.
	\par
	\noindent
	$\triangleright$ \textbf{Settings:}
	\par
	$\bullet$ Set the threshold of ARI and AMI for good embeddings to be $\geq 0.8$, and set the threshold for bad embeddings to be $\leq 0.3$.
	\par
	\noindent
	$\triangleright$ \textbf{Steps:}
	\par
	(1) Select all good and bad node embeddings by the criteria. 
	\par
	(2) For each $K$ and for each selected embedding, compute $\mathcal{M}_{\mathbb{1}}^K$'s and $\mathcal{M}^K$'s. Note that \textbf{LN-VX} is configured with $20$ learning trials.
	\par
	(3) Examine $\nabla \mathcal{M}_{\mathbb{1}}^K(i) = \big|\mathcal{M}_{good\mathbb{1}}^K(i) - \mathcal{M}_{bad\mathbb{1}}^K(i)\big| \in [0, 1]$ of the initial conditions of the good and bad embeddings. The lower the values of $\nabla \mathcal{M}_{\mathbb{1}}^K$'s, the less likely the initial conditions are biased.
	\par
	(4) Compute $\mathbb{E}\big[ \nabla \mathcal{M}^K_{\mathbb{1}} \big]$'s and $||\nabla \mathcal{M}^K_{\mathbb{1}}||_{\infty}$'s of the initial conditions of all good-bad embedding pairs (denoted by GB), good-good pairs (denoted by GG) and bad-bad pairs (denoted by BB). Then, compute the Wasserstein distances of $\mathbb{E}\big[ \nabla \mathcal{M}^K_{\mathbb{1}} \big]$'s and $||\nabla \mathcal{M}^K_{\mathbb{1}}||_{\infty}$'s respectively between GB and GG, GB and BB, as well as GG and BB. Note that the Wasserstein distances of $\mathbb{E}\big[ \nabla \mathcal{M}^K_{\mathbb{1}} \big]$'s and $||\nabla \mathcal{M}^K_{\mathbb{1}}||_{\infty}$'s are bounded between $0$ and $1$ because $\nabla \mathcal{M}^K_{\mathbb{1}} \in [0, 1]$. The lower the distances, the stronger the indistinguishability of the good and bad embeddings from the initial condition perspective, and thus the less likely the initial conditions are biased.
	\par
	(5) As $||\mathcal{M}^K||_2$'s are significantly affected by $||\nabla s^K||_2$, it is also necessary to justify that the $||\nabla s^K||_2$'s of initial conditions do not introduce bias into the learning. Specifically, compute the Pearson correlation coefficients (PPMCCs) \protect\footnotemark between ARIs (resp. AMIs) and $||\nabla s^K||_2$'s to examine if the good and bad embeddings can be distinguished by $||\nabla s^K||_2$'s. The lower the PPMCCs, the less likely $||\nabla s^K||_2$'s introducing bias. The PPMCCs between ARIs (resp. AMIs) and $||\mathcal{M}^K||_2$'s are also examined, and the results are expected to be similar to those on $||\nabla s^K||_2$'s.
\end{framed}
\footnotetext{Spearman's $\rho$ can also be used in this step, and it has been verified to produce very similar results. Thus, only the PPMCCs are discussed in this task.}
\par
At the step (1), $82$ good embeddings and $84$ bad embeddings out of $500$ trials are found \footnote{Two videos respectively visualizing the learning processes of a good embedding and a bad embedding can be found in Appendix \ref{sec:sup_file_list} \textbf{SUP-T5-2} and \textbf{SUP-T5-3}.}. 
\par
\begin{figure}
	\centering
	\includegraphics[width=1\textwidth]{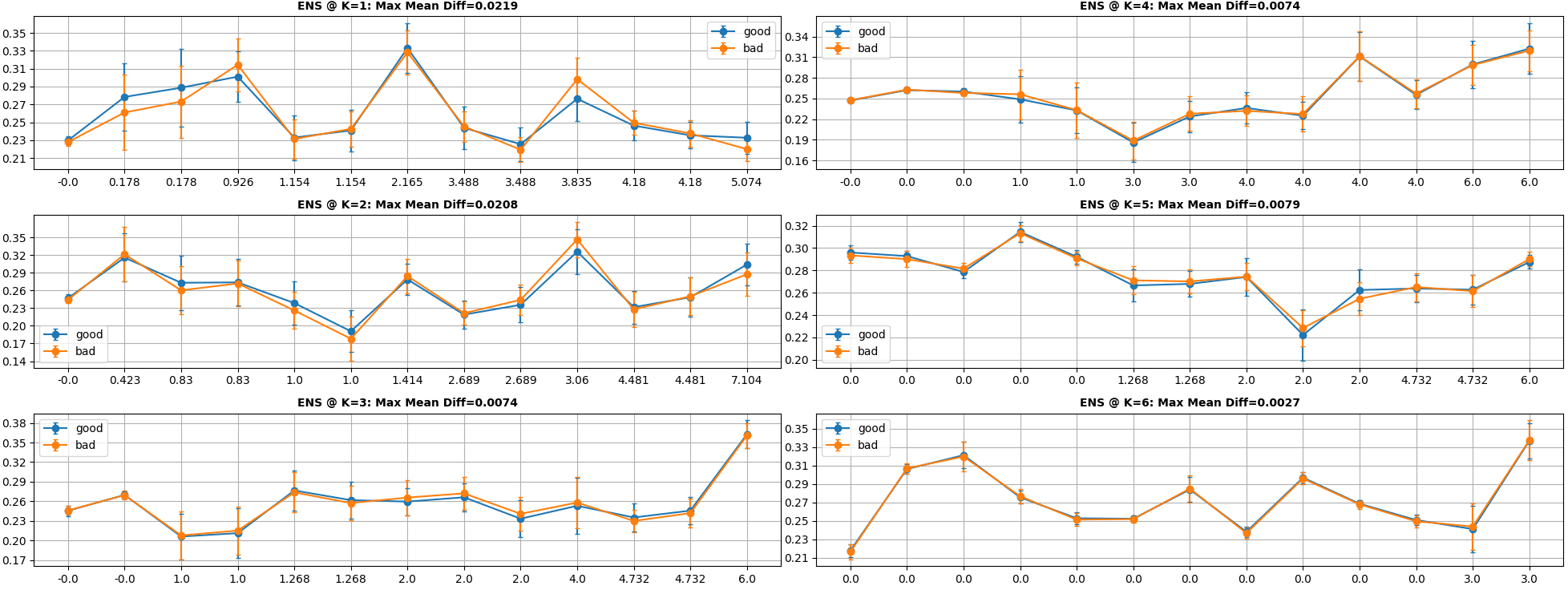}
	\caption{\textbf{Task 5}. The $\mathcal{M}^K_{\mathbb{1}ENS}$ of the initial conditions of the good and bad embeddings \protect\footnotemark. \textbf{Max Mean Diff}s are the maximum values of $\nabla \mathcal{M}_{\mathbb{1}}^K$'s between the good and bad curves.}
	\label{fig:good_vs_bad_init_only_ens_norm}
\end{figure}

At the step (3), $\mathcal{M}_{\mathbb{1}}^K$'s of the good and bad embeddings are illustrated in Figure \ref{fig:good_vs_bad_init_only_ens_norm}. And thus $\nabla \mathcal{M}_{\mathbb{1}}^K$'s are the differences between the magnitude curves of the good and bad embeddings. Particularly, the maximum means of $\nabla \mathcal{M}_{\mathbb{1}}^K$'s are remarked by \textbf{Max Mean Diff}s which are typically lower than $0.03$ over all $K$'s. This implies that the good and bad embeddings could hardly be distinguished by their $\mathcal{M}_{\mathbb{1}}^K$'s. To further justify the indistinguishability over all initial conditions with respect to $\mathcal{M}_{\mathbb{1}}^K$, the step (4) is performed. 
\par
\begin{figure}
	\centering
	\includegraphics[width=1\textwidth]{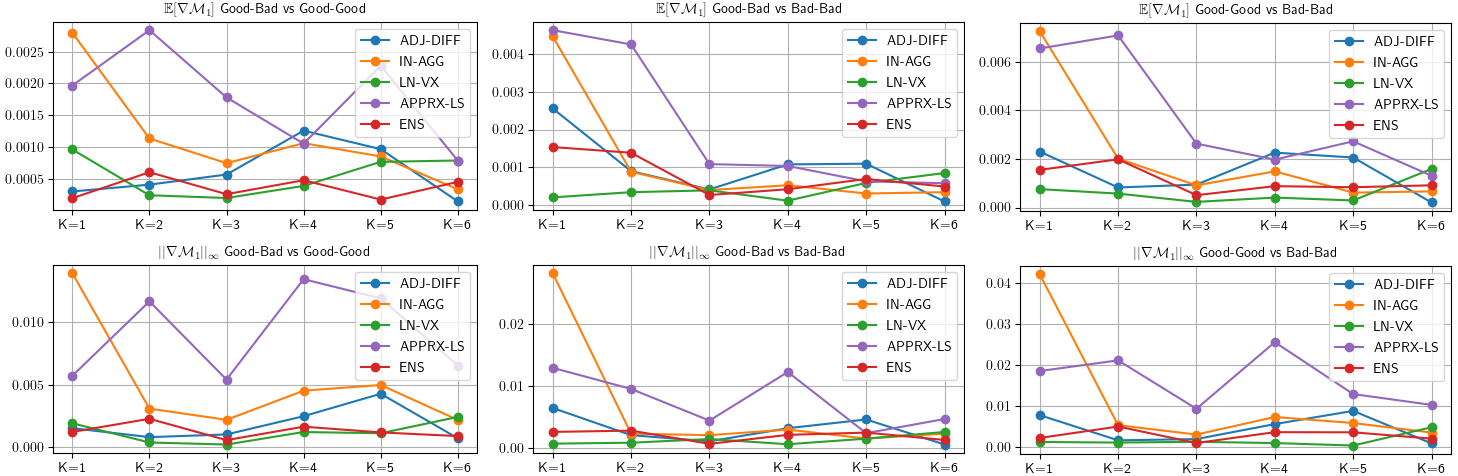}
	\caption{\textbf{Task 5}. The Wasserstein distances of GB-GG, GB-BB, and GG-BB with respect to $\mathcal{M}_{\mathbb{1}}^K$ of the initial conditions.}
	\label{fig:sgs_delta_distr_wass_init}
\end{figure}

At the step (4), the Wasserstein distances of the three pairs (i.e. GB-GG, GB-BB, and GG-BB) respectively are illustrated in Figure \ref{fig:sgs_delta_distr_wass_init}. The distances of $\mathbb{E}\big[ \nabla \mathcal{M}^K_{\mathbb{1}} \big]$'s are lower than $0.008$, and those of $||\nabla \mathcal{M}^K_{\mathbb{1}}||_{\infty}$'s are lower than $0.05$. These low distances strongly imply the indistinguishability over all initial conditions, and thus further justify the unbiasedness of the initial conditions.
\par
\begin{figure}
	\centering
	\includegraphics[width=0.8\textwidth]{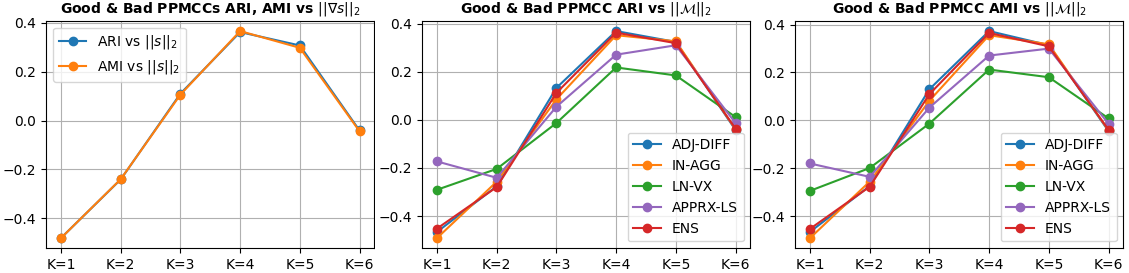}
	\caption{\textbf{Task 5}. The PPMCCs between ARIs (resp. AMIs) and $||s||_2$'s, and the PPMCCs between ARIs (resp. AMIs) and $||\mathcal{M}^K||_2$'s over $K$'s.}
	\label{fig:ppmcc_ari_ami_vs_init_sgs_amp}
\end{figure}

At the step (5), the resulting PPMCCs are shown in Figure \ref{fig:ppmcc_ari_ami_vs_init_sgs_amp}. The values of PPMCCs are clamped in the range $[-0.5, 0.4]$ over all $K's$. This range fails to indicate a strong correlation between the performance of node clustering and $||\nabla s||_2$'s (resp. $||\mathcal{M}^K||_2$'s), which justifies that $||\nabla s||_2$'s of initial conditions are not likely to introduce biases into the learning.
\par
\footnotetext{Full sets of $\mathcal{M}_{\mathbb{1}}^K$'s can be found in Appendix \ref{sec:sup_file_list} \textbf{SUP-T5-1}.}

\begin{framed}
	\noindent\textbf{Task 6: Learning Processes and Unstableness}\\
	$\triangleright$ \textbf{Objective:}
	\par
	Examine the existence of missing local optima potentially caused by ill-controlled learning processes. Specifically, it is examined if there exist turning points in the trend of the curves of $\tau$, ARI and AMI over all epochs. The absence of turning point implies little likelihood of missing local optima, and thus, if so, missing local optima could hardly be a cause of the unstableness.
	\par
	\noindent
	$\triangleright$ \textbf{Settings:}
	\par
	$\bullet$ All intermediate node embeddings, $\tau$'s, ARIs and AMIs of both the good and bad embeddings over all epochs are collected during the learning processes.
	\par
	\noindent
	$\triangleright$ \textbf{Steps:}
	\par
	(1) For each intermediate node embedding, compute $\mathcal{M}_{\mathbb{1}}^K$'s and $\mathcal{M}^K$'s.
	\par
	(2) Differentiate the curves of $\tau$, ARI, AMI, $\mathcal{M}_{\mathbb{1}}^K(i)$ and $\mathcal{M}^K(i)$ for each eigencomponent on epoch (denoted by $t$) to examine the existence of turning points in the trend. 
\end{framed}
\par
\begin{figure}
	\centering
	\includegraphics[width=1\textwidth]{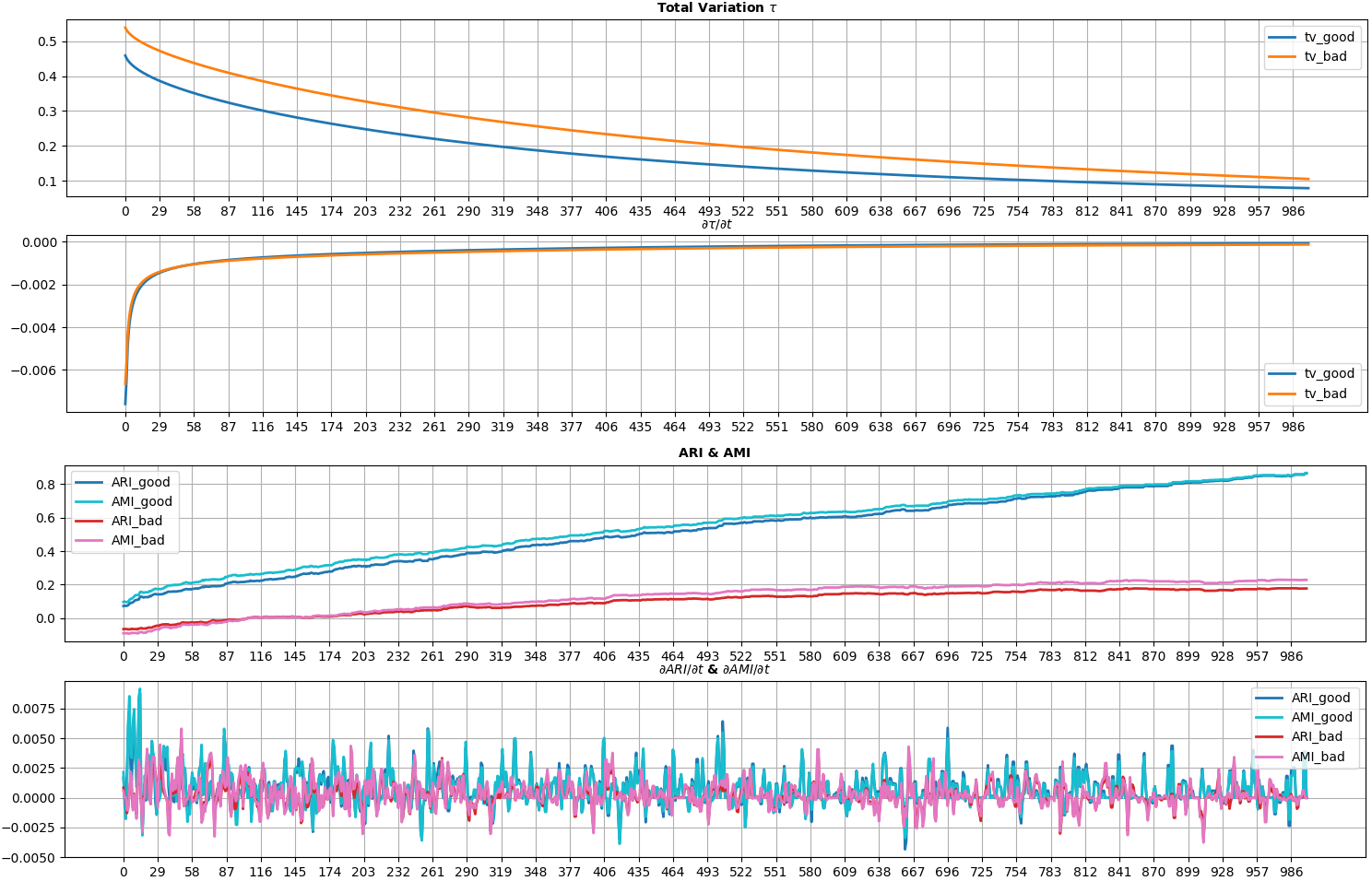}
	\caption{\textbf{Task 6}. $\tau$ and $\frac{\partial \tau}{\partial t}$ are shown in the upper two rows. ARI (resp. AMI) and $\frac{\partial \text{ARI}}{\partial t}$ (resp. $\frac{\partial \text{AMI}}{\partial t}$) are shown in the lower two rows. $t$ denotes epochs in the learning.}
	\label{fig:grad_tv_ari_ami}
\end{figure}

The differentiation of $\tau$, ARI and AMI is shown in Figure \ref{fig:grad_tv_ari_ami}. The curves of $\frac{\partial \tau}{\partial t}$ for both good and bad embeddings are smooth and asymptotic to $0$. This trend is a sign of over-smoothing according to the meaning of $\tau$. The curves of $\frac{\partial \text{ARI}}{\partial t}$ and $\frac{\partial \text{AMI}}{\partial t}$ are oscillating but bounded in a narrow range (i.e. bounded mean oscillation functions), and the trend is close to $0$. These observations indicate that ARIs and AMIs do not have sharp turning point in their trends during the learning, and thus no better solution is missing up to the current learning processes. 
\par
\begin{figure}
	\centering
	\includegraphics[width=1\textwidth]{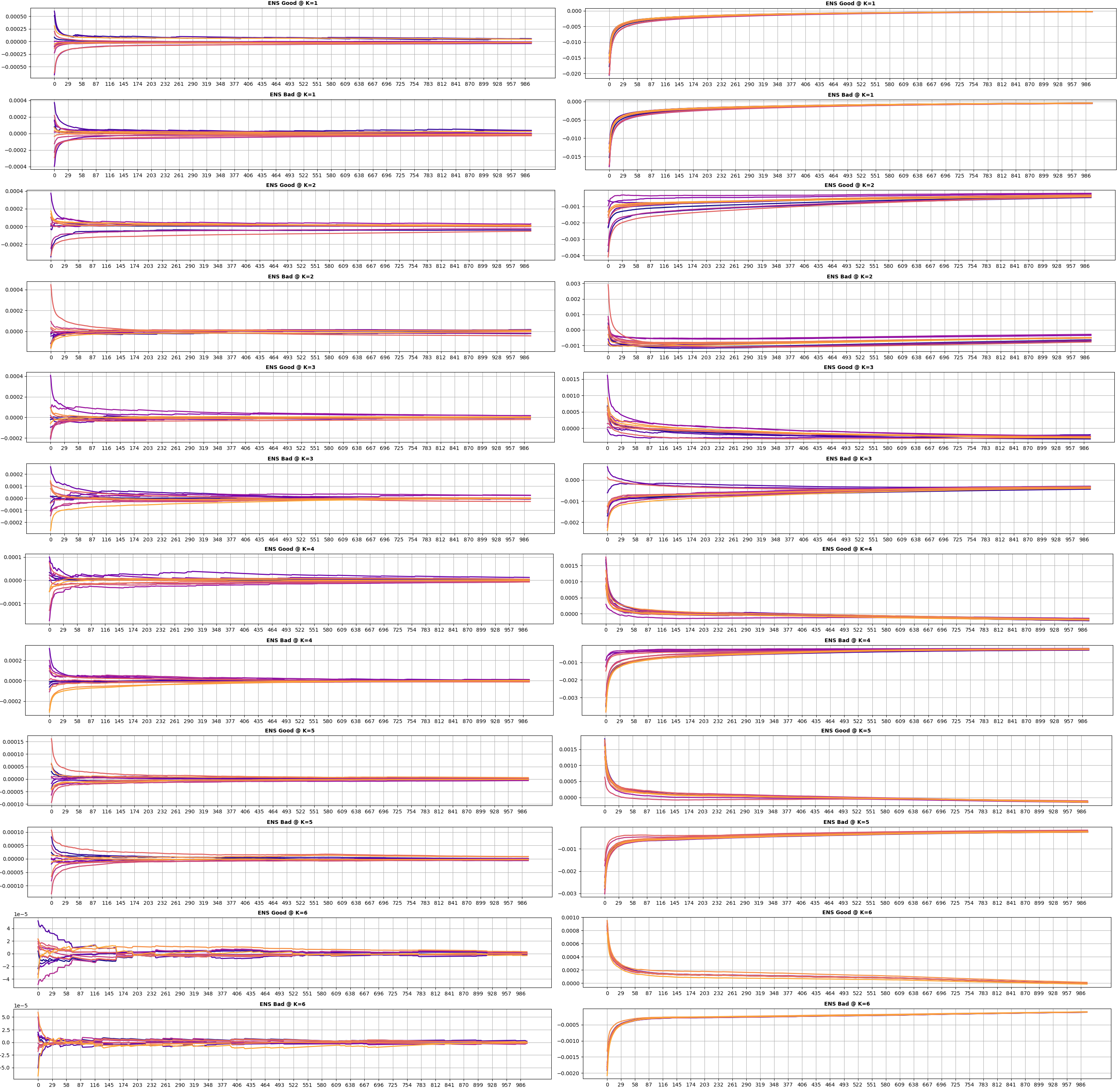}
	\caption{\textbf{Task 6}. $\frac{\partial \mathcal{M}_{ENS\mathbb{1}}^K(i)}{\partial t}$'s (left) and $\frac{\partial \mathcal{M}_{ENS}^K(i)}{\partial t}$'s (right) of good and bad embeddings respectively for all eigencomponents over $K$'s \protect\footnotemark. Darker curves correspond to lower-frequency eigencomponents, and lighter curves correspond to higher-frequency eigencomponents.}
	\label{fig:ens_grad}
\end{figure}

On the other hand, the curves of $\frac{\partial \mathcal{M}_{ENS\mathbb{1}}^K(i)}{\partial t}$ and $\frac{\partial \mathcal{M}_{ENS}^K(i)}{\partial t}$ are shown in Figure \ref{fig:ens_grad}. All curves therein are generally smooth and all asymptotic to $0$. This implies that the model encounters increasing damping in changing the magnitudes of eigencomponents in the learning processes, and, at the end of the processes, the magnitudes (or, in other words, the model) have reached a relatively inertial state. Thus, together with the conclusion attained from $\frac{\partial \text{ARI}}{\partial t}$ and $\frac{\partial \text{AMI}}{\partial t}$, not only no local optima are missing, but also it would be not very likely to gain significant improvement in performance even if the learning processes kept running beyond the $1000$ epochs. 
\par
In \textbf{Task 5} and \textbf{6}, it has been justified that the unstableness is pathological, and not caused by biased initial conditions or missing local optima. In \textbf{Task 7}, the resulting embeddings are profiled in the spectral domain, and the over-smoothing is identified. 
\par
\footnotetext{Full sets of $\mathcal{M}_{\mathbb{1}}^K$'s and $\mathcal{M}^K$'s over epochs as well as full sets of  $\frac{\partial \mathcal{M}_{\mathbb{1}}^K(i)}{\partial t}$'s and $\frac{\partial \mathcal{M}^K(i)}{\partial t}$ can be found in Appendix \ref{sec:sup_file_list} \textbf{SUP-T6-1}.}

\begin{framed}
	\noindent\textbf{Task 7: Profile Embeddings in Spectral Domain and Identify Over-Smoothing}\\
	$\triangleright$ \textbf{Objective:}
	\par
	Profile the spectral characteristics of learned embeddings, and identify the over-smoothing phenomenon. Specifically, a couple of spectral patterns of the over-smoothing are particularly examined. Since the smoothing objective (i.e. $\tau$) does not force non-adjacent nodes to repel each other, and, under this objective, the backpropagation, as an innate behavior of the learning, always prefers to ``penalizing" the adjacent nodes that are most dissimilar, then as $\tau$ keeps declining, the dissimilarities between adjacent nodes trend to become increasingly low and uniform. In consequence, theoretically, (i) the resulting embeddings are expected to have indistinguishable spectral characteristics, regardless of their performance in node clustering; and (ii) the initial conditions and the resulting embeddings are expected to have similar spectral characteristics up to scaling. 
	\par
	\noindent
	$\triangleright$ \textbf{Settings:}
	\par
	$\bullet$ $\mathcal{M}^K$'s and $\mathcal{M}_{\mathbb{1}}^K$'s of both the initial conditions and the resulting embeddings of good and bad instances are considered. 
	\par
	\noindent
	$\triangleright$ \textbf{Steps:}
	\par
	(1) Similarly to the steps (3), (4) and (5) of \textbf{Task 5}, compute $\nabla \mathcal{M}^K_{\mathbb{1}}$'s of all good and bad resulting embeddings, and compute the Wasserstein distances of $\mathbb{E}\big[ \nabla \mathcal{M}^K_{\mathbb{1}} \big]$'s and $||\nabla \mathcal{M}^K_{\mathbb{1}}||_{\infty}$'s examining if the good and bad embeddings can be distinguished by their $\mathcal{M}^K_{\mathbb{1}}$'s.
	\par
	(2) To examine if the initial conditions and the resulting embeddings can be distinguished by their $\mathcal{M}^K_{\mathbb{1}}$'s, the steps (1) is modified and performed. The good and bad embeddings are substituted by the initial conditions and the resulting embeddings of all good and bad instances. Also, for abbreviation, IF, FF and II denote initial-final embedding pairs, final-final pairs and initial-initial pairs respectively.
\end{framed}
\par
\begin{figure}
	\centering
	\includegraphics[width=1\textwidth]{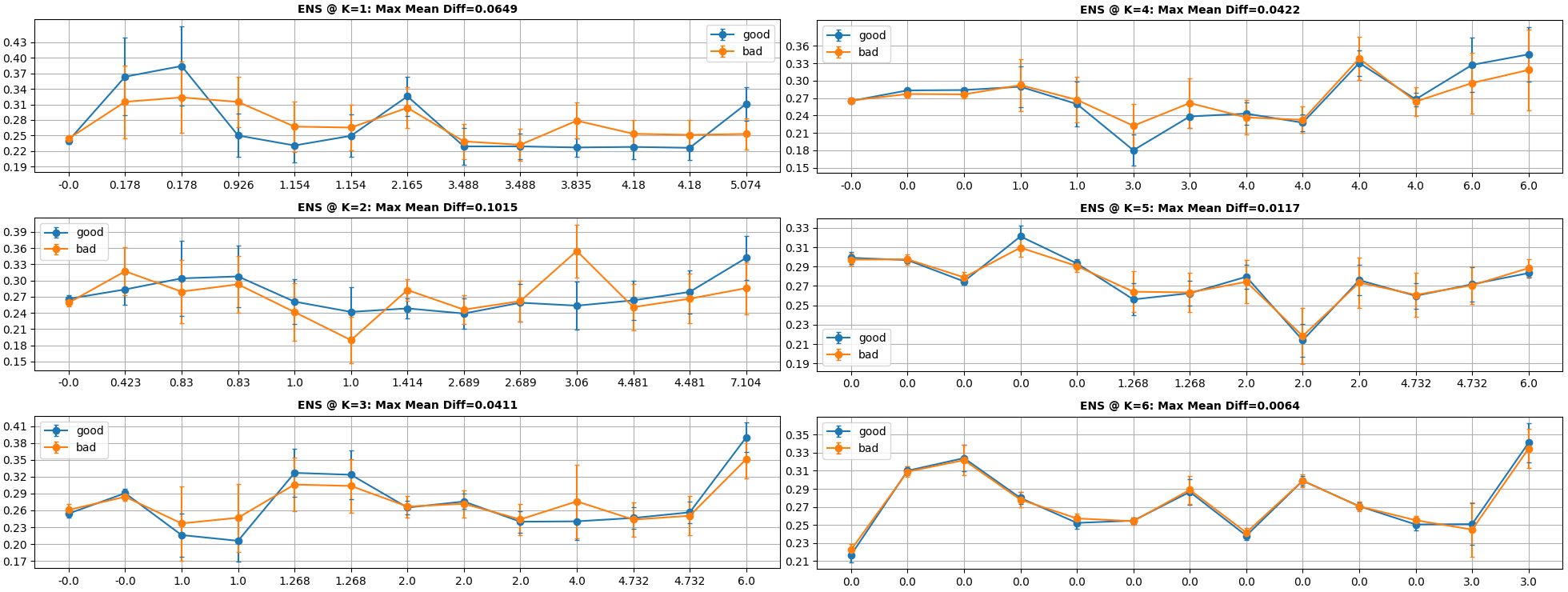}
	\caption{\textbf{Task 7}. The $\mathcal{M}_{ENS\mathbb{1}}^K$'s of the resulting good and bad embeddings \protect\footnotemark.}
	\label{fig:good_vs_bad_final_only_ens_norm}
\end{figure}
\begin{figure}
	\centering
	\includegraphics[width=1\textwidth]{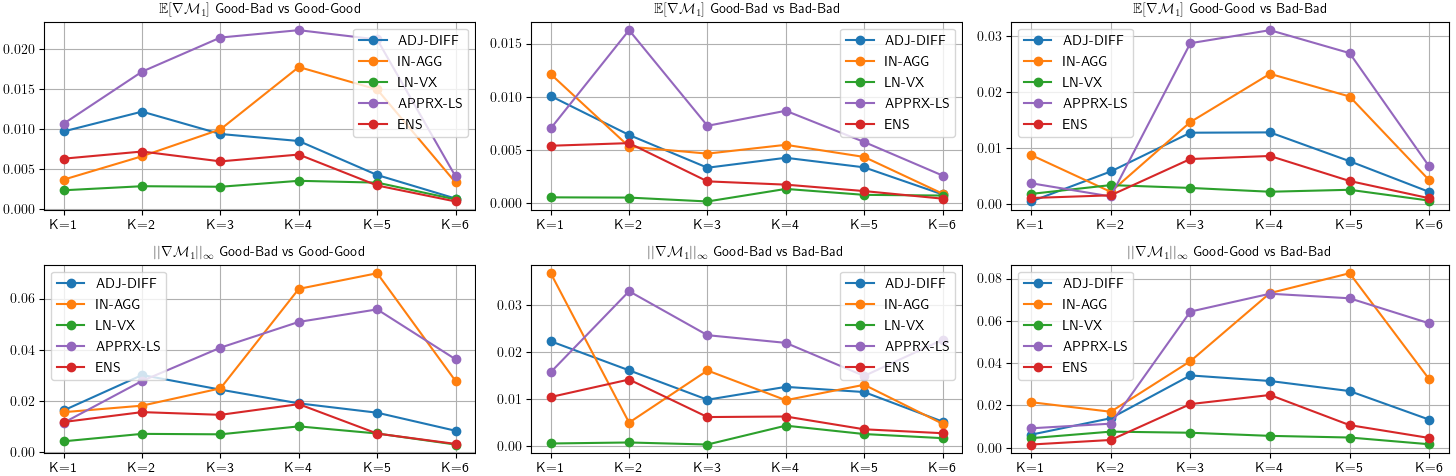}
	\caption{\textbf{Task 7}. The Wasserstein distances of GB-GG, GB-BB, and GG-BB with respect to $\mathcal{M}_{\mathbb{1}}^K$ of the resulting embeddings.}
	\label{fig:sgs_delta_distr_wass_final}
\end{figure}

The results at the step (1) are illustrated in Figures \ref{fig:good_vs_bad_final_only_ens_norm} and \ref{fig:sgs_delta_distr_wass_final}. The \textbf{Max Mean Diff}s of $\nabla \mathcal{M}^K_{\mathbb{1}}$ between the good and bad embeddings, as shown in Figure \ref{fig:good_vs_bad_final_only_ens_norm}, are lower than $0.07$, which strongly evidences the indistinguishability between the $\mathcal{M}^K_{\mathbb{1}}$'s of the good and bad embeddings. The Wasserstein distances of $\mathbb{E}\big[ \nabla \mathcal{M}^K_{\mathbb{1}} \big]$'s, as shown in Figure \ref{fig:sgs_delta_distr_wass_final}, are lower than $0.04$, and those of $||\nabla \mathcal{M}^K_{\mathbb{1}}||_{\infty}$'s are typically lower than $0.08$. These results also justify the indistinguishability.
\par
\begin{figure}
	\centering
	\includegraphics[width=1\textwidth]{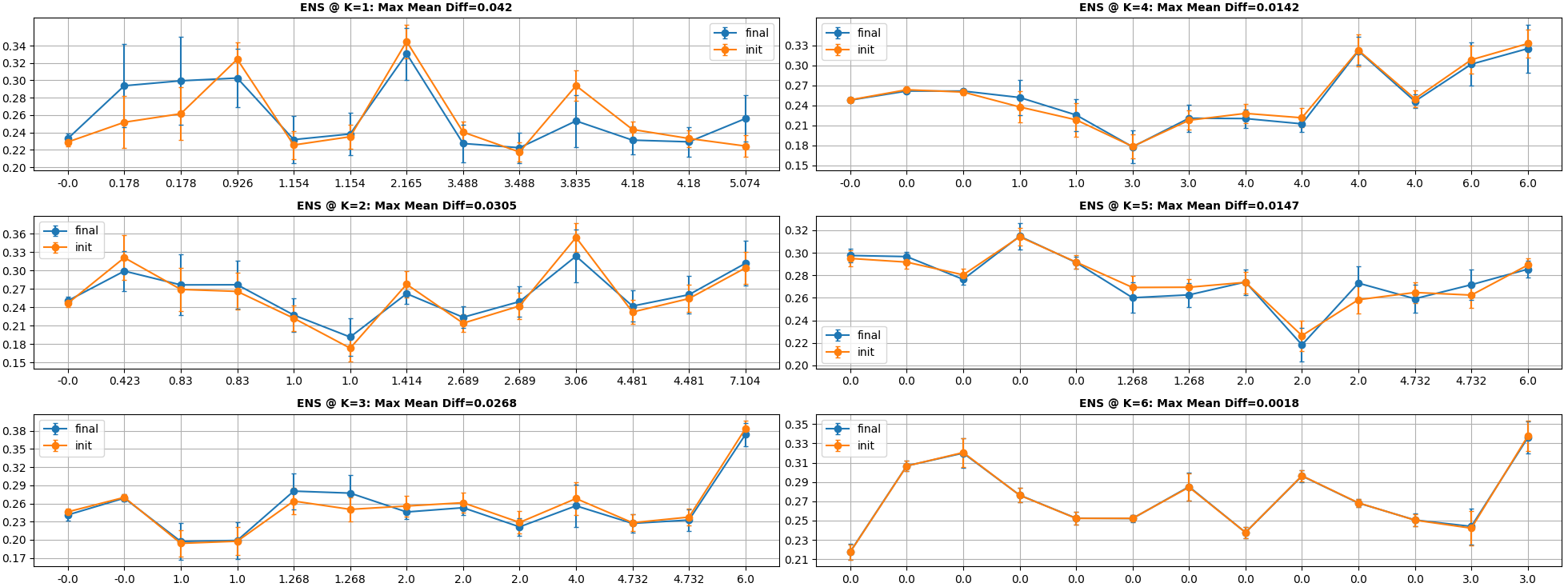}
	\caption{\textbf{Task 7}. The $\mathcal{M}_{ENS\mathbb{1}}^K$'s of the initial conditions and the resulting embeddings of the entire set of good and bad instances. \protect\footnotemark.}
	\label{fig:init_vs_final_ens_norm}
\end{figure}
\begin{figure}
	\centering
	\includegraphics[width=1\textwidth]{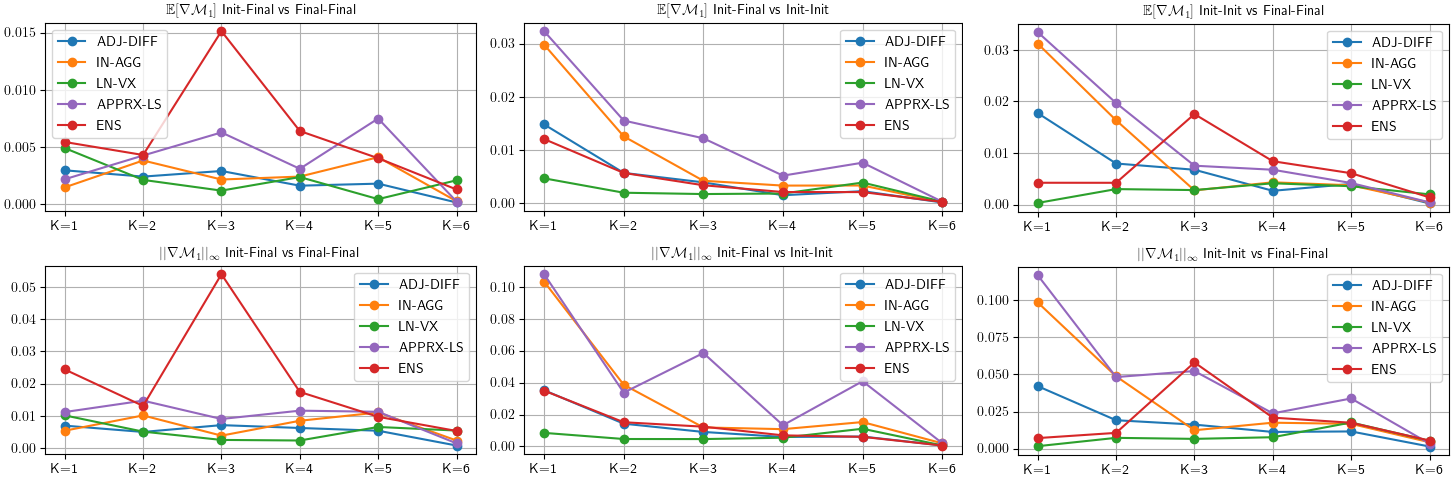}
	\caption{\textbf{Task 7}. The Wasserstein distances of IF-FF, IF-II, and II-FF with respect to $\mathcal{M}_{\mathbb{1}}^K$ of the initial conditions and the resulting embeddings.}
	\label{fig:sgs_delta_distr_wass_init_vs_final}
\end{figure}

The results at the step (2) are illustrated in Figure \ref{fig:init_vs_final_ens_norm} and \ref{fig:sgs_delta_distr_wass_init_vs_final}. The \textbf{Max Mean Diff}s of $\nabla \mathcal{M}^K_{\mathbb{1}}$ between the initial conditions and the resulting embeddings are lower than $0.05$. The Wasserstein distances of $\mathbb{E}\big[ \nabla \mathcal{M}^K_{\mathbb{1}} \big]$'s are lower than $0.04$, and those of $||\nabla \mathcal{M}^K_{\mathbb{1}}||_{\infty}$'s are lower than $0.13$. Clearly, the indistinguishability between the initial conditions and the resulting embeddings on $\mathcal{M}^K_{\mathbb{1}}$'s is justified by these results. 
\par
Therefore, the expected patterns of the over-smoothing are matched with these observations, which effectively identifies the over-smoothing phenomenon. Despite the effectiveness of SGS methods in the spectral profiling, in this task, one of the limitations is also witnessed. In the classic GSP, an over-smoothed (real-valued) signal is expected to have larger magnitudes at the zero eigencomponents (i.e. $\lambda_i = 0$) than all others. However, the SGS methods fail to reflect accurate magnitudes at zero eigencomponents. This failure can be observed from Figures \ref{fig:good_vs_bad_final_only_ens_norm} and \ref{fig:init_vs_final_ens_norm}, and it evidences the limitation common to all SGS methods as discussed at the end of Section \ref{sec:sgs}. 
\par
The last topic of this case study is to interpret the performance of the good and bad embedding in node clustering. The discussion is hosted in \textbf{Task 8}.
\par
\footnotetext{Full sets of result of $\mathcal{M}_{ENS\mathbb{1}}^K$'s of the resulting good and bad embeddings can be found in Appendix \ref{sec:sup_file_list} \textbf{SUP-T7-1}.}
\footnotetext{Full sets of result of $\mathcal{M}_{ENS\mathbb{1}}^K$'s of the initial conditions and the resulting embeddings can be found in Appendix \ref{sec:sup_file_list} \textbf{SUP-T7-2}.}
\par

\begin{figure}
	\centering
	\includegraphics[width=1\textwidth]{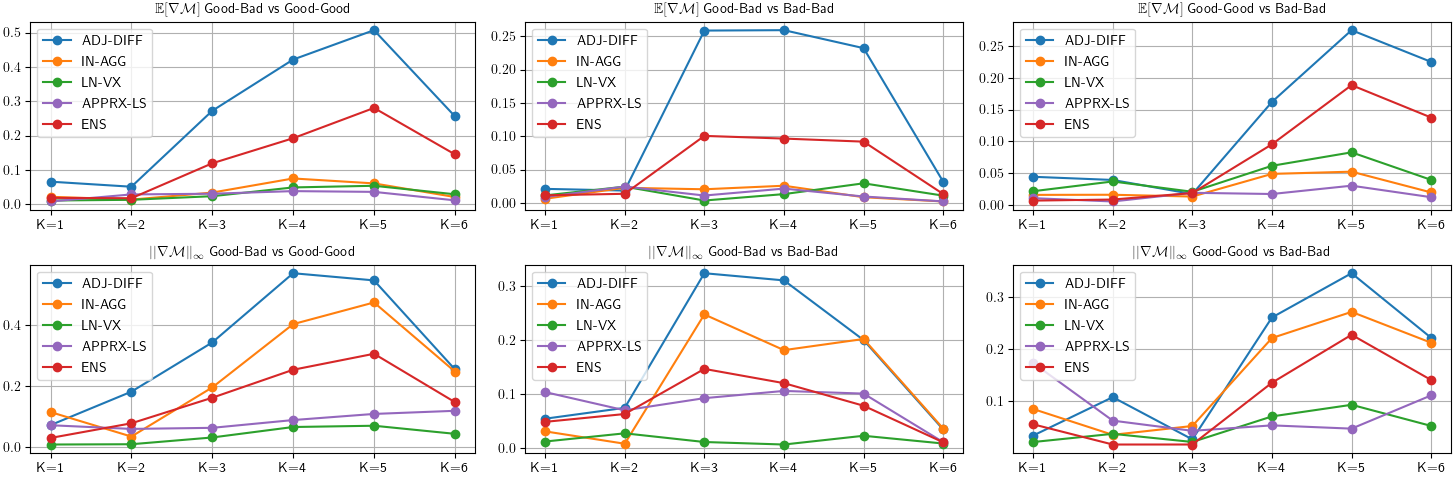}
	\caption{\textbf{Task 8}. The Wasserstein distances of GB-GG, GB-BB, and GG-BB with respect to $\mathcal{M}^K$ of the resulting embeddings.}
	\label{fig:sgs_delta_distr_wass_final_non_norm}
\end{figure}

\begin{figure}
	\centering
	\includegraphics[width=0.8\textwidth]{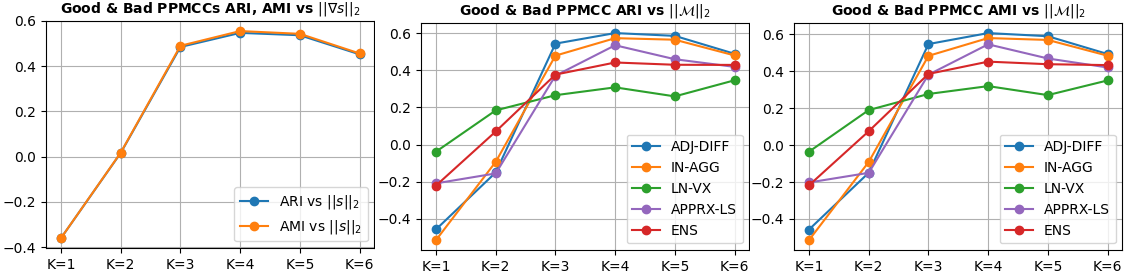}
	\caption{\textbf{Task 8}. The PPMCCs between ARIs (resp. AMIs) and $||s^K||_2$'s, and the PPMCC's between ARIs (resp. AMIs) and $||\mathcal{M}^K||_2$'s.}
	\label{fig:ppmcc_ari_ami_vs_final_sgs_amp}
\end{figure}

\begin{framed}
	\noindent\textbf{Task 8: Interpret Node Clustering Performance in Over-Smoothed Learning}\\
	$\triangleright$ \textbf{Objective:}
	\par
	Seek evidence to distinguish the resulting good and bad embeddings from $\nabla s^K$'s and $\mathcal{M}^K$'s.
	\par
	\noindent
	$\triangleright$ \textbf{Settings:}
	\par
	$\bullet$ $\nabla s^K$'s and $\mathcal{M}^K$'s of the resulting good and bad embeddings are considered. 
	\par
	\noindent
	$\triangleright$ \textbf{Steps:}
	\par
	(1) Compute the Wasserstein distances of $\mathbb{E}\big[ \nabla \mathcal{M}^K \big]$'s and $||\nabla \mathcal{M}^K||_{\infty}$'s and the PPMCCs between $||\nabla s^K||_2$'s (resp. $||\mathcal{M}^K||_2$'s) and ARIs and AMIs. The higher the values of the Wasserstein distances and the PPMCCs, the more distinguishable between the good and bad embeddings by $\nabla s^K$'s and $\mathcal{M}^K$'s.
\end{framed}
The Wasserstein distances are shown in Figure \ref{fig:sgs_delta_distr_wass_final_non_norm}, and the PPMCCs are shown in Figure \ref{fig:ppmcc_ari_ami_vs_final_sgs_amp}. Two observations are highlighted. First, the Wasserstein distances with respect to $\mathcal{M}^K$'s, especially at $K=4$ and $5$, are significantly greater than those with respect to $\mathcal{M}^K_{\mathbb{1}}$'s (as shown in Figure \ref{fig:sgs_delta_distr_wass_final}), though the values are not yet great enough to indicate strong distinguishability between the good and bad embeddings. As the essential difference between $\mathcal{M}^K$ and $\mathcal{M}^K_{\mathbb{1}}$ is that $||\nabla s^K||_2$ is preserved in $\mathcal{M}^K$, this observation is a weak sign of $||\nabla s^K||_2$ (or similarly $||\mathcal{M}^K||_2$) being able to distinguish the good and bad embeddings. Second, the PPMCCs, especially at $K=3, 4, 5$ and $6$, are non-trivial. Specifically, the PPMCCs between $||\nabla s^K||_2$'s and ARIs (resp. AMIs) are higher than $0.4$ at $K=3, 4, 5$ and $6$, and close to $0.6$ at $K=4$ and $5$. Moreover, $||\mathcal{M}^K||_2$'s have similar results. These correlations, though not strong enough to offer sufficient criteria in distinguishing the good and bad embeddings, indeed afford an interpretation explaining how the good embeddings are primarily different from the bad ones. Additionally, these correlations accord with the fact that is, being ruled by the objective $\tau$, when the over-smoothing happens, the resulting similarities between adjacent nodes in the $K$-SGs are not uniform over $K$'s, but rather, the lower the $K$, the more similar the adjacent nodes. Thus, at least a weak conclusion can be made on the interpretation of the node clustering performance. That is, the good resulting embeddings are more likely to have greater $||\nabla s^K||_2$'s and $||\mathcal{M}^K||_2$'s than bad ones, especially at higher $K$'s. In other words, at higher $K$'s, the adjacent nodes of good embeddings are not ``squeezed" as much as those of bad embeddings, which indirectly helps preserve the graph partition structures. 
\par
\par
The discusses from \textbf{Task 4} to \textbf{Task 8} have demonstrated a full use case of the SGS methods in model diagnostics, and the utility can be summarized as follows:
\begin{framed}
	\noindent\textbf{Conclusions of Tasks 4, 5, 6, 7 and 8}\\
	(1) The magnitudes of eigencomponents computed by the SGS methods are effective quantities in analyzing spectral characteristics of vector-valued signals by utilizing various methods (e.g. the differentiation over time and the PPMCC), especially $\nabla \mathcal{M}_{\mathbb{1}}^K$'s and $\nabla \mathcal{M}^K$'s.\\
	(2) Examining the magnitudes over different $K$'s helps make understanding behaviors of graph learning models and the model diagnostics more straightforward. 
\end{framed}

\section{Related Work}\label{sec:related_work}
The popular methods for computing magnitudes of eigencomponents for vector-valued signals have been discussed in Section \ref{sec:introduction}. In this section, some important work related to the SGs is briefed. 
\par
The SGs can be considered as a platform serving the ``multiresolution" spectral analysis. Typical studies in this area include graph wavelets (\citet{hammond2011wavelets}), the multiresolution matrix factorization (\citet{kondor2014multiresolution}), combining the spectral clustering and the Guassian kernel producing mutiresolution graph partitioning (\citet{park2004support}), utilizing the kernel method in multiresolution community detection (\citet{zhang2009seeding}) and etc. Particularly, the wavelets have been widely used in various applications especially in neural networks (\citet{zhang1995wavelet}; \citet{adeli2006dynamic}; \citet{chen2006time}; \citet{xu2019graph}; \citet{donnat2018learning}; \citet{rustamov2019wavelets}).
\par
In addition, the essence of SGs is to consider node relations beyond the 1-hop adjacency. In many real tasks on graphs (e.g. node clustering, classification, edge prediction and influence maximization), understanding such relations is critical. Plenty of studies have made progress on the related problems from random walk based models (\citet{wu2019simplifying}; \citet{abu2020n}; \citet{abu2019mixhop}) to line graph based models (\citet{chen2017supervised}), and from applications of the Hashimoto matrix (\citet{bordenave2015non}; \citet{morone2015influence}) to multiresolution analysis (\citet{kondor2014multiresolution}).

\section{Discussion \& Conclusion} \label{sec:discssion_and_conclusion}
The SGs serving landscapes at different levels of adjacency and the SGS methods computing the magnitudes of eigencomponents for vector-valued signals have been presented. In summary, these methods can be used as a general tool in analyzing vector-valued signals in the spectral domain, and thus they are useful in designing and diagnosing graph learning models from the node embedding perspective. Particularly, comparing signals from difference sources (e.g. randomly generated and model generated) with respect to their magnitudes of eigencomponents has been shown to be an effective approach in various use cases. Moreover, conducting such analysis over $K$'s helps gain more insightful understanding of the learning models. 
\par
The SGS methods, on the other hand, have a few limitations as primarily discussed in Section \ref{sec:sgs} and witnessed in Section \ref{sec:experiments}. \textbf{APPRX-LS} is sensitive to the structures of input graphs, and its performance depends on how the SVD-based approximation performs, which can be arbitrarily bad. \textbf{IN-AGG} is only effective on pulse-like signals. \textbf{ADJ-DIFF} is also sensitive to the structures of input graphs, and $\nabla u_i$'s being not orthogonal can undermine its performance. It also has a potential scaling issue. \textbf{LN-VX} relies on a learned transform which may not be unique, and it may suffer from numerical biases introduced by the transform. The ensemble method, \textbf{ENS}, has to face the scaling issue when using $\mathcal{M}^K$'s. Furthermore, all of the SGS methods are weak at handling the zero eigencomponents and constant signals. Finally, the disconnected components at higher $K$'s further weaken the effectiveness of the methods. These issues need to be paid particular attention in practice. The future work will concentrate on addressing these issues. 
\newpage

\appendix

\section{Supplementary Material File List}\label{sec:sup_file_list}

\subsection{Section \ref{sec:low_pass_filter} Task 3}
$\bullet$ \textbf{SUP-T3-1}: A video visualizing the entire learning process of the selected embedding: SUP-T3-1.mp4
\\
$\bullet$ \textbf{SUP-T3-2}: Full sets of eigenvalues and eigenvectors of all SGs: SUP-T3-2.zip
\\
$\bullet$ \textbf{SUP-T3-3}: Full sets of $\mathcal{M}_{\mathbb{1}}^K$'s of the initial condition and the selected resulting embedding: SUP-T3-3.zip

\subsection{Section \ref{sec:amplitude} Task 5}
$\bullet$ \textbf{SUP-T5-1}: Full Sets of $\mathcal{M}_{\mathbb{1}}^K$'s of the initial conditions of the good and bad embeddings: SUP-T5-1.zip
\\
$\bullet$ \textbf{SUP-T5-2}: A video visualizing the entire learning process of a good embedding: SUP-T5-2.mp4
\\
$\bullet$ \textbf{SUP-T5-3}: A video visualizing the entire learning process of a bad embedding: SUP-T5-3.mp4
\\
$\bullet$ \textbf{SUP-T6-1}: Full sets of $\mathcal{M}_{\mathbb{1}}^K$'s and $\mathcal{M}^K$'s over epochs, and full sets of  $\frac{\partial \mathcal{M}_{\mathbb{1}}^K(i)}{\partial t}$'s and $\frac{\partial \mathcal{M}^K(i)}{\partial t}$: SUP-T6-1.zip

\subsection{Section \ref{sec:amplitude} Task 7}
$\bullet$ \textbf{SUP-T7-1}: Full sets of $\mathcal{M}_{\mathbb{1}}^K$'s of the resulting good and the bad embeddings: SUP-T7-1.zip
\\
$\bullet$ \textbf{SUP-T7-2}: Full sets of $\mathcal{M}_{\mathbb{1}}^K$'s of the initial conditions and the resulting embeddings of the entire set of good and the bad instances: SUP-T7-2.zip

\vskip 0.2in
\bibliography{stratified_graph_spectra}

\end{document}